\documentclass[times,twocolumn,final]{elsarticle}

\usepackage{medima}
\usepackage{framed,multirow}

\usepackage{amssymb,amsfonts}
\usepackage[fleqn]{amsmath}
\usepackage{latexsym}
\usepackage{multirow}
\usepackage[table]{xcolor}
\usepackage{makecell}
\usepackage{gensymb}
\usepackage{balance}
\usepackage{float}
\usepackage{url}
\usepackage{array}
\usepackage[bottom]{footmisc}
\usepackage{booktabs}
\usepackage{xcolor}
\usepackage[colorlinks,
            linkcolor=cyan,
            urlcolor=cyan,
            citecolor=cyan,
            ]{hyperref}

\renewcommand\arraystretch{1.2}
\definecolor{newcolor}{rgb}{.8,.349,.1}

\journal{Medical Image Analysis}

\begin{document}
\verso{Li Lin \textit{et~al.}}

\begin{frontmatter}

\title{YoloCurvSeg: You Only Label One Noisy Skeleton for Vessel-style Curvilinear Structure Segmentation}


\author[1,2,3]{Li \snm{Lin}}

\author[1]{Linkai \snm{Peng}} 
\author[1,3]{Huaqing \snm{He}}
\author[1,3]{Pujin \snm{Cheng}}
\author[1]{Jiewei \snm{Wu}}
\author[2]{Kenneth K. Y. \snm{Wong}}

\author[1,3]{Xiaoying \snm{Tang}\corref{cor1}}
\cortext[cor1]{Corresponding author.}
\ead{tangxy@sustech.edu.cn}

\address[1]{Department of Electronic and Electrical
Engineering, Southern University of Science and Technology,
Shenzhen, China}
\address[2]{Department of Electrical and Electronic Engineering, the University of Hong Kong, Hong Kong, China}
\address[3]{Jiaxing Research Institute, Southern University of Science and Technology, Jiaxing, China}

\received{15 Jan 2023}


\begin{abstract}
Weakly-supervised learning (WSL) has been proposed to alleviate the conflict between data annotation cost and model performance through employing sparsely-grained (i.e., point-, box-,  scribble-wise) supervision and has shown promising performance, particularly in the image segmentation field. However, it is still a very challenging task due to the limited supervision, especially when only a small number of labeled samples are available. Additionally, almost all existing WSL segmentation methods are designed for star-convex structures which are very different from curvilinear structures such as vessels and nerves. In this paper, we propose a novel sparsely annotated segmentation framework for curvilinear structures, named YoloCurvSeg. A very essential component of YoloCurvSeg is image synthesis. Specifically, a background generator delivers image backgrounds that closely match the real distributions through inpainting dilated skeletons. The extracted backgrounds are then combined with randomly emulated curves generated by a Space Colonization Algorithm-based foreground generator and through a multilayer patch-wise contrastive learning synthesizer. In this way, a synthetic dataset with both images and curve segmentation labels is obtained, at the cost of only one or a few noisy skeleton annotations. Finally, a segmenter is trained with the generated dataset and possibly an unlabeled dataset. The proposed YoloCurvSeg is evaluated on four publicly available datasets (OCTA500, CORN, DRIVE and CHASEDB1) and the results show that YoloCurvSeg outperforms state-of-the-art WSL segmentation methods by large margins. With only one noisy skeleton annotation (respectively 0.14\%, 0.03\%, 1.40\%, and 0.65\% of the full annotation), YoloCurvSeg achieves more than 97\% of the fully-supervised performance on each dataset. Code and datasets will be released at 
\href{https://github.com/llmir/YoloCurvSeg}{https://github.com/llmir/YoloCurvSeg}.
\end{abstract}

\begin{keyword}
\MSC \\41A05\\ 41A10\\ 65D05\\ 65D17
\KWD 
\\Sparse Annotation
\\One-shot
\\Curvilinear Structure Segmentation
\\Weakly-supervised Learning
\\Medical Image Synthesis
\end{keyword}

\end{frontmatter}

\section{Introduction}
\label{sec1}

Curvilinear structures are elongated, curved, multi-scale structures that often appear tree-like and are commonly found in natural images (e.g., cracks and aerial road maps) and biomedical images (e.g., vessels, nerves and cell membranes). Automatic and precise segmentation of these curvilinear structures plays a significant role in both computer vision and biomedical image analysis. For example, road mapping serves as a prerequisite in both autonomous driving and urban planning. In the biomedical field, studies \citep{pritchard2014longitudinal, lin2021blu, kawasaki2009retinal, lin2020sustech} have suggested that the morphology and topology of specific curvilinear anatomy (e.g., retinal vessels and corneal nerve fibers) are highly relevant to the presence or severity of various diseases such as hypertension, arteriolosclerosis, keratitis, age-related macular degeneration, diabetic retinopathy, and so on. Retinal vessels are observable in retinal fundus images and optical coherence tomography angiography (OCTA) images, while corneal nerve fibers are identifiable in confocal corneal microscopy (CCM) images. It has been suggested that early signs of many ophthalmic diseases are reflected by microvascular and capillary abnormalities \citep{allon2021retinal,lin2021bsda}. Collectively, accurate segmentation of various curvilinear structures is of great importance for computer-aided diagnosis, quantitative analysis and early screening of various diseases, especially in ophthalmology.

\begin{figure}[t]
  \vspace{-0.1cm}
  \centering{\includegraphics[width=\columnwidth]{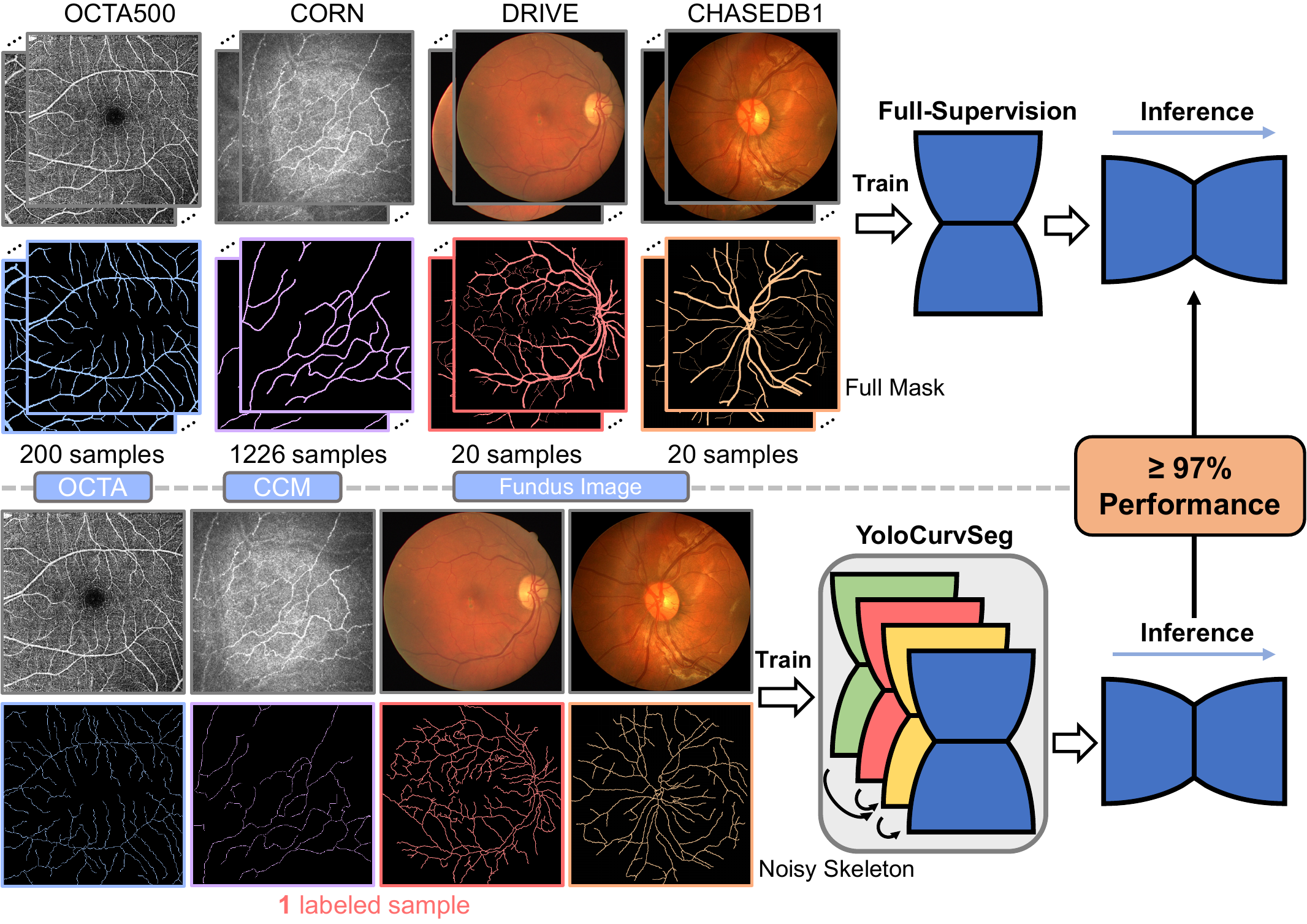}}
  \vspace{-0.7cm}
  \caption{YoloCurvSeg achieves more than 97\% of the fully-supervised performance on each of four representative datasets utilizing only one noisy skeleton annotation, which means physicians can save largely save labeling time and still obtain satisfactory segmentation results.}
  \label{fig1}
  \vspace{-0.45cm}
\end{figure}

In recent years, benefiting from the development of deep learning (DL), many DL-based segmentation algorithms for curvilinear structures have been proposed and have shown overwhelming performance compared to traditional (e.g., matched filter-based and morphological processing-based \citep{nguyen2013effective,singh2016retinal}) methods. Most existing works are dedicated to designing sophisticated network architectures \citep{peng2021fargo, mou2021cs2, he2022curv} and deploying strategies to preserve curvilinear structures' topology by employing generative adversarial networks (GANs) \citep{lin2021blu, son2019towards} or topology-preserving loss functions \citep{cheng2021joint, shit2021cldice}. These methods are typically fully-supervised, wherein large-scale well-annotated datasets are required. However, collecting and labeling a large-scale dataset with full annotation is very costly and time-consuming, particularly for medical images since their annotation requires expert knowledge and clinical experience. Furthermore, annotating curvilinear structures is even more challenging, given that curvilinear structures are slender, multi-scale, and complex in shape with fine details.

More recently, many efforts have been made to reduce the annotation cost for DL model training. For example, semi-supervised learning (SSL) trains models by combining limited amounts of annotated data with massive unlabeled data \citep{xu2022semicurv,hou2022semi, mittal2019semi}. While effective, most state-of-the-art (SOTA) SSL methods still require about 5\%-30\% of the accurately and precisely labeled data to achieve about 85\%-95\% of the fully-supervised performance, which is still not sufficiently cost-effective and still time-consuming when it comes to labeling curvilinear structures.
Weakly supervised learning (WSL) attempts to alleviate the annotation issue from another perspective by performing sparsely-grained (i.e., point-, scribble-, bounding box-wise) supervision and attains promising performance \citep{liang2022tree,lin2016scribblesup,tang2018normalized,tang2018regularized,kervadec2019constrained}. Compared with either point or bounding box, scribble is a relatively more flexible and generalizable form of sparse annotation that can be used to annotate complex structures \citep{luo2022scribble}. Existing scribble-supervised segmentation methods mainly fall into two categories. The first line of research exploits structural or volumetric priors to expand scribbles into more accurate pseudo proposals; for example, grouping pixels with similar grayscale intensities or locations into the same class \citep{liang2022tree,lin2016scribblesup,ji2019scribble}. However, the expansion process may introduce noisy proposals, which may induce error accumulation and deteriorate the performance of the segmentation model. Some work \citep{huo2021atso} also points out the inherent weakness of these methods, namely models retain their own predictions and thus resist updating. The second line learns adversarial shape priors utilizing extra unpaired but fully-annotated masks. Such approaches somewhat contradict the motivation of saving annotation costs, especially for complex curvilinear structures \citep{larrazabal2020post, valvano2021learning,zhang2020accl}.
Moreover, most WSL methods still require sparsely labeling the entire dataset (or a large portion), and they are mainly designed and validated on relatively simple structures (e.g., cardiac structures or abdominal organs) with assumptions and priors that may not apply to complex structures (e.g., curvilinear structures).

To address these aforementioned challenges, we here present a novel WSL segmentation framework for vessel-style curvilinear structures, namely \textit{You Only Label One Noisy Skeleton for Curvilinear Structure Segmentation (YoloCurvSeg)}. For curvilinear structures, label noises/errors are inevitable, and a good segmentation approach should be noise tolerant. Therefore, instead of utilizing only the annotated pixels for supervision, YoloCurvSeg ingeniously converts the weakly-supervised problem into a fully- or semi-supervised one via image synthesis. It employs a trained inpainting network as a background generator, which takes one (or multiple depending on availability) noisy skeleton (as shown in Fig. \ref{fig1}) and dilates it to serve as an inpainting mask to obtain a background that closely matches the real distribution. The extracted background is then augmented and combined with randomly emulated curves generated by a Space Colonization Algorithm-based foreground generator, from which a synthetic dataset is obtained through a multilayer patch-wise contrastive learning synthesizer. Finally, a segmenter performs coarse-to-fine two-stage segmentation using the synthetic dataset and an unlabeled dataset (if available).
Our main contributions are summarized as follows:
\begin{itemize}
  \vspace{-0.1cm}
  \item We propose a novel weakly-supervised framework for one-shot skeleton/scribble-supervised curvilinear structure segmentation, namely YoloCurvSeg. {To the best of our knowledge, YoloCurvSeg is a pioneering weakly-supervised segmentation method for curvilinear structures utilizing noisy and sparsely-annotated data.}
  \item YoloCurvSeg novelly converts a WSL problem into a fully supervised one through four steps: curve generation, image inpainting, image translation and coarse-to-fine segmentation. The proposed framework is noise-robust, sample-insensitive and easily extensible to various curvilinear structures.
  \item We evaluate YoloCurvSeg on four challenging curvilinear structure segmentation datasets, namely OCTA500 \citep{li2020ipn}, CORN \citep{zhao2020automated}, DRIVE \citep{staal2004ridge} and CHASEDB1 \citep{fraz2012ensemble}. Experimental results show that YoloCurvSeg outperforms SOTA WSL and noisy label learning methods by large margins. Meanwhile, we demonstrate that $\geq$ 97\% of the fully-supervised performance can be achieved with only one noisy skeleton label (approximately 0.1\% or 1\% of the full annotation), which shall also inspire subsequent works on WSL and curvilinear dataset construction.
\end{itemize}





\section{Related Works}
Related works mainly involve curvilinear structure segmentation, weakly-supervised segmentation and medical image synthesis, which we introduce below one by one.

\subsection{Curvilinear Structure Segmentation}
Existing automatic curvilinear structure segmentation algorithms can be roughly divided into two categories. The first category is traditional unsupervised methods, mainly including mathematical morphology methods and various filtering methods \cite{mou2021cs2}. 
For instance, \cite{zana2001segmentation} segment vascular-like patterns using a hybrid framework of morphological filtering and cross-curvature analysis. \cite{passat2006magnetic} present a preliminary approach to strengthen the segmentation of cerebral vessels by incorporating high-level anatomical knowledge into the segmentation process.
Filtering methods include Hessian matrix-based filters \citep{frangi1998multiscale}, matched filters \citep{singh2016retinal,hoover2000locating}, multi-oriented filters \citep{soares2006retinal}, symmetry filter \citep{zhao2017automatic}, etc. 
The other category is supervised methods, wherein data with ground truth labels are used to train segmenters based on predefined or model-extracted features. Traditional machine-learning-based approaches are dedicated to pixel-level classification using handcrafted features \citep{zhang2017retinal, holbura2012retinal}. Recently, DL-based approaches have made significant progress in various segmentation tasks. For example, \cite{ronneberger2015u} propose U-Net, which has been widely used in numerous medical image segmentation tasks.
Existing curvilinear structure segmentation works focus on well-designed network architectures by introducing multi-scale \citep{he2022curv, wu2018multiscale}, multi-task \citep{lin2021bsda, peng2021fargo, hao2022retinal}, or various attention mechanisms \citep{mou2021cs2, yu2022vision} 
as well as well-playing morphological and topological properties by introducing GANs or morphology-/topology-preserving loss functions \citep{cheng2021joint, shit2021cldice}. Still, data availability and annotation quality are the main limitations of these methods.

\subsection{Weakly-supervised Segmentation}

Weakly-supervised segmentation aims to reduce the labeling costs by training segmentation models on data annotated with coarse granularity \citep{liang2022tree}. Among various formats of sparse annotations, scribble is recognized as the most flexible and versatile one that can be used to annotate even very complex structures \citep{luo2022scribble,valvano2021learning}. Existing scribble-supervised segmentation methods fall into two main categories. The first one exploits structural or volumetric priors to expand scribble annotations by assigning a same class to pixels with similar intensities or nearby locations \citep{lin2023unifying, liang2022tree,lin2016scribblesup,ji2019scribble}. 
The main limitation of such approaches is that they heavily rely on pseudo proposals and often contain multiple stages, which can be time-consuming and prone to errors that may be propagated during model training.
The second category learns adversarial shape priors utilizing extra unpaired but fully-annotated masks. Such approaches somewhat contradict the motivation of saving annotation costs, especially for complex curvilinear structures \citep{larrazabal2020post, valvano2021learning,zhang2020accl}.
Additionally, these methods still require sparsely labeling the entire dataset or a large portion, and they are mainly designed and validated on relatively simple structures like cardiac structures or abdominal organs with assumptions and priors that may not apply to complex structures (e.g., curvilinear ones).
In this paper, we make use of noisy skeletons that differ from scribbles in two ways: (1) skeletons are more label demanding since all branches are supposed to be covered; (2) noisy skeletons are more likely to contain errors or noises, which are inevitable when quickly labeling slender structures. We convert sparse and noisy skeleton annotations to accurate ones via an image synthesis pipeline, thus requiring only one noisy skeleton label. This significantly reduces the annotation cost.

\begin{figure*}[t]
  \vspace{-0.2cm}
  \centering{\includegraphics[width=\textwidth]{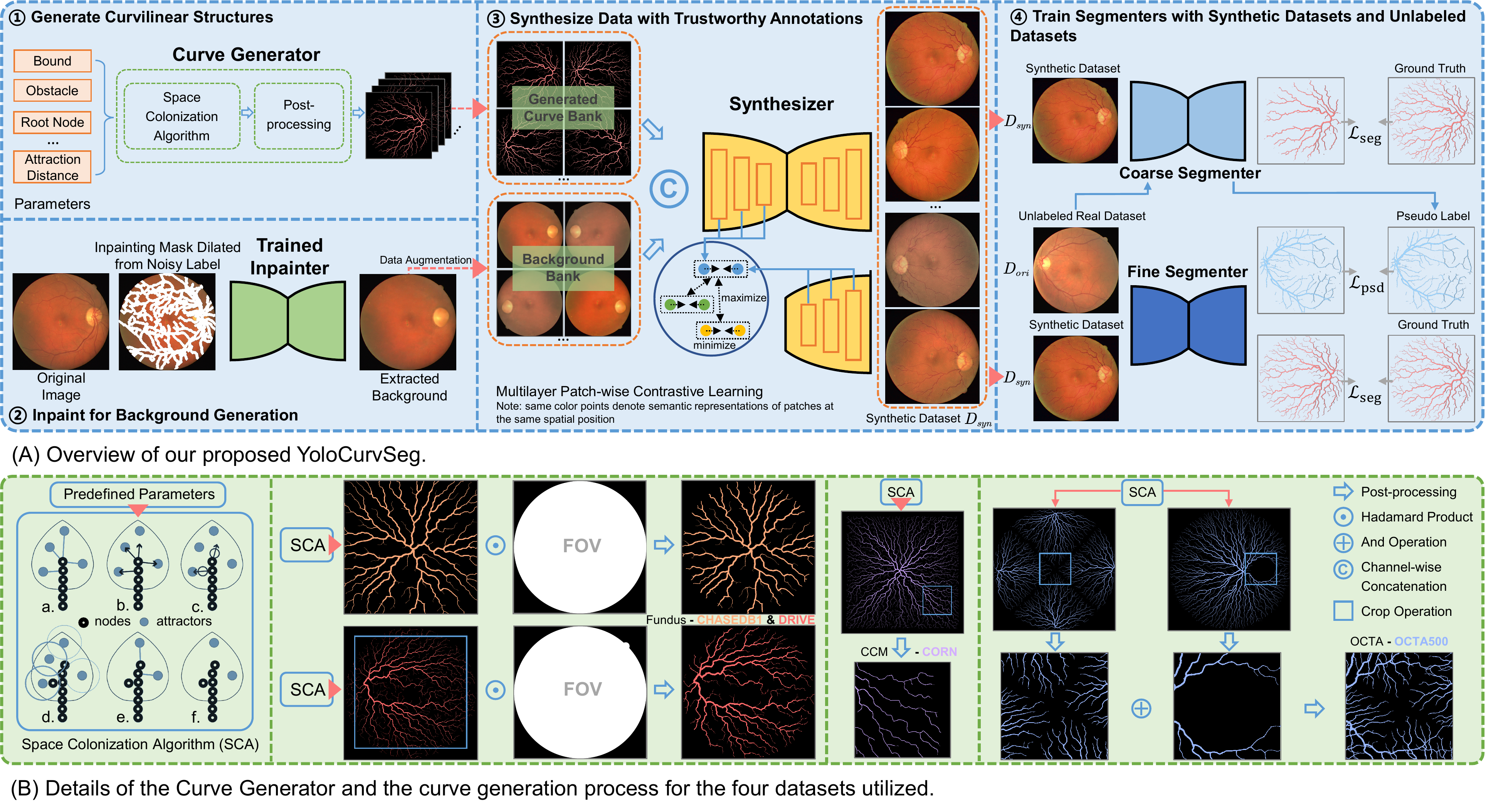}} 
  \vspace{-0.7cm}
  {\caption{
      Top: Overview of our proposed YoloCurvSeg, which comprises four main components: a space colonization algorithm-based curve generator, a background inpainter, a multilayer patch-wise contrastive foreground-background fusion based synthesizer, and a two-stage coarse-to-fine segmenter.
      Bottom: Details of the Curve Generator and the curve generation process for the four datasets utilized.
  }\label{fig:mainfig}}
  \vspace{-0.2cm}
\end{figure*}

\subsection{Medical Image Synthesis}

GAN \citep{goodfellow2020generative} has become the mainstay of medical image synthesis, with common applications in intra-modality augmentation \citep{zhou2020dr}, cross-domain image-to-image translation \citep{peng2022unsupervised}, quality enhancement \citep{cheng2021secret}, missing modality generation \citep{huang2022ds3, huang2022multi}, etc. Below we briefly review previous works on retinal image synthesis, the topic of which is relevant to our work. \cite{costa2017towards} employ a U-Net trained with paired fundus images and vessel masks. It employs a conditional GAN, i.e., Pix2pix \citep{isola2017image} to learn a mapping from vessel masks to the corresponding fundus images. To simplify the framework, they propose an adversarial autoencoder (AAE) for retinal vascularity synthesis and a GAN for generating retinal images \citep{costa2017end}. Similarly, \cite{guibas2017synthetic} present a two-stage approach that consists of a DCGAN for generating vasculature from noise and a cGAN (Pix2pix) to synthesize the corresponding fundus image. Note that cGAN requires paired images and vessel masks for training, which is a strict condition to some extent. These methods require an extra set of vessel annotations to train AAE or DCGAN and may sometimes generate vessels with unrealistic morphology. The generated images also lack diversity. \cite{zhao2018synthesizing} develop Tub-sGAN, which incorporates style transfer into the GAN framework to generate more diverse outputs. In another work, SkrGAN \citep{zhang2019skrgan} is proposed to introduce a sketch prior related constraint to guide the image generation process. Yet, the sketches utilized are extracted by the Sobel edge operator, and cannot be used as segmentation masks. In this paper, we employ a multilayer patch-wise contrastive foreground-background fusion GAN for several considerations. According to previous research, training a GAN to learn a direct mapping from a curvilinear structure mask to the corresponding image is difficult, especially under few-shot conditions \citep{lin2021automated}. Therefore, we provide GAN with extracted real backgrounds, enabling implicit skip-connection that allows the GAN to focus more on mapping the foreground regions. Such a design not only enhances performance but also accelerates convergence. Multilayer patch-wise contrastive learning allows the provided mask and the foreground region of the generated image to be spatially aligned (via unpaired training), which further benefits the subsequent segmenter.

\section{Method}

YoloCurvSeg comprises four main components: (1) a \textit{Curve Generator} that produces binary curve masks that well accommodate the corresponding image modality of interest; (2) an \textit{Inpainter} for extracting backgrounds from labeled samples; (3) a  \textit{Synthesizer} that synthesizes images from the generated curve masks and the image backgrounds; and (4) a two-stage \textit{Segmenter} trained with the synthetic dataset and an unlabeled dataset. 
The overall framework is shown in Fig. \ref{fig:mainfig}.


\subsection{Curvilinear Structure Generation}

Space colonization is a procedural modeling algorithm in computer graphics that simulates the growth of branching networks or tree-like structures \citep{runions2005modeling,runions2007modeling}, including vasculature, leaf venations, root systems, etc. It is employed in YoloCurvSeg for modeling the iterative growth of curvilinear structures with two fundamental elements: attractors and nodes. Its core steps are described in the bottom left panel of Fig. \ref{fig:mainfig}, wherein blue dots denote attractors and black ones denote nodes: a) place a set of attractors randomly or following a predefined pattern, and then associate nodes with nearby attractors (if the distance between a node and an attractor is within an attraction distance $D_a$); b) for each node, calculate its average direction from all attractors affecting it; c) calculate the position of new nodes via normalizing the average direction to a unit vector and scaling it by a predefined \textit{segment length} $L_s$; d) place nodes at the calculated positions and check if any nodes are within an attractor's kill zone; e) prune an attractor if there are nodes staying within its \textit{kill distance} $D_k$; f) repeat steps b)-e) until the maximum number of nodes is reached.
Through observing the pattern of the foreground/curve in a single image or a few images that are accessible, including the curves' starting point, boundary, and degree of curvature, etc., it is relatively straightforward to set the corresponding hyperparameters, such as the root node coordinates $C_r$ (e.g., the starting point of the vessels in fundus lies in the optic disc region), as well as the bounds and obstacles. For $D_a$, $D_k$ and $L_s$, the commonly used values of 5, 30 and 5 can be re-tuned as needed.
{Regarding the attractors, we use a grid placement strategy to control the number of attractors by setting the number of grids in both horizontal and vertical directions. To simplify, we set the same grid number $A_g$ for both directions. Each attractor can be jittered within a certain range $A_j$ to introduce randomness. Attractors located outside the boundary or inside the obstacles are removed.}
Table \ref{table1} summarizes the parameters and post-processing operations we employ for generating the four types of curves and representative examples are demonstrated in the bottom panel of Fig. \ref{fig:mainfig}. Please note that our adopted settings and post-processing operations only represent our empirical choices and are not necessarily the best-performing ones; users can make further adjustments based on their observations and experiences. 
{ In our configuration, multiplying with the field of view (FOV) region is performed to align the curve with the corresponding image background, ensuring that the curve does not exceed the FOV area. Random Crop and Random Flip are employed to further enhance the diversity of the curve, while Erode and Dilate are utilized to fine-tune the thickness of the curve.}
In addition to the curvilinear shape, we also need to simulate the thickness of each branch
\begin{equation}
R^n = R_1^n + R_2^n
\end{equation}
where $R$, $R_1$ and $R_2$ respectively denote the radii of a father branch and its two child branches. $n$ is set to be 3 according to Murray's law \citep{painter2006pulsatile}. The calculation is performed recursively from the branch tips (whose radii are set to be 1) towards the tree base.
Several intuitive demos can be accessed at link\footnote{\href{https://jasonwebb.github.io/2d-space-colonization-experiments/}{https://jasonwebb.github.io/2d-space-colonization-experiments/}}. 
By setting random grid attractors and root nodes via predefined parameters, we construct a bank of curves of the same type but with varied shapes for each dataset of interest, and then employ them to train the \textit{synthesizers} and the \textit{segmenters}.

\subsection{Inpainting for Background Extraction}

Inpainting is the task of reconstructing missing or masked regions in an image. Similar to removing watermarks or extraneous pedestrians from images, we employ an inpainting model here to remove foregrounds (e.g., vessels and nerve fibers) from the images of interest, under the hypothesis that the dilated noisy skeletons can fully cover the foregrounds. In inpainting, common concerns are the network's ability to grasp local and global context and to generalize to a different (especially higher) resolution.

\subsubsection{Architecture}

\begin{figure*}[!t]
  \centering
  \includegraphics[width=0.7\linewidth]{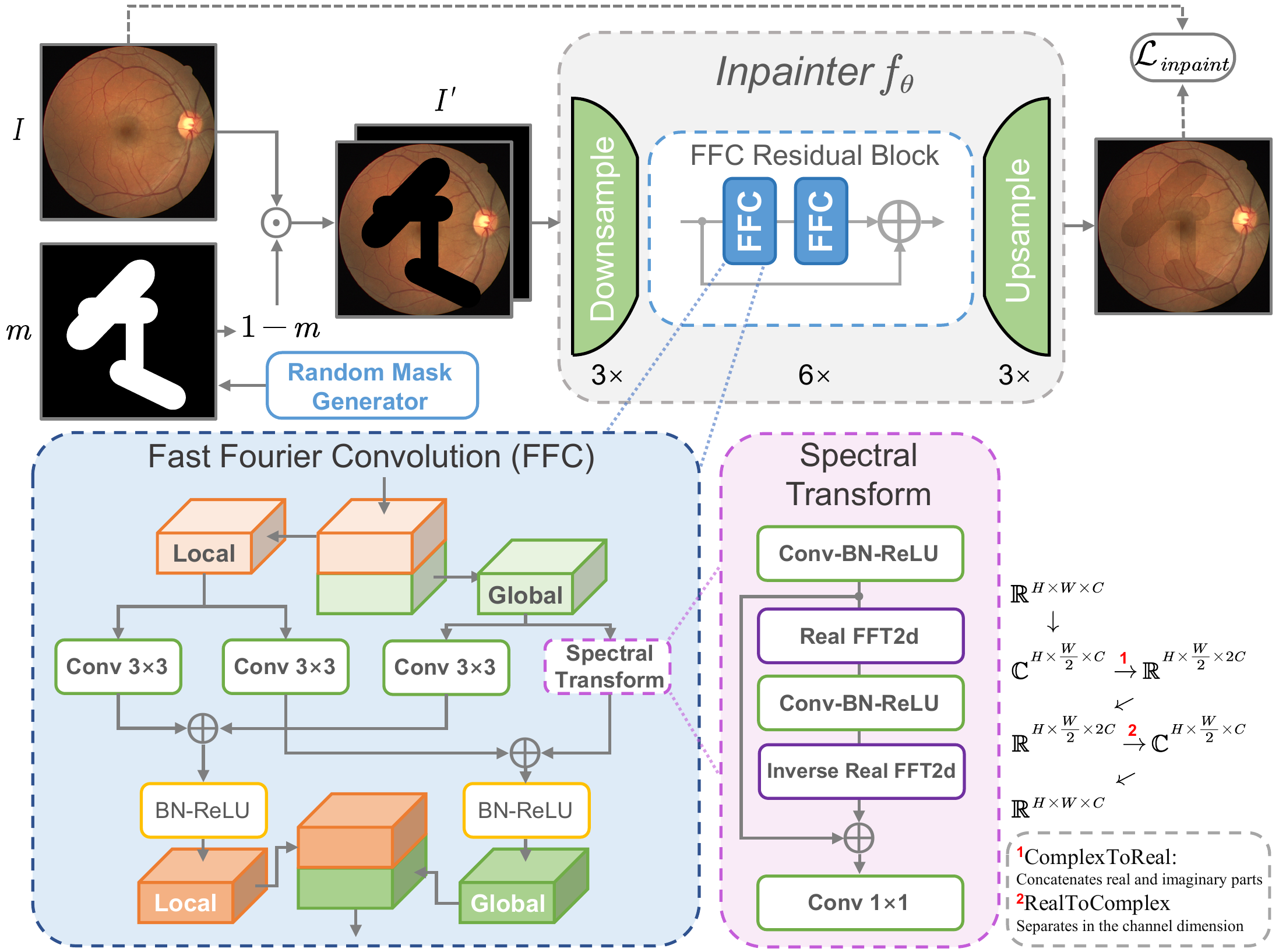}
  \vspace{-0.2cm}
  \caption{The architecture of the \textit{Inpainter}. The input is a four-channel image with the first three channels being the original image and the last channel being the binary mask of the inpainting regions. The output is the inpainted image. The dimensional change of the feature map in FFC is shown in the lower right panel.}
  \label{fig:inpainting}
\end{figure*}

Inspired by \citep{suvorov2022resolution}, we adopt an inpainting network based on the recently proposed \textit{fast Fourier convolutions} (FFCs) \citep{chi2020fast} with image-wide receptive fields, strong generalizability and relatively few parameters. Given a masked image $I \odot (1-m)$, where $I$ and $m$ respectively denote the original image and the binary mask of the inpainting regions, the feed-forward inpainting network $f_\theta(\cdot)$ aims to output an inpainted image $\hat{I}$ = $f_\theta(I^{\prime})$ taking a four-channel input $I^{\prime}=\operatorname{concat}(I\odot (1-m),m)$. FFC builds its basis on channel-wise fast Fourier transform (FFT) and has a receptive field covering the whole image. It splits channels into two parallel branches: a \textit{local} branch uses conventional convolutions and a \textit{global} branch uses \textit{real} FFT to capture global context, as shown in Fig. \ref{fig:inpainting}. \textit{real} FFT is only applicable to real-valued signals, and inverse \textit{real} FFT ensures the output is real-valued. Compared to FFT, \textit{real} FFT uses only half of the spectrum. In FFC, \textit{real} FFT is first applied to the input tensor and a \textit{ComplexToReal} operation is performed by concatenating the real and imaginary parts. Then, it applies convolutions in the frequency domain. Inverse \textit{real} FFT is performed to transform features from the frequency domain to the spatial domain through the \textit{RealToComplex} operation. Finally, the local and global branches are fused. For the upsampling and downsampling of the \textit{Inpainter} and the architecture of the discriminator in adversarial training, we follow the ResNet settings respectively employed in \cite{he2016deep} and \cite{suvorov2022resolution}. The training is performed on [image, randomly synthesized mask] pairs. We adopt the mask generation strategy in \cite{suvorov2022resolution}, containing multiple rectangles with arbitrary aspect ratios and wide polygonal chains.

\begin{table*}[htbp]
  \centering
  {\caption{Parameters for generating different types of curves. $D_a$, $D_k$, $L_s$ and $C_r$ respectively denote the attraction distance, kill distance, segment length and root node coordinates. $r$ and $l$ denote radius and length. $A_g$ and $A_j$ are hyperparameters that respectively control the number of attractor grids and the range of random jitter. $\bigoplus$ and $\bigodot$ represent the union and element-wise multiplication operations. There are two rows for OCTA500 since two components are used to separately generate the horizontal large vessels and the centripetal small vessels.}  \label{table1} }
  \setlength{\tabcolsep}{1mm}
  \resizebox{\textwidth}{!}{      
  \begin{tabular}{cccccccccc}    
      \specialrule{0.13em}{0pt}{0pt}
      Dataset & Bound & Obstacle & $C_r$ & $A_g$& $A_j$& $D_a$ & $D_k$ & $L_s$ & Post-processing \\ \specialrule{0.13em}{0.5pt}{0.8pt}
      
      \multirow{2}{*}{OCTA500} &Circle, $r=450$&-& Midpoints of the Four Sides & \multirow{2}{*}{130}& \multirow{2}{*}{20}& \multirow{2}{*}{5} & \multirow{2}{*}{30} & \multirow{2}{*}{5} & \multirow{2}{*}{Crop, $\bigoplus$}                  \\
                & Circle, $r=450$      &Circle with Center (650,450), $r\in[60,90]$   & $x,y\in[\frac{1}{2}r-150,\frac{1}{2}r+150]$  &       &       &       &                  \\
      \specialrule{0.05em}{0pt}{0pt}
      CORN      &Square, $l=1300$&-& $x,y\in[\frac{1}{2}l-30,\frac{1}{2}l+30]$ &110&15& 5 & 30 & 5 & Erode, Random Crop        \\
      \specialrule{0.05em}{0pt}{0pt}
      DRIVE     & Circle, $r=400$ & Concentric Circle, $r\in[40,60]$ & $x\in[\frac{1}{4}r-30,\frac{1}{4}r+30],y\in[r-50,r+50]$ &85&30& 5 & 30 & 5 & $\bigodot$ FOV, Random Flip                 \\
      \specialrule{0.05em}{0pt}{0pt}
      CHASEDB1     &Square, $l=960$&-& $x,y\in[\frac{1}{2}l-40,\frac{1}{2}l+40]$ &100&12& 3 & 35 & 10 & $\bigodot$ FOV, Dilate                 \\ 
      \specialrule{0.13em}{0pt}{0pt}
  \end{tabular}
  }

\vspace{-0.1cm}
\end{table*}

\vspace{-0.2cm}

\subsubsection{Objective}

Compared with naive supervised losses which may result in blurry predictions, perceptual loss \citep{johnson2016perceptual} evaluates the distance between feature maps of the inpainted image and the original image via a pre-trained network $\phi(\cdot)$. It does not require exact reconstruction and allows for variations in the reconstructed image. Given that inpainting focuses on understanding the global structure, we introduce a perceptual loss of a large receptive field $\mathcal{L}_{HRP}$ through a pre-trained ResNet50 $\phi_{HRF}(\cdot)$ with dilated convolutions
\begin{equation}
    \mathcal{L}_{HRP}(I, \hat{I})=\mathcal{M}([\phi_{HRF}(I)-\phi_{HRF}(\hat{I})]^2)
\end{equation}
where $\mathcal{M}$ is a sequential two-stage mean operator, i.e., obtaining the inter-layer mean of intra-layer means. Additionally, an adversarial loss $\mathcal{L}_{adv}$ is utilized to encourage the inpainted image to be realistic. Specifically, we use a PatchGAN \citep{isola2017image} discriminator $\mathcal{D}_{\xi}(\cdot)$ and label patches that overlap with the mask as \textit{fake} and the others as \textit{real}. The non-saturating adversarial loss is defined as
\begin{equation}
    \begin{gathered}
        \mathcal{L}_D=-E_I[\log D_{\xi}(I)]-E_{I, m}[\log D_{\xi}(\hat{I}) \odot (1-m)] \\
        -E_{I, m}[\log(1-D_{\xi}(\hat{I})) \odot m]
    \end{gathered}
\end{equation}
\vspace{-0.3cm}
\begin{equation}
    \mathcal{L}_G=-E_{I, m}[\log D_{\xi}(\hat{I})]
\end{equation}
\vspace{-0.3cm}
\begin{equation}
    \mathcal{L}_{adv}=\operatorname{sg}_\theta(\mathcal{L}_D)+\operatorname{sg}_{\xi}(\mathcal{L}_G) \rightarrow \min _{\theta, \xi}
\end{equation}
where $\hat{I} = f_\theta(I^\prime)$ is the output of the inpainting network and $\operatorname{sg}_{var}$ represents stop gradient w.r.t. $var$. To further stabilize the training process, we use a gradient penalty $\mathcal{L}_{GP} = E_I\|\nabla D_{\xi}(I)\|_2^2$ \citep{ross2018improving} and a perceptual loss defined on features of the discriminator $\mathcal{L}_{DP}$ \citep{wang2018high}. The final objective of the \textit{Inpainter} is
\begin{equation}
    \mathcal{L}_{inpaint}=\mathcal{L}_{HRP}+\lambda_{adv}\mathcal{L}_{adv}+\lambda_{DP}\mathcal{L}_{DP}+\lambda_{GP}\mathcal{L}_{GP}
\end{equation}      
where $\lambda_{adv}$, $\lambda_{DP}$ and $\lambda_{GP}$ are hyper-parameters balancing the contributions of different losses. $\mathcal{L}_{HRP}$ is responsible for supervised signals and global structure consistency while $\mathcal{L}_{adv}$ and $\mathcal{L}_{DP}$ are responsible for local details and realism.

\subsubsection{Training}

Given that the training of the \textit{Inpainter} does not require annotation and it learns a general ability to recover missing regions through contextual understanding, we initialize the model with pre-trained parameters from the Places-Challenge dataset \citep{zhou2017places} and fine-tune it on images accessible within each training set. 
{The validation set of the Inpainter consists of both accessible training set images and validation set images (each paired with 10 predefined masks obtained using the same generation strategy employed in \cite{suvorov2022resolution}).}
The training is conducted with a batch size of 8 and an Adam optimizer is adopted with a learning rate of $10^{-3}$ for 50 epochs. 
{ Data augmentation consists of random flipping, rotation and color jittering.
For each training image, we first apply the aforementioned data augmentation strategy to offlinely generate 20 augmented images and then employ the same strategy for online augmentation during training.}
We empirically set $\lambda_{adv}=3$, $\lambda_{DP}=10$ and $\lambda_{GP}=10^{-4}$. Once trained, the \textit{Inpainter} is used to remove the foregrounds from the skeleton-labeled samples taking the dilated noisy annotations as the masks. Then we construct a background bank for each dataset by augmenting the extracted backgrounds through random horizontal and vertical flipping as well as rotation (spanning from 0\degree to 90\degree).

\subsection{Patch-wise Contrastive Learning Based Synthesis}

Now we have a curve (foreground) bank $B_{curv} = \{c^{1}, \cdots, c^{N}\}$ and a background bank $B_{bg} = \{b^{1}, \cdots, b^{N}\}$ respectively from the \textit{Curve Generator} and the \textit{Inpainter}, for each given dataset. We construct an intermediate dataset $X_{inter} = \{x^{1}, \cdots, x^{N}\}$ through randomly sampling a curve $c^{i}$ from $B_{curv}$ and a background $b^{i}$ from $B_{bg}$, and then concatenating them to form a temporary sample $x^{i} = \operatorname{concat}(b^{i},c^{i})$.
The problem now turns into an unpaired image-to-image translation task, i.e., designing a synthesizer to learn a mapping from $X_{inter}$ to the corresponding real dataset $Y$. It is desirable that the local context especially the foreground of the synthetic image $\hat{y}^{i}$ is spatially aligned with that of the corresponding intermediate image $x^{i}$ (especially $c^{i}$) as much as possible.

Previously, for unpaired image translation, most existing methods apply GANs with a cycle structure, relying on cycle-consistency to ensure high-level correspondence \citep{zhu2017unpaired}. While effective, the underlying bijective assumption behind cycle-consistency is sometimes too restrictive, which may reduce the diversity of the generated samples. More importantly, the cycle-consistency is not suitable for our task since it does not guarantee any explicit or implicit spatial constraining. In such context, we introduce a multilayer patch-wise contrastive learning based synthesizer to learn a mapping from $X_{inter}$ to $Y$ inspired by \cite{chen2020simple} and \cite{park2020contrastive}, {which is illustrated in the middle panel of Fig. \ref{fig:mainfig} (a)}. It is trained in a generative adversarial manner with an internal contrastive learning pretext task.

The generator (i.e., synthesizer) $G$ is a U-shape network, which firstly down-samples the input image into high-level features via an encoder $E$ with three residual blocks equipped with instance normalization and ReLU activation. As such, each pixel in the high-level feature map represents the embedding feature vector of a patch in the original image. Several layers of interest $E_{l \in L}(x)$ in $E$ are selected to extract multi-scale features of patches and each passes through a two-layer multilayer perceptron (MLP) $H_l$ ($l$ indexes a layer), obtaining a feature stack $\{v_{l \in L}=H_{l \in L}[E_{l \in L}(x)]\}$. Given patchwise features $v_{l}$ and the corresponding pair $\{H_l(E_l(x))^{s_1}, H_l(E_l(G(x)))^{s_2}\}$ with $s_1$ and $s_2$ denoting the spatial locations of the patches of interest, we set $v^+$ to represent a patch at the same location as $v$ and $v^-_n$ to denote the $n^{th}$ among $N$ patches at different locations. The objective of the contrastive learning task is to maintain the local information at the same spatial location. Similar to the noise contrastive estimation loss \citep{oord2018representation}, our objective function can be written as
\begin{equation}
    \mathcal{L}_{c}=-\sum_{l \in L} \log \frac{\exp (v_l \cdot v_l^{+} / \tau)}{\exp (v_l \cdot v_l^{+} / \tau)+\sum_{n=1}^N \exp (v_l \cdot v_{l n}^{-} / \tau)}
\end{equation}
where $\tau$ is a temperature hyper-parameter. Besides, we employ the identity loss, which was first proposed in \cite{zhu2017unpaired} for regularizing the generator $G$. We pass each real sample $y \in Y$ through the encoder $E$ and obtain the patchwise features $v^*$, the negative samples $v^{*-}$ and the positive samples $v_n^{*+}$. The identity loss is formulated as
\begin{equation}
    \mathcal{L}_{id}=-\sum_{l \in L} \log \frac{\exp (v_l^* \cdot v_l^{*+} / \tau)}{\exp (v_l^* \cdot v_l^{*+} / \tau)+\sum_{n=1}^N \exp (v_l^* \cdot v_{l n}^{*-} / \tau)}
\end{equation}

We use the LSGAN loss as our adversarial loss $\mathcal{L}_{adv}$ \citep{mao2017least} to make the synthetic images as realistic as possible. Therefore, with trade-off parameters $\lambda_{adv}$, $\lambda_{c}$ and $\lambda_{id}$, the overall loss of the synthesizer is defined as
\begin{equation}
    \mathcal{L}_{syn} = \lambda_{adv}\mathcal{L}_{adv} + \lambda_{c}\mathcal{L}_{c} + \lambda_{id}\mathcal{L}_{id}
\end{equation}

The training of the synthesizer is conducted employing an Adam optimizer with a learning rate of $10^{-4}$ and a cosine decay strategy, together with a batch size of 1. We utilize images that are accessible within each training set as the corresponding real dataset $Y$ for training the synthesizer. The hyperparameters and model weights are selected based on the Fréchet Inception Distances (FIDs) between the synthesized images and $Y$ \citep{heusel2017gans}. We set $\lambda_{adv}$ as 1, $\lambda_{c}$ as 1, $\lambda_{id}$ as 0.5 and $\tau$ as 0.07. The training process lasts for 300 epochs.

\subsection{Two-stage Coarse-to-Fine Segmentation}

A synthetic dataset $\mathcal{D}_{syn} = \{(\hat{y}^{1},c^1),\cdots,(\hat{y}^{N},c^N)\}$, with $\hat{y}^{i}$ being a synthetic image and $c^i$ being the corresponding curve ground truth, is created by the \textit{Synthesizer}. The weakly-supervised task is then transformed into a fully- or semi- one when making use of solely the synthetic dataset or a combination of an unlabeled dataset $\mathcal{D}_{ori}$ and the synthetic dataset $\mathcal{D}_{syn}$. In this section, we introduce a two-stage coarse-to-fine segmentation pipeline to tackle the task.

A specific segmentation network is first trained on $\mathcal{D}_{syn}$ to obtain a coarse model $S_{coarse}$ with a segmentation loss $\mathcal{L}_{seg}$
\begin{equation}
    \label{eq:segloss}
    \mathcal{L}_{seg} = 0.5\times \mathcal{L}_{ce} + 0.5\times \mathcal{L}_{dice}
\end{equation}
where $\mathcal{L}_{ce}$ and $\mathcal{L}_{dice}$ respectively denote the cross-entropy loss and the Dice loss. We observe and conclude that the performance of $S_{coarse}$ is mainly limited by two issues; one is that the curve generated by the \textit{Curve Generator} still has a certain \textbf{morphological gap} with the foreground of the real image, and the other one is that there is also a slight but inevitable \textbf{intensity gap} between the \textit{Synthesizer}-generated image and the real image. We target at relieving the latter issue by making use of $\mathcal{D}_{ori}$ to further boost the segmentation performance. {We employ predictions on $\mathcal{D}_{ori}$ from $S_{coarse}$ as pseudo-labels, and train a fine model $S_{fine}$ on the combined dataset of $\mathcal{D}_{ori}$ and $\mathcal{D}_{syn}$ through random batch sampling. The final loss function, denoted as $\mathcal{L}_{final}$, is formulated as follows:}
\begin{equation}
    \mathcal{L}_{final} = \mathcal{L}_{seg} + \lambda_{psd}\mathcal{L}_{psd}
\end{equation}
where $\mathcal{L}_{psd}$ denotes the loss on $\mathcal{D}_{ori}$ sharing the same {loss calculation approach as $\mathcal{L}_{seg}$ in Eq. (\ref{eq:segloss}), namely $0.5\times \mathcal{L}_{ce} + 0.5\times \mathcal{L}_{dice}$, and $\lambda_{psd}$ is a trade-off parameter. Please note that each of the two losses is calculated only for the corresponding data samples, i.e. $\mathcal{L}_{seg}$ for $\mathcal{D}_{syn}$ and $\mathcal{L}_{psd}$ for $\mathcal{D}_{ori}$.} We employ the vanilla U-Net with feature channels of 16, 32, 64, 128 and 256 as our $S_{coarse}$'s and $S_{fine}$'s architecture. We use an SGD optimizer (weight decay = $10^{-4}$, momentum = 0.9) for training both $S_{coarse}$ and $S_{fine}$ with a batch size of 12 and an initial learning rate of $10^{-2}$. The total iterations and $\lambda_{psd}$ are respectively set to be 30$k$ and 1.

\vspace{-0.1cm}
\begin{figure*}[!t]
  \centering

  \includegraphics[width=0.85\textwidth]{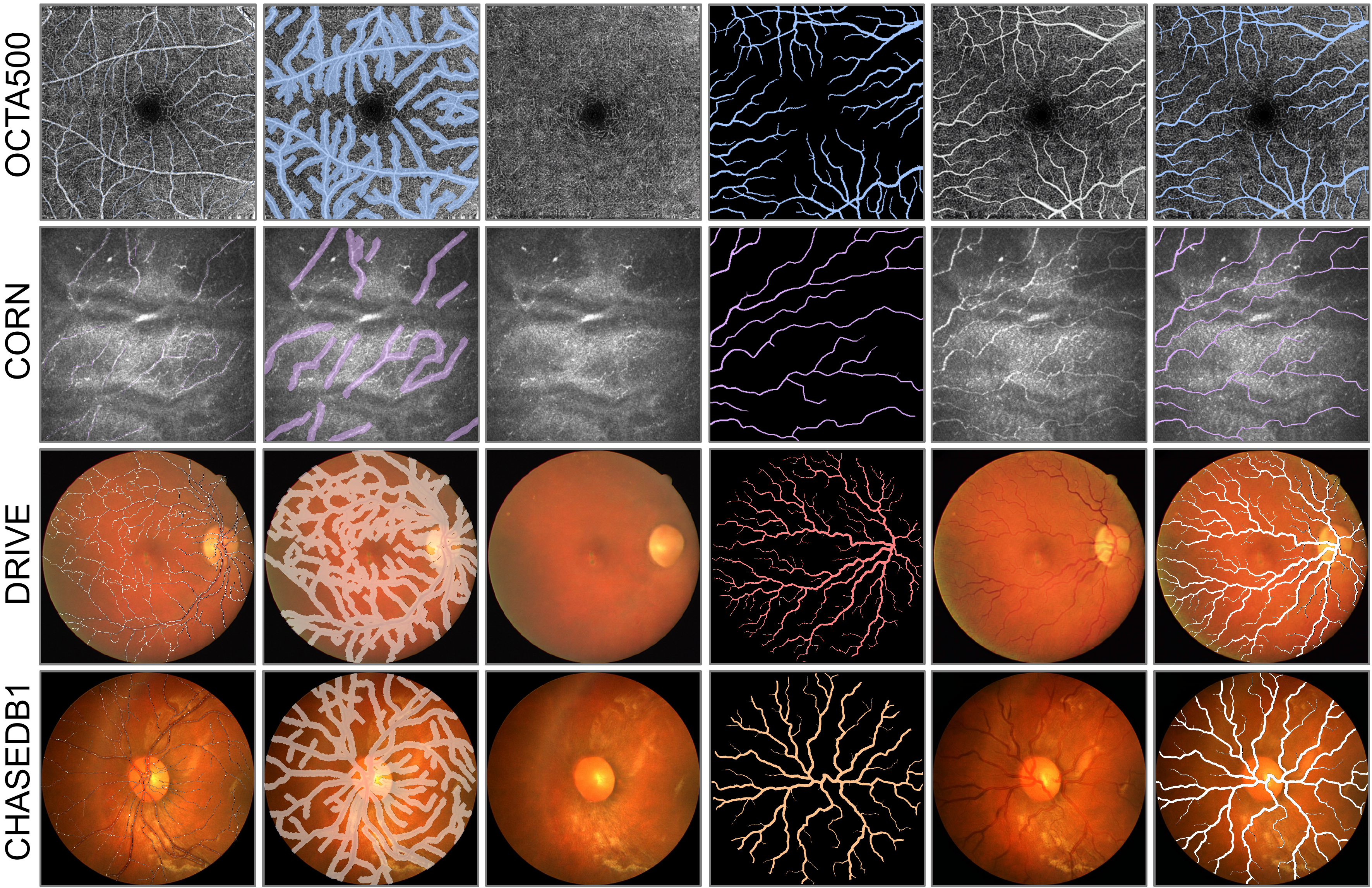}
  \vspace{-0.1cm}
  \caption{Visualization of synthetic data from YoloCurvSeg. From left to right are examples of the noisy skeleton label, the inflated inpainting mask, the extracted background, the generated foreground, the synthesized image and the generated foreground superimposed on the synthesized image.}
  \vspace{-0.2cm}
  \label{fig:synthesis}
\end{figure*}

\section{Experiments}

In this section, we extensively evaluate the effectiveness of our YoloCurvSeg framework on four representative curvilinear structure segmentation datasets.

\subsection{Datasets and Preprocessing}

We comprehensively evaluate YoloCurvSeg on four ophthalmic datasets: OCTA500, CORN, DRIVE and CHASEDB1. OCTA500 is used for retinal microvascular segmentation, and only the subset that contains 300 samples with a $6\times6$ $mm^2$ field of view (FOV) and a $400\times400$ resolution is utilized. We only make use of the $en$-$face$ images generated by maximum projection between the internal limiting membrane layer and the outer plexiform layer. CORN consists of 1578 CCM images for nerve fiber segmentation. It also provides two subsets respectively consisting of 340 low-quality and 288 high-quality images. All CCM images have a resolution of $384\times384$ and an FOV of $400\times400$ $\mu m^2$. Instead of following the dataset's original division, we use 1532 images (samples that overlap with the test set are removed and the validation split ratio is 0.2) for training and validation, and test on 60 relatively accurately labeled samples provided in its subset. DRIVE and CHASEDB1 are used for retinal vessel segmentation and respectively have resolutions of $565\times584$ and $999\times960$. These two fundus datasets are cropped via the provided FOV masks and are respectively resized to $576\times576$ and $960\times960$. For DRIVE, we utilize the original division of 20 training samples and 20 testing samples. For CHASEDB1, we follow the division in \cite{lin2021blu} and \cite{he2022curv}, with the first 20 images serving as the training set and the remaining 8 used for testing. For OCTA500, we respectively utilize 200, 10 and 90 samples as the training, validation and testing sets. Images are first normalized and {online} data augmentation consists of random rotation, flipping and Bézier Curve transformation \citep{zhou2019models}.

\begin{figure*}[t]
  \vspace{-0.2cm}
  \centering

  \includegraphics[width=0.22\linewidth]{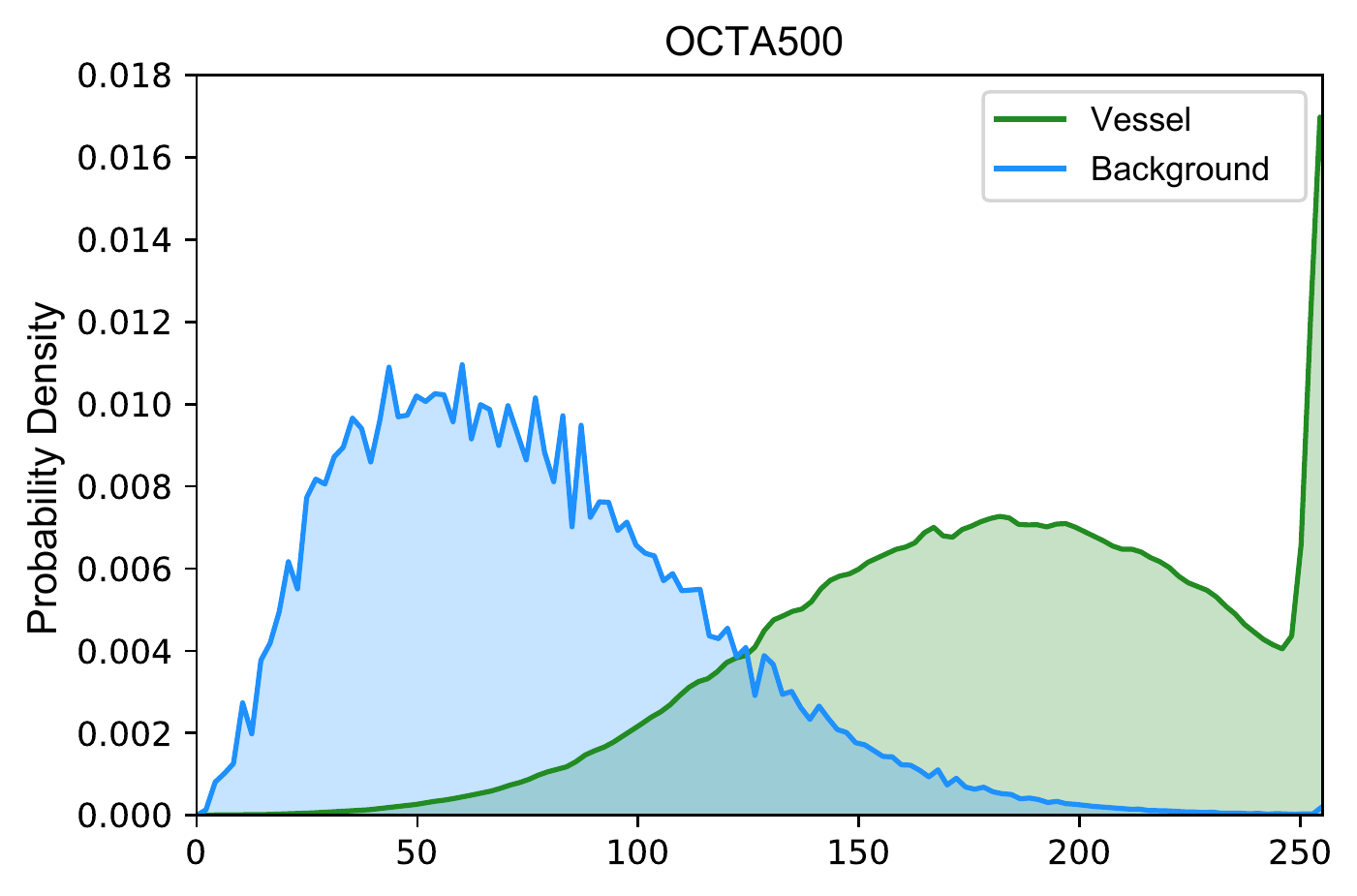}\hspace{+0.1cm}
  \includegraphics[width=0.22\linewidth]{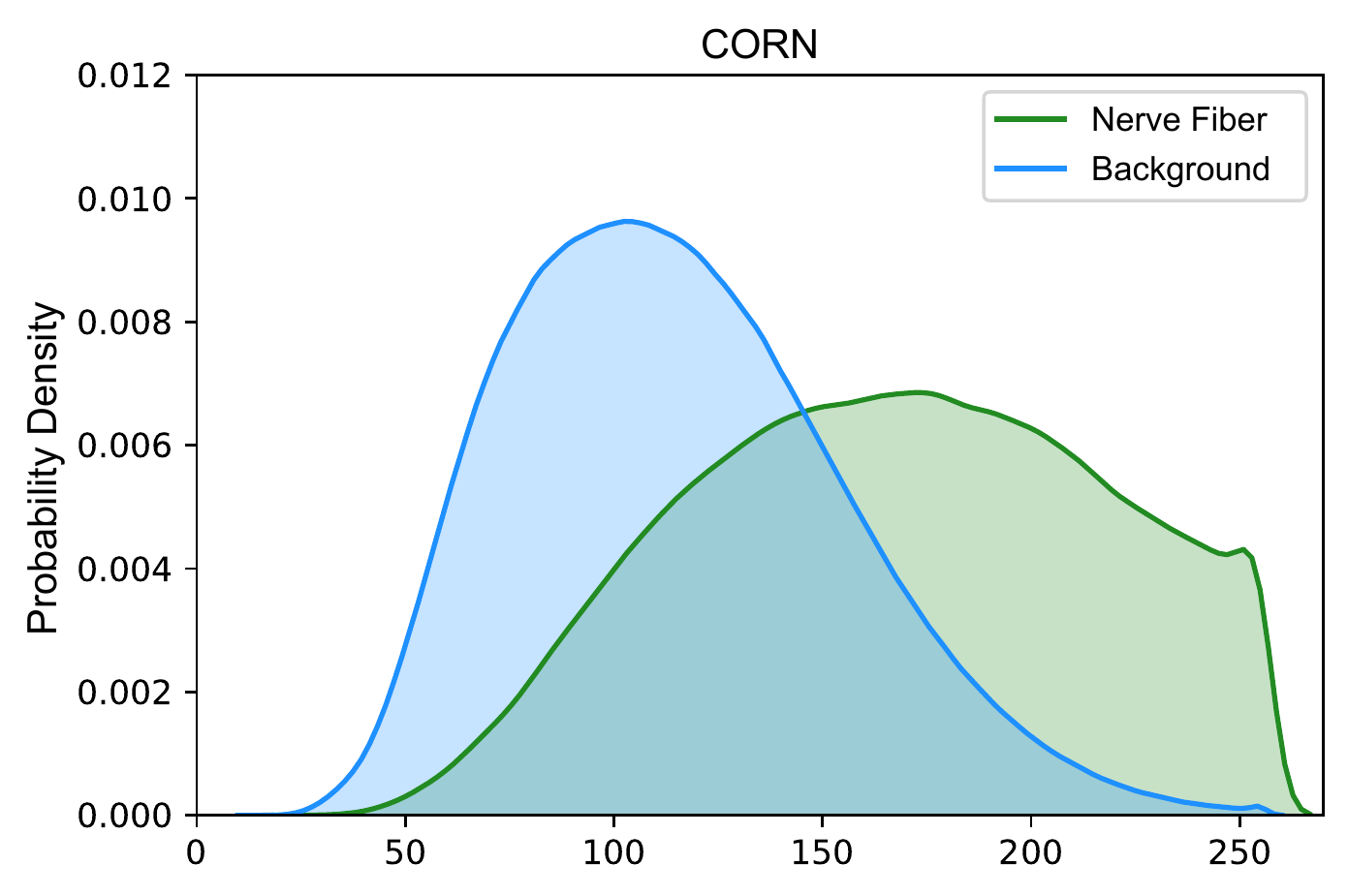}\hspace{+0.1cm}
  \includegraphics[width=0.22\linewidth]{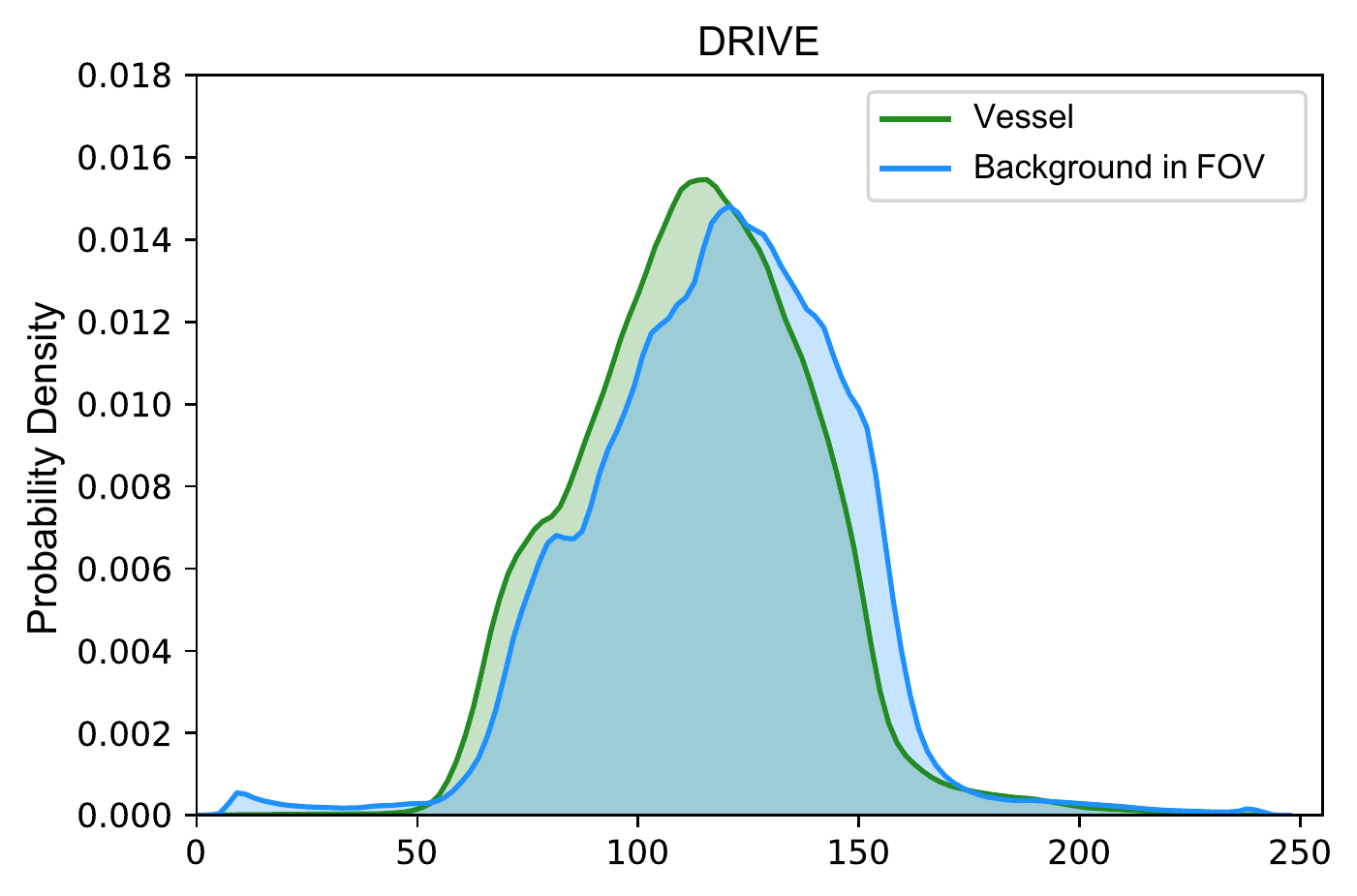}\hspace{+0.1cm}
  \includegraphics[width=0.22\linewidth]{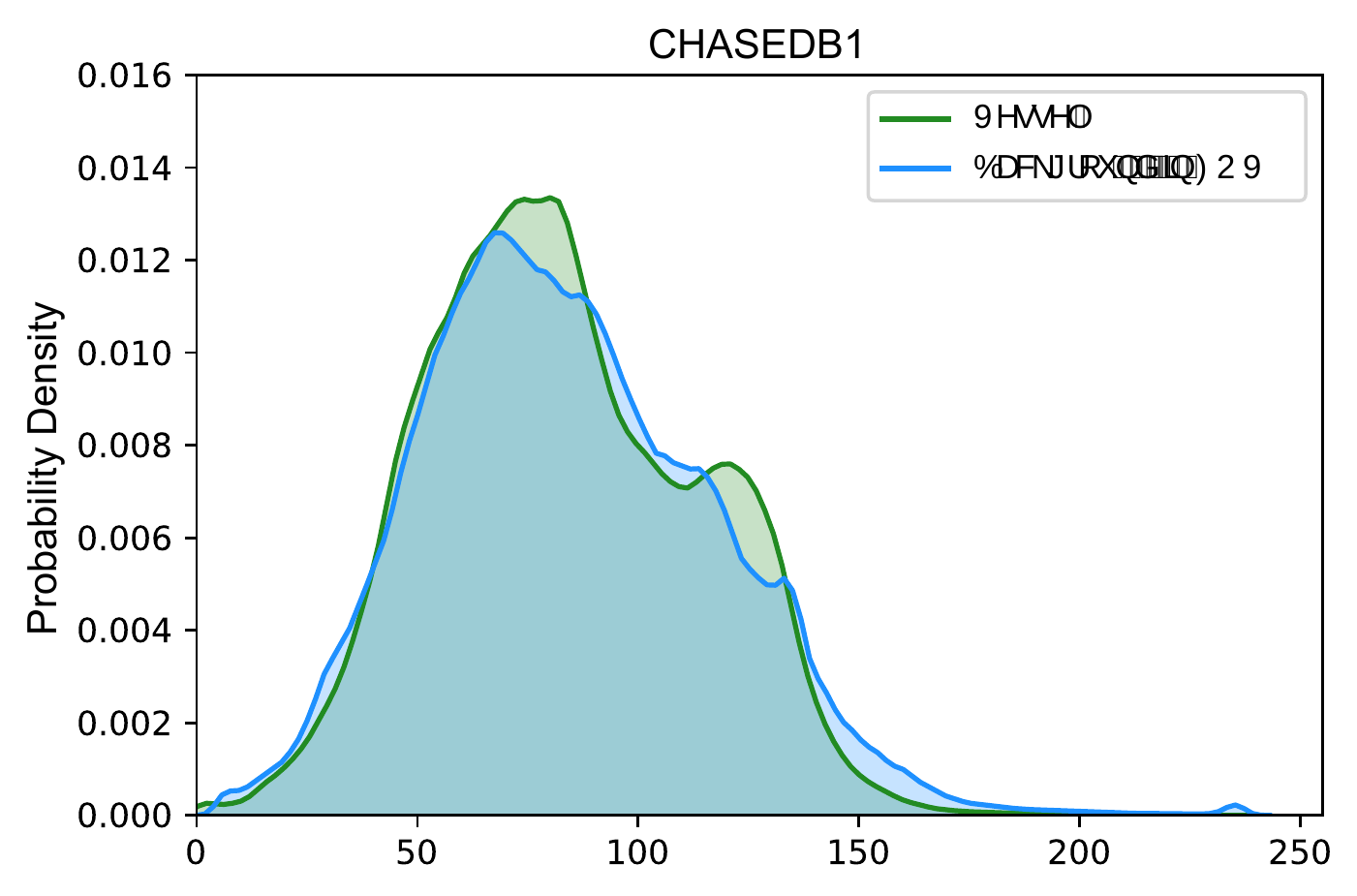}

  \vspace{+0.2cm}

  \centering
  \includegraphics[width=0.22\linewidth]{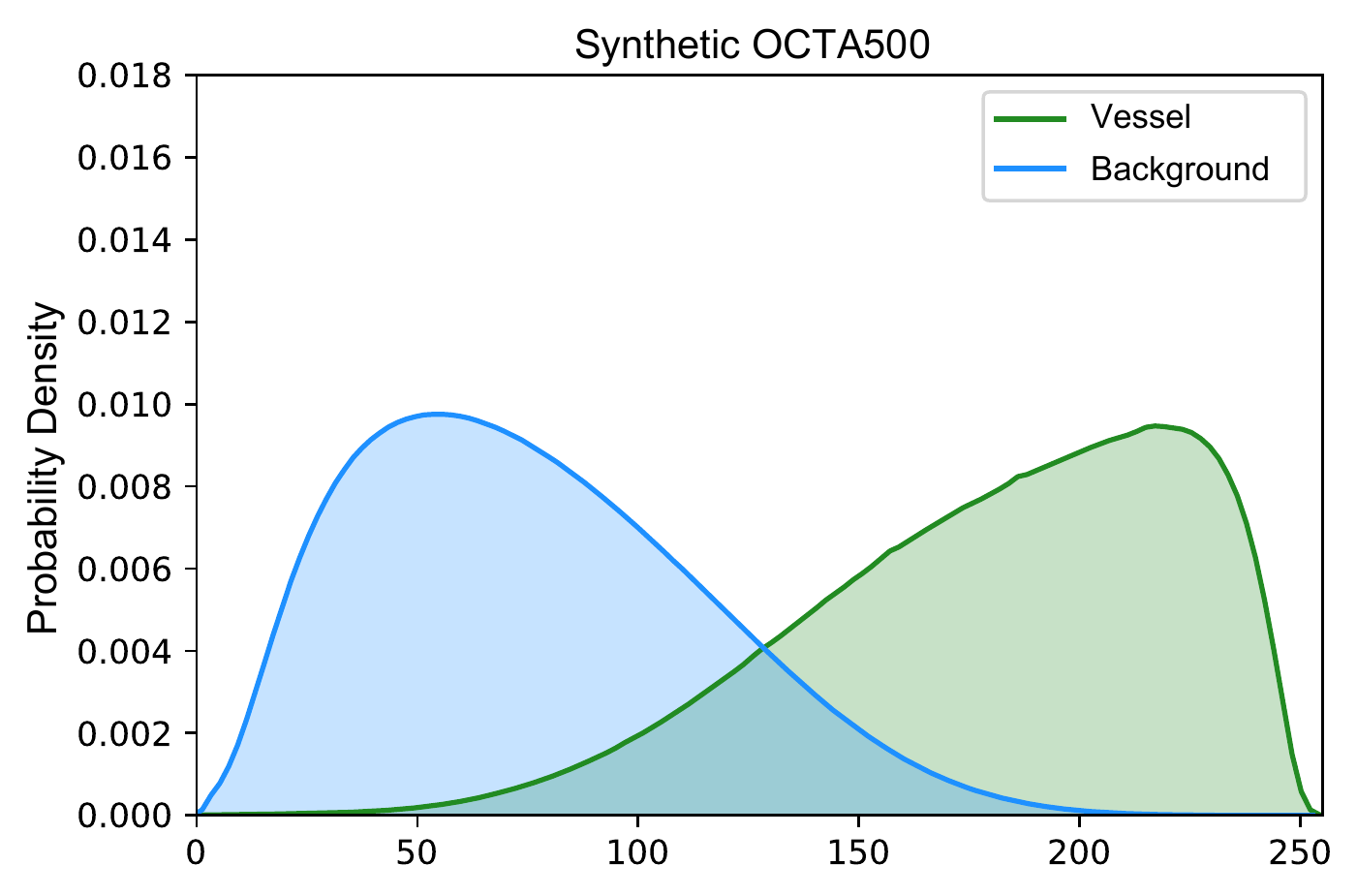}\hspace{+0.1cm}
  \includegraphics[width=0.22\linewidth]{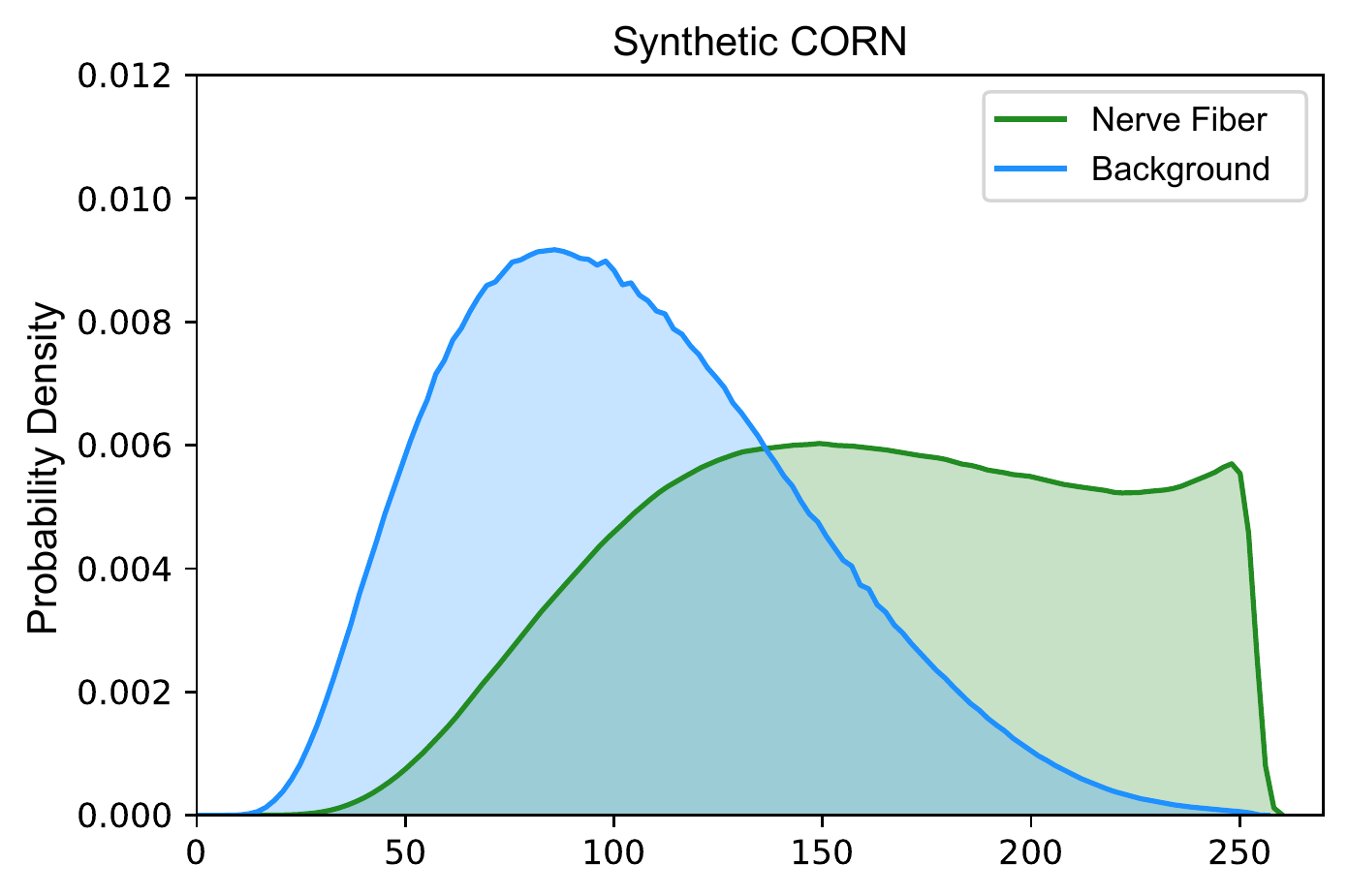}\hspace{+0.1cm}
  \includegraphics[width=0.22\linewidth]{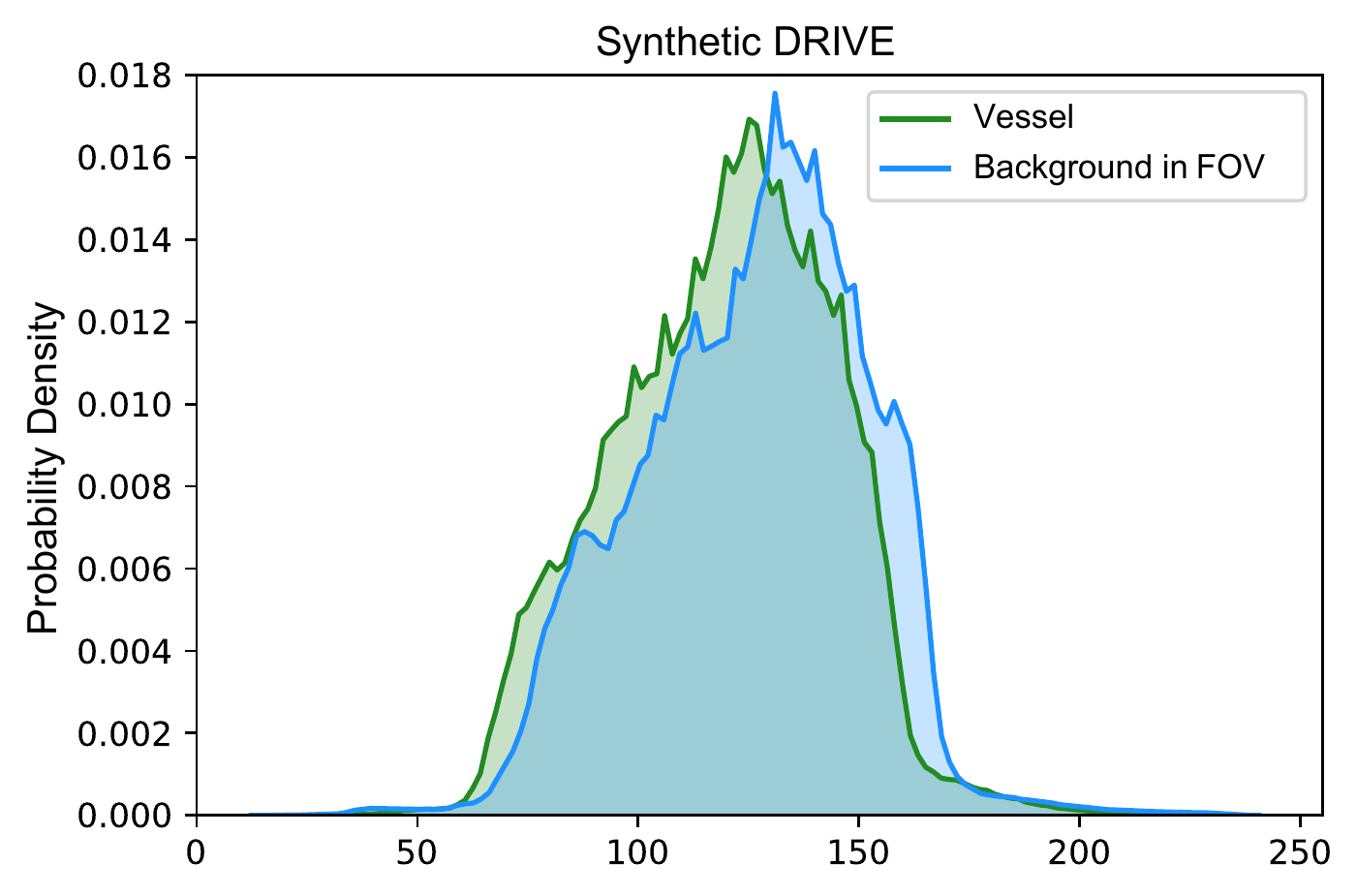}\hspace{+0.1cm}
  \includegraphics[width=0.22\linewidth]{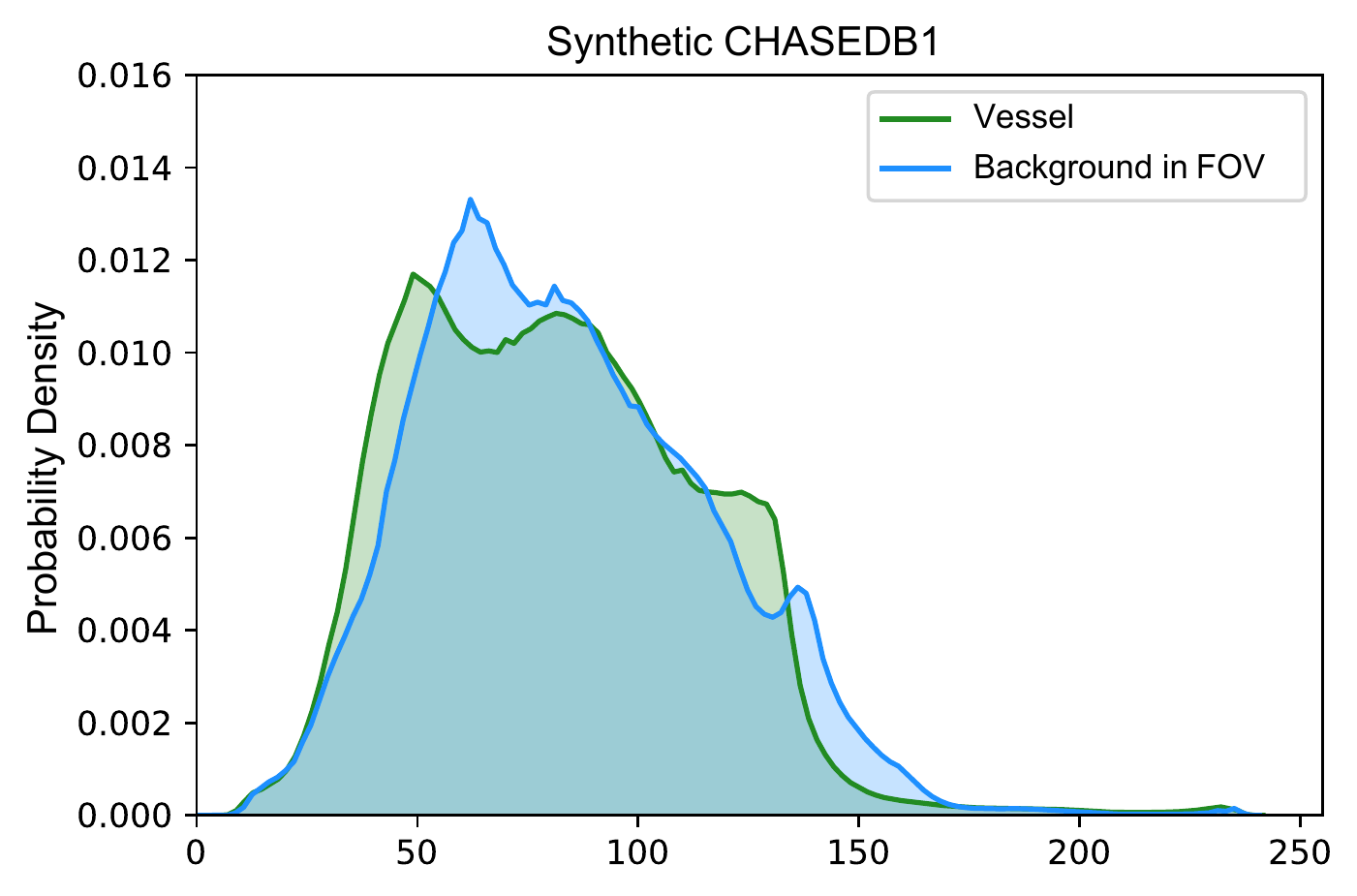}

\vspace{-0.2cm}
\caption{Histograms of the four datasets in terms of the real data (top) and the corresponding synthetic data (bottom).}
\vspace{-0.1cm}

\label{fig:histogram} 
\end{figure*}

\begin{figure*}[htbp]
  \centering

  \includegraphics[width=0.24\linewidth]{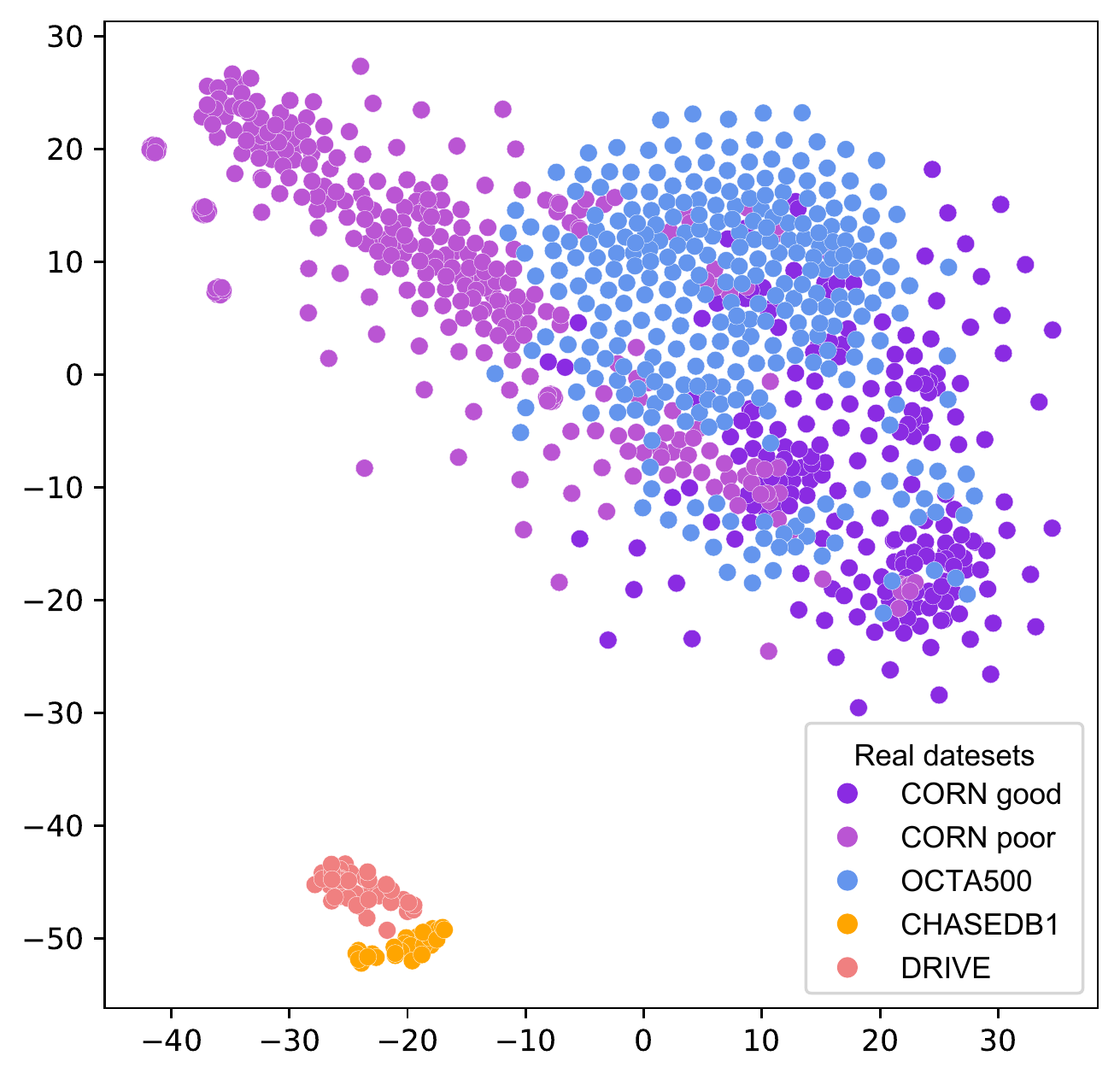}\hspace{+0.1cm}
  \includegraphics[width=0.24\linewidth]{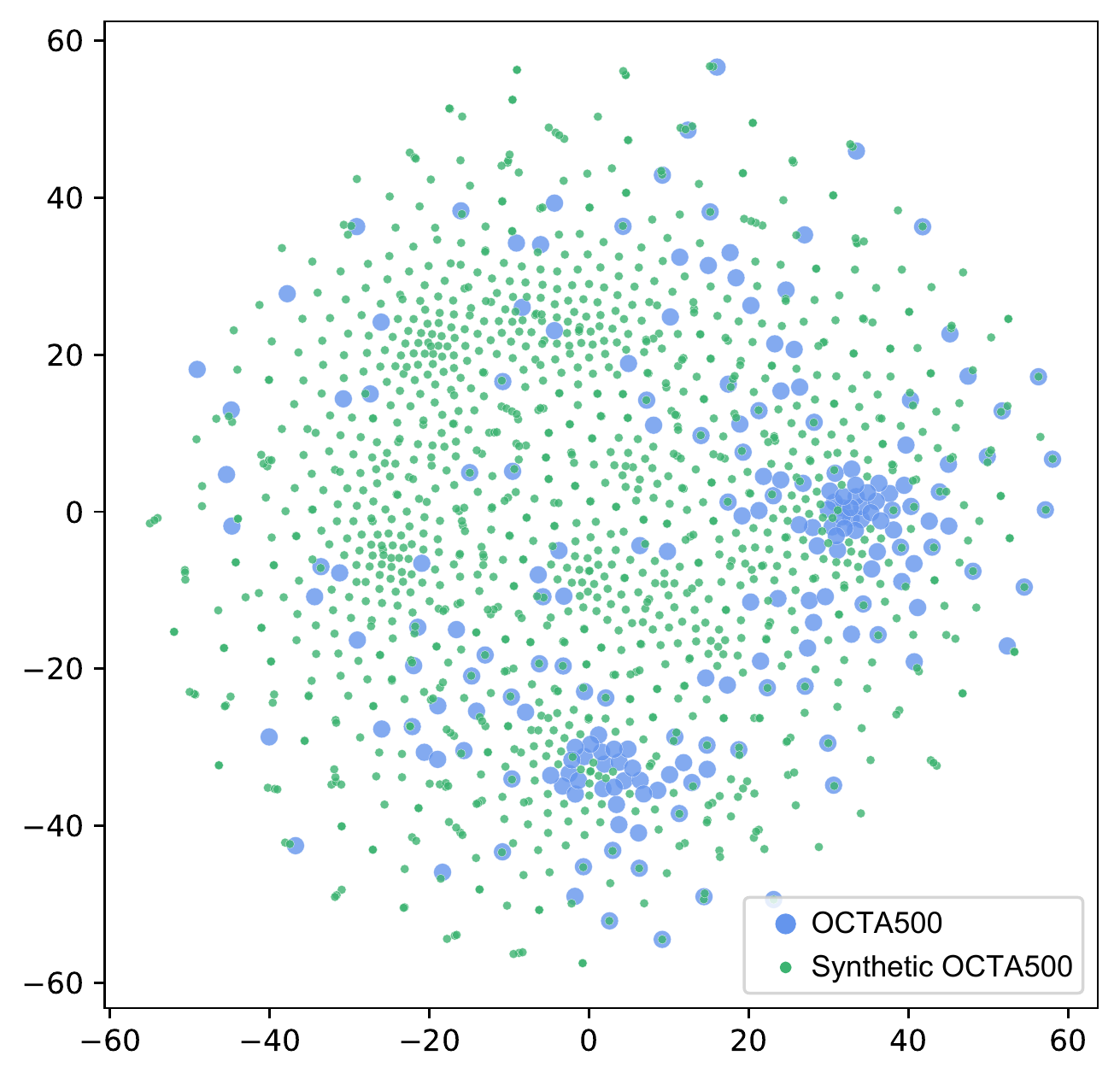}\hspace{+0.1cm}
  \includegraphics[width=0.24\linewidth]{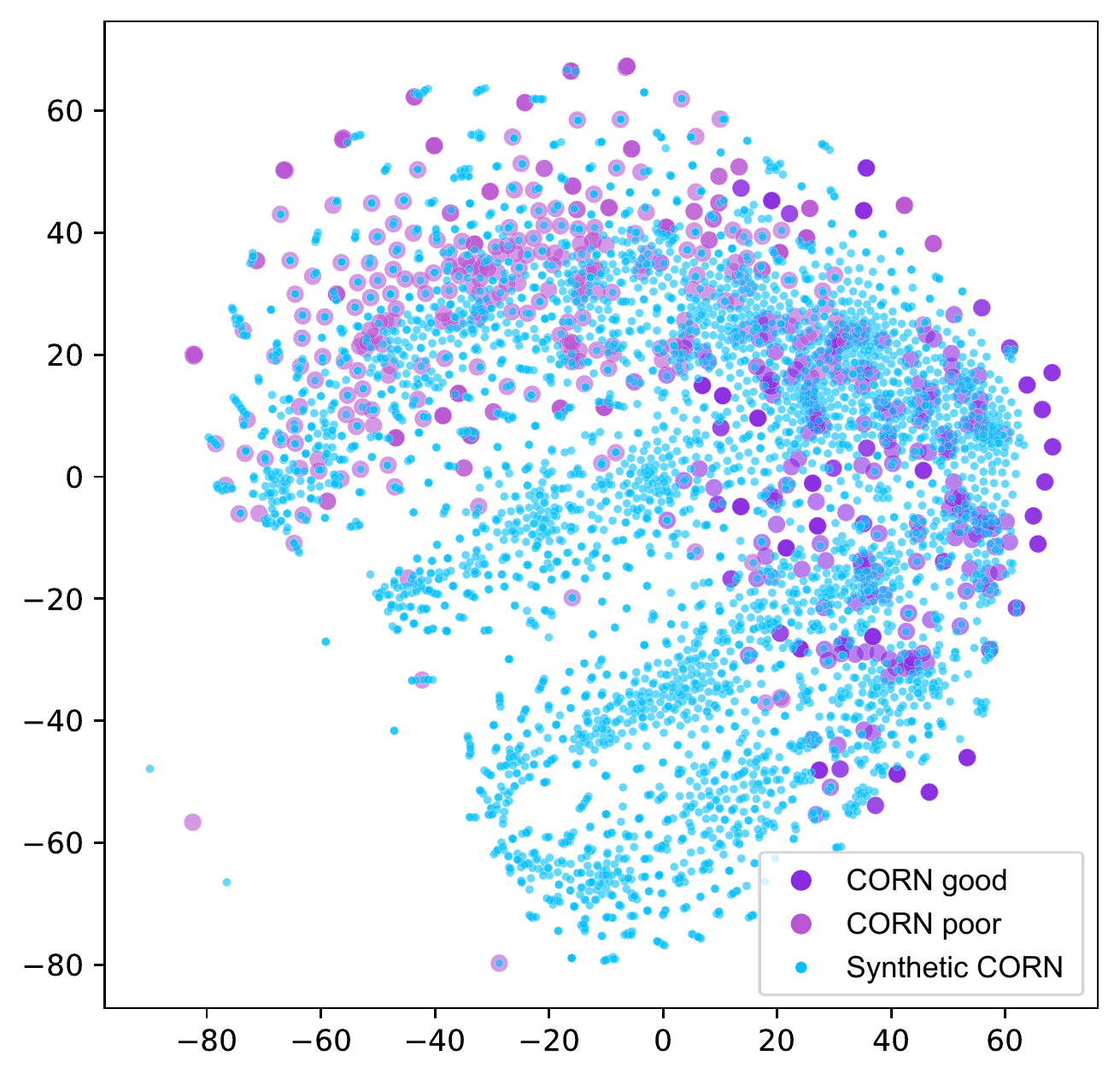}\hspace{+0.1cm}
  \includegraphics[width=0.24\linewidth]{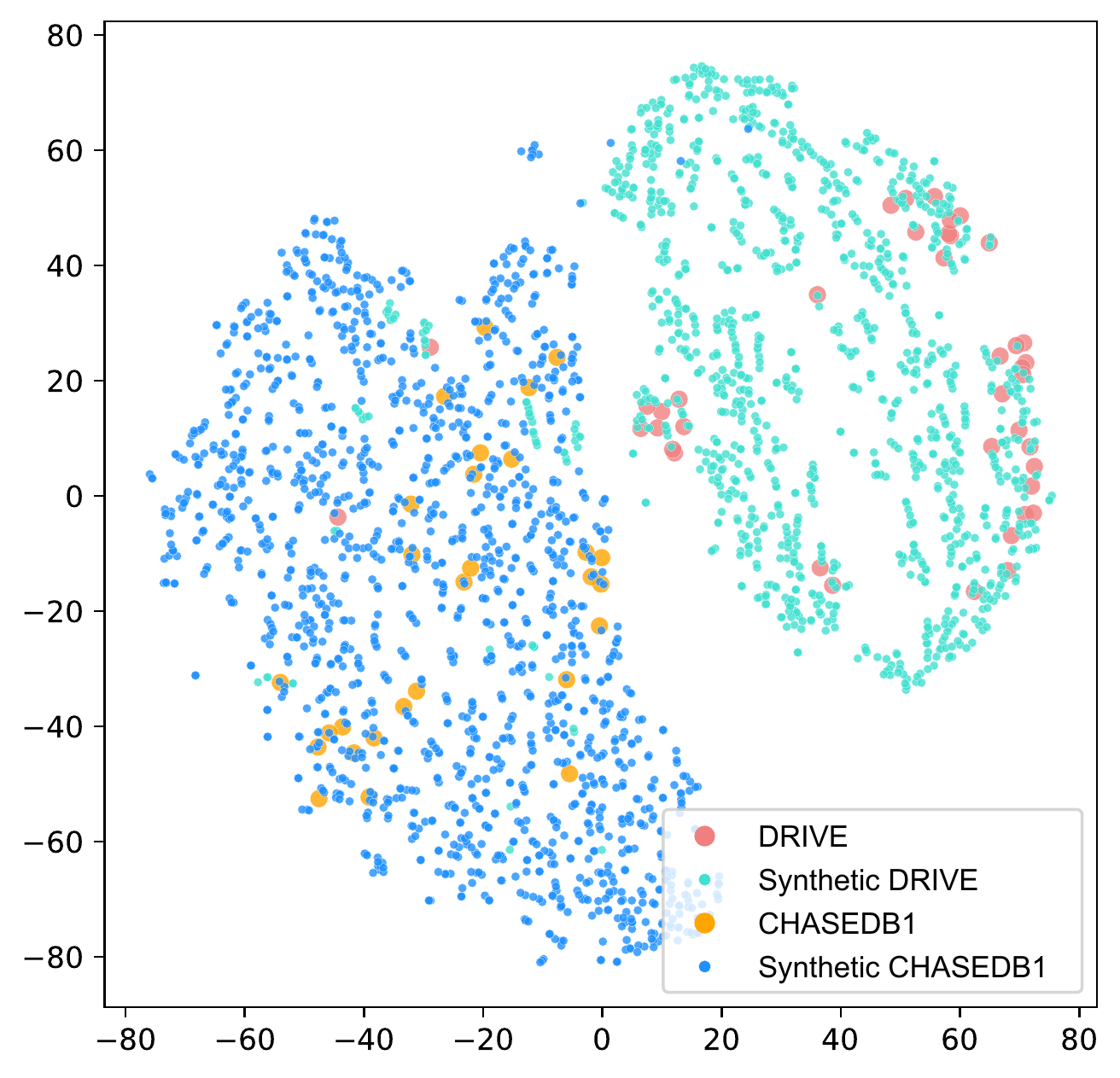}
\vspace{-0.1cm}
\caption{t-SNE visualization of the four real and synthetic datasets. \textbf{CORN good} and \textbf{CORN poor} respectively denote the high-quality and low-quality subsets in CORN.}
\vspace{-0.3cm}
\label{fig:tsne} 
\end{figure*}

\subsection{Implementation Details}

We implement YoloCurvSeg and other compared methods by PyTorch on a workstation equipped with 8 RTX 3090Ti GPUs. In the \textit{Synthesizer}, the indices of the layers selected to calculate $\mathcal{L}_c$ include $\{0,4,8,12,16\}$. For training the \textit{Segmenters} $S_{coarse}$ and $S_{fine}$, the polynomial policy with $power=0.9$ is used to adjust the learning rate online \citep{mishra2019polynomial}. Other hyperparameters, training details and model architecture are already provided in previous sections. It is worth noting that manually-delineated vessel segmentation labels are provided for OCTA500, DRIVE and CHASEDB1. To generate noisy skeleton annotations for those three datasets, we perform the \textit{skeletonize} operation in \textit{scikit-image} \citep{van2014scikit} to obtain the skeletons of the original ground truth masks and then employ \textit{elastic transformation} to simulate jitter noises that may be introduced during fast manual labeling. For CORN, only noisy skeleton labels are provided, and thus are directly used in all our experiments. For this dataset, we dilate each skeleton to a 3-pixel width to serve as the full mask in its fully-supervised learning setting, and the same operation is also applied to the testing set annotations. For sparse labels used in other comparative WSL methods, skeletons of the backgrounds are also generated via skeletonization. The synthesis process in YoloCurvSeg can be online or offline. For better reproducibility and fair comparison, we use the offline version in our experiments, i.e., we first generate the synthetic dataset and then train the \textit{Segmenters}. By randomly combining samples from the pre-generated curve bank and the augmented background bank, we generate 1276, 5005, 1240 and 1604 synthetic samples respectively for OCTA500, CORN, DRIVE and CHASEDB1 if all training samples are labeled. If only one sample is labeled, we respectively generate 100, 100, 60 and 80 synthetic samples.
{ The segmentation models in all our compared methods and YoloCurvSeg are trained for 30000 iterations.}


\subsection{Synthesis Performance}

Before comparing with SOTA WSL methods, we first qualitatively and quantitatively evaluate the synthesis performance of YoloCurvSeg. We visualize representative examples, in terms of the noisy skeleton labels, dilated masks for inpainting, extracted backgrounds, generated curves and synthesized images, in Fig. \ref{fig:synthesis}. It can be observed from the last column that the generated curves well match the synthetic images.  
\begin{table}[htbp]
  \vspace{-0.2cm}
  \centering
  \caption{FID scores between various synthetic components and the real ones. A: synthetic mask vs. real mask, B: synthetic mask vs. real image, C: synthetic background vs. real image, D: synthetic image vs. real image, E: real training image vs. real test image.}
  \small
  \label{tab:FID}
  \renewcommand\arraystretch{1.25}
  \tiny
  \resizebox{\linewidth}{!}{
  \begin{tabular}{llllll}
      \specialrule{0.1em}{0pt}{0pt}
      Dataset  & A & B & C & D & E \\ 
      \specialrule{0.05em}{0pt}{0pt}
      OCTA500 & 44.93 & 202.82 & 195.66 & 40.88 & 25.89 \\
      CORN & 97.17 & 313.04 & 174.42 & \textbf{38.26} & 60.91 \\
      DRIVE & 157.53 & 298.93 & 143.69 & 67.91 & 51.49 \\
      CHASEDB1 & 165.74 & 367.72 & 136.29 & \textbf{56.11} & 75.96 \\ \specialrule{0.1em}{0pt}{0pt}
  \end{tabular}
  }
  \vspace{-0.2cm}
\end{table}
We also compare the intensity distributions of the synthetic datasets with the real ones in Fig. \ref{fig:histogram}, exhibiting high intensity similarities between the synthetic and real images in terms of both background and foreground. 
From the t-SNE \citep{van2008visualizing} visualization in Fig. \ref{fig:tsne}, the synthetic datasets are generally in line and well mixed with the real ones. In most cases, the synthetic data are even more uniformly and widely distributed, having a similar effect as data augmentation. The FIDs between the synthetic components and the corresponding real ones are tabulated in Table \ref{tab:FID} to measure the difficulty of learning a mapping between the two distributions. Column A shows the morphological gap between the synthetic and real curvilinear structures. Columns B and C illustrate there is great difficulty in translating directly to the image distribution from foreground masks, while the background image with the foreground removed is less distant from the original distribution. Since YoloCurvSeg incorporates background information, which implicitly acts as skip connections, it reduces the difficulty of image translation and thus can synthesize images close to the real ones even under few-shot settings. Our method achieves competitive FID scores on all four datasets, two of which are even smaller than those between the real training and test sets, as shown in column E of Table \ref{tab:FID}. Successful alignments between synthetic curves and regions of interest in synthetic images and high similarities between synthetic images and real ones are well established, both of which are crucial factors for YoloCurvSeg to achieve powerful segmentation performance in the following experiments.

\begin{table*}[!t]
  \centering

  {\caption{Comparison with existing WSL methods on the OCTA500 and CORN datasets. The best results are highlighted in bold, and the second-best results are underlined. \textbf{FS} denotes fully-supervised learning.}\label{tab:comparison1}}

  \renewcommand\arraystretch{1.08}
  \setlength{\tabcolsep}{1.2mm}
  \renewcommand\theadgape{\Gape[1.8mm][0mm]}
  \resizebox{0.91\linewidth}{!}{
  \begin{tabular}{clcccccccc}
      \specialrule{0.1em}{0pt}{0pt}
\multirow{2}{*}{\begin{tabular}[c]{@{}c@{}}Sample \thead{}\\ Size\end{tabular}} & \multirow{2}{*}{Method} & \multicolumn{4}{c}{OCTA500} & \multicolumn{4}{c}{CORN} \\ 
\Xcline{3-10}{0.05em}
&                    \thead{}& DSC$\uparrow$           & ASSD$\downarrow$          & SE$\uparrow$         & SP$\uparrow$         & DSC$\uparrow$          & ASSD$\downarrow$         & SE$\uparrow$         & SP$\uparrow$           \\ \specialrule{0.1em}{0pt}{0pt}
\multirow{13}{*}{All}
& pCE \citep{tang2018normalized}                     & 67.43$\pm$4.51              &  2.44$\pm$0.66           &  64.35$\pm$6.21           &   97.42$\pm$0.70         &   42.47$\pm$5.66           &   8.01$\pm$4.90         &  66.43$\pm$14.37             &   92.97$\pm$2.76           \\
& RW \citep{grady2006random}         &  61.61$\pm$4.10             &  2.84$\pm$0.64           &  63.69$\pm$5.80           &   95.76$\pm$0.62         &  32.44$\pm$6.07            &  8.40$\pm$4.46          &   51.54$\pm$12.90            &   91.62$\pm$3.94           \\
& USTM \citep{liu2022weakly}                    & 71.64$\pm$3.18              & \underline{1.74$\pm$0.37}            &   74.06$\pm$3.93          &  96.78$\pm$0.53          &  40.70$\pm$5.49            &   8.93$\pm$5.74         & 70.57$\pm$14.92              &  91.49$\pm$3.04            \\
& S2L \citep{lee2020scribble2label}                     &  69.75$\pm$5.07             &  2.37$\pm$0.73           &   74.56$\pm$6.96          &    96.08$\pm$1.63        &  41.47$\pm$7.22            &  8.81$\pm$4.22          &  71.14$\pm$9.55             &  91.40$\pm$3.63            \\
& MLoss \citep{kim2019mumford}                   & \underline{72.26$\pm$3.57}              &  1.96$\pm$0.48           &   73.40$\pm$4.71          &  97.08$\pm$0.48          &  45.34$\pm$3.59            &   6.53$\pm$3.39         &  79.01$\pm$13.15            &  91.55$\pm$3.26            \\
& EM \citep{grandvalet2004semi}                      &  71.10$\pm$3.47             &  1.86$\pm$0.44           &  \underline{77.87$\pm$4.30}           &  95.94$\pm$0.63          &  41.19$\pm$3.02            &   7.18$\pm$2.76         &  \textbf{82.55$\pm$9.85}             &   89.23$\pm$3.76           \\
& Dense CRF \citep{tang2018regularized}                &   71.95$\pm$3.35            & 1.99$\pm$0.44           & 72.33$\pm$4.40            &  97.18$\pm$0.42          &  42.56$\pm$7.24            &   8.55$\pm$5.55         &  63.62$\pm$15.49             &   93.46$\pm$2.88           \\
& Gated CRF \citep{obukhov2019gated}                 &  70.31$\pm$3.90             &  2.73$\pm$0.61           &   66.05$\pm$4.78          & \underline{97.87$\pm$0.36}           &  45.24$\pm$7.32            & 7.64$\pm$4.52           &  58.86$\pm$14.39             & \underline{95.12$\pm$2.22}            \\
& AC \citep{chen2019learning}                      &     70.15$\pm$3.01          & 1.89$\pm$0.33            & 72.09$\pm$4.08          & 96.70$\pm$0.58         &   41.49$\pm$8.46           &  7.91$\pm$4.39          &  61.02$\pm$13.12             &    93.51$\pm$2.60          \\
& DBDM \citep{luo2022scribble}                    & 71.52$\pm$3.67              & 1.95$\pm$0.51            & 74.59$\pm$4.87          & 96.65$\pm$0.69            & 43.78$\pm$9.30             &   7.66$\pm$4.07         &  59.67$\pm$11.69              &   94.49$\pm$2.10           \\
& Tree Energy \citep{liang2022tree}             &  72.21$\pm$3.88             &    2.12$\pm$0.54         &  72.05$\pm$5.15          & 97.31$\pm$0.49           &  \underline{51.22$\pm$4.33}            & \underline{5.29$\pm$2.31}           & \underline{82.47$\pm$10.42}              &  93.04$\pm$2.62            \\
& Ours (coarse)                    & \textbf{{85.24$\pm$2.51}}              &  \textbf{{0.87$\pm$0.23}}            & \textbf{84.21$\pm$4.40}          & \textbf{98.70$\pm$0.35}           & \textbf{{72.15$\pm$5.61}}             &  \textbf{{3.53$\pm$2.20}}           & 68.99$\pm$8.16              &  \textbf{{98.96$\pm$0.42}}            \\
& FS\cellcolor{gray!30}                   & 88.93$\pm$2.23\cellcolor{gray!30}        & 0.60$\pm$0.20\cellcolor{gray!30}       & 90.63$\pm$2.38\cellcolor{gray!30}            &  98.71$\pm$0.41\cellcolor{gray!30}        &  67.19$\pm$6.67\cellcolor{gray!30}            &  3.83$\pm$2.17\cellcolor{gray!30}          & 67.60$\pm$9.80\cellcolor{gray!30}              &   98.44$\pm$0.53\cellcolor{gray!30}           \\ \specialrule{0.05em}{0pt}{0pt}
\multirow{13}{*}{One}
& pCE \citep{tang2018normalized}                     &  68.19$\pm$4.55             &  3.07$\pm$0.81           &  65.70$\pm$6.94           &  97.36$\pm$0.87          & 37.74$\pm$10.80             &  12.14$\pm$6.24          &   34.79$\pm$13.75            &  97.76$\pm$1.59            \\
& RW \citep{grady2006random}                      &   62.13$\pm$4.17            & 3.69$\pm$0.93            &  64.78$\pm$7.15           &  95.68$\pm$1.18          &  23.54$\pm$7.71            &  17.10$\pm$9.33          & 27.32$\pm$13.86              &  94.86$\pm$4.72            \\
& USTM \citep{liu2022weakly}                    &    67.13$\pm$3.83           &  3.07$\pm$0.71           &  66.53$\pm$5.61           &   96.89$\pm$0.53         &  42.55$\pm$7.80            &  8.86$\pm$5.70          &  \underline{58.13$\pm$12.30}             &  94.24$\pm$2.79            \\
& S2L \citep{lee2020scribble2label}                     &   66.70$\pm$4.00            &  2.67$\pm$0.67           &  \underline{75.88$\pm$5.97}           &  94.87$\pm$1.25          &  39.65$\pm$9.09            &  9.71$\pm$4.74          &  39.07$\pm$12.58             &  97.39$\pm$1.63            \\
& MLoss \citep{kim2019mumford}                   &  67.19$\pm$3.73             &  2.90$\pm$0.69           &  68.02$\pm$5.38           &   96.62$\pm$0.52         &  39.45$\pm$7.79            & 10.35$\pm$5.52           &  51.03$\pm$14.22             &  94.53$\pm$3.66            \\
& EM \citep{grandvalet2004semi}                      &  67.70$\pm$4.57             &  2.86$\pm$0.78           &  70.19$\pm$7.03           &   96.32$\pm$1.28         &  38.40$\pm$9.58            &   10.95$\pm$5.71         & 39.97$\pm$14.49              &  96.80$\pm$2.14            \\
& Dense CRF \citep{tang2018regularized}                &   68.77$\pm$4.25            &   2.88$\pm$0.77          &  66.85$\pm$6.40           &  97.30$\pm$0.76          &  35.49$\pm$8.72            &   10.99$\pm$5.67         &  43.62$\pm$16.32             &   95.14$\pm$3.48           \\
& Gated CRF \citep{obukhov2019gated}                 & \underline{69.16$\pm$3.92}              &  \underline{2.49$\pm$0.65}           &  69.75$\pm$5.73           &  96.86$\pm$0.74          &   38.23$\pm$11.23           & 11.42$\pm$5.93           &  34.54$\pm$13.72             &  \underline{97.99$\pm$1.26}            \\
& AC \citep{chen2019learning}                      &   65.94$\pm$4.93             & 3.82$\pm$0.95            & 59.26$\pm$7.05            & \underline{98.03$\pm$0.60}           &   37.51$\pm$9.15           &  11.21$\pm$6.03          &  40.88$\pm$14.97             &   96.67$\pm$2.41          \\
& DBDM \citep{luo2022scribble}                    &  66.35$\pm$4.23             &  2.80$\pm$0.71           & 74.17$\pm$6.46            & 95.09$\pm$1.32            &  41.09$\pm$7.30            &   8.26$\pm$4.71         &  53.37$\pm$15.23             &  94.71$\pm$2.99            \\
& Tree Energy \citep{liang2022tree}             &  64.98$\pm$4.11             &  3.30$\pm$0.79           & 67.83$\pm$6.60            & 95.97$\pm$0.95            & \underline{53.52$\pm$9.03}       &  \underline{7.18$\pm$5.07}          &  57.05$\pm$10.34            &   97.25$\pm$1.54           \\
& Ours (coarse)                    & \textbf{{85.47$\pm$2.39}}              & \textbf{{0.81$\pm$0.21}}            & \textbf{86.27$\pm$3.83}            & \textbf{98.47$\pm$0.42}           &  \textbf{{71.45$\pm$6.48}}            & \textbf{{4.01$\pm$2.59}}          & \textbf{69.25$\pm$7.86}            &  \textbf{98.85$\pm$0.50}           \\
& FS\cellcolor{gray!30}                      &84.60$\pm$3.35\cellcolor{gray!30}               &0.98$\pm$0.35\cellcolor{gray!30}             &88.89$\pm$2.91\cellcolor{gray!30}             &97.93$\pm$0.81\cellcolor{gray!30}            &43.23$\pm$9.01\cellcolor{gray!30}              &10.09$\pm$4.40\cellcolor{gray!30}            &55.99$\pm$11.05\cellcolor{gray!30}               &95.03$\pm$2.25\cellcolor{gray!30}              \\ \specialrule{0.1em}{0pt}{0pt}
  \end{tabular}
  }
  \vspace{-0.2cm}

\end{table*}

\begin{table*}[!t]
  \centering

  {\caption{Comparison with existing WSL methods on the DRIVE and CHASEDB1 datasets. The best results are highlighted in bold, and the second-best results are underlined. \textbf{FS} denotes fully-supervised learning.}\label{tab:comparison2}}

  \renewcommand\arraystretch{1.08}
  \setlength{\tabcolsep}{1.2mm}
  \renewcommand\theadgape{\Gape[1.8mm][0mm]}
  \resizebox{0.91\linewidth}{!}{
  \begin{tabular}{clcccccccc}
      \specialrule{0.1em}{0pt}{0pt}
\multirow{2}{*}{\begin{tabular}[c]{@{}c@{}}Sample \thead{}\\ Size\end{tabular}} & \multirow{2}{*}{Method} &  \multicolumn{4}{c}{DRIVE} & \multicolumn{4}{c}{CHASEDB1} \\ 
\Xcline{3-10}{0.05em}
&                    \thead{}& DSC$\uparrow$           & ASSD$\downarrow$          & SE$\uparrow$         & SP$\uparrow$         & DSC$\uparrow$          & ASSD$\downarrow$         & SE$\uparrow$         & SP$\uparrow$           \\ \specialrule{0.1em}{0pt}{0pt}
\multirow{13}{*}{All}
& pCE \citep{tang2018normalized}                               &   62.59$\pm$5.82           &   4.60$\pm$0.87  & \underline{76.24$\pm$9.50} &  91.92$\pm$4.46      &  55.20$\pm$3.13             &   10.09$\pm$2.02   & 71.90$\pm$5.55 &  92.80$\pm$2.37      \\
& RW \citep{grady2006random}                 &  42.13$\pm$3.05            &  9.86$\pm$1.50    & 63.58$\pm$9.99&  83.86$\pm$4.45    &   39.34$\pm$3.11            &   14.23$\pm$1.44     & 70.88$\pm$3.00& 85.00$\pm$2.25     \\
& USTM \citep{liu2022weakly}                    &  62.52$\pm$6.26            &   4.71$\pm$0.81     & 73.89$\pm$9.63 & 92.41$\pm$4.96    & 56.74$\pm$2.48              &  7.43$\pm$1.18    & 74.27$\pm$8.29 & 92.87$\pm$2.64        \\
& S2L \citep{lee2020scribble2label}                                &  56.81$\pm$3.45            &  5.71$\pm$0.98    & 67.63$\pm$9.00& 91.85$\pm$2.20      &  61.75$\pm$3.29             &  7.28$\pm$1.11       & 75.32$\pm$3.58&  94.45$\pm$1.73    \\
& MLoss \citep{kim2019mumford}                           &  58.62$\pm$4.88            &   4.65$\pm$0.58   & \textbf{80.09$\pm$7.49} &  88.91$\pm$4.15     &  63.82$\pm$2.85             &  5.74$\pm$1.42     & \textbf{87.21$\pm$6.78} &  93.02$\pm$2.22      \\
& EM \citep{grandvalet2004semi}                              &  61.64$\pm$3.99            &   5.19$\pm$1.05   & 61.00$\pm$8.37&   95.79$\pm$1.20    &  54.37$\pm$2.94             &   9.23$\pm$2.26     & 82.04$\pm$6.20&  90.30$\pm$2.93     \\
& Dense CRF \citep{tang2018regularized}                &  58.88$\pm$4.29            &   4.23$\pm$0.57   & 74.65$\pm$7.51&  90.68$\pm$3.33     &  56.48$\pm$5.28             &   9.23$\pm$2.12      & 74.05$\pm$7.56& 92.47$\pm$3.70     \\
& Gated CRF \citep{obukhov2019gated}                    &  47.81$\pm$3.69             & 8.46$\pm$1.31    & 49.09$\pm$9.07 & 93.59$\pm$2.37      &  51.54$\pm$3.16             &  9.69$\pm$1.42     & \underline{82.20$\pm$4.73} &  88.93$\pm$3.12      \\
& AC \citep{chen2019learning}                       &   62.41$\pm$6.40           &  5.39$\pm$1.36   & 68.02$\pm$10.55&   94.07$\pm$3.91     &  54.89$\pm$2.00             &    9.48$\pm$3.08      & 62.69$\pm$9.47& 94.69$\pm$2.36    \\
& DBDM \citep{luo2022scribble}                    & \underline{64.18$\pm$3.88}             &   \underline{3.36$\pm$0.66}   & 69.50$\pm$7.01&  \underline{94.45$\pm$2.28}     &  53.87$\pm$4.24              &   10.25$\pm$3.25    & 73.47$\pm$9.27 &  91.82$\pm$3.37      \\
& Tree Energy \citep{liang2022tree}                       &  60.21$\pm$4.39            & 6.50$\pm$1.70    & 66.37$\pm$10.91 &  93.81$\pm$2.48      & \underline{73.07$\pm$1.75}              &  \underline{3.71$\pm$0.95}      & 77.32$\pm$2.43& \underline{97.30$\pm$0.40}      \\
& Ours (coarse)                     & \textbf{{78.30$\pm$2.04}}             &  \textbf{{1.59$\pm$0.30}}  & 74.37$\pm$4.87&  \textbf{98.23$\pm$0.36}      & \textbf{{78.32$\pm$2.63}}              &  \textbf{{3.22$\pm$0.78}}       & 82.16$\pm$4.93&  \textbf{97.83$\pm$0.28}   \\
& FS\cellcolor{gray!30}                   &  81.84$\pm$1.47\cellcolor{gray!30}            &  1.19$\pm$0.20\cellcolor{gray!30}    & 81.99$\pm$4.75\cellcolor{gray!30}&  97.90$\pm$0.46\cellcolor{gray!30}    & 81.29$\pm$1.59\cellcolor{gray!30}              &   2.45$\pm$0.47\cellcolor{gray!30}   & 85.21$\pm$2.64\cellcolor{gray!30} & 98.07$\pm$0.36\cellcolor{gray!30}       \\ \specialrule{0.05em}{0pt}{0pt}
\multirow{13}{*}{One}
& pCE \citep{tang2018normalized}                               & 55.84$\pm$5.85             &  \underline{4.59$\pm$0.59}       & \textbf{80.92$\pm$6.82} &   86.83$\pm$5.65&   52.62$\pm$1.92            &  9.07$\pm$2.10      & 63.52$\pm$9.76& 93.57$\pm$3.01     \\
& RW \citep{grady2006random}                                &  39.43$\pm$3.30            &  13.62$\pm$2.78    & 62.16$\pm$10.37&  82.06$\pm$5.29     & 40.41$\pm$1.71              &  13.38$\pm$1.31   & 72.93$\pm$5.27& 84.98$\pm$3.37         \\
& USTM \citep{liu2022weakly}                    &  56.07$\pm$4.03            &  6.57$\pm$1.20    & 64.90$\pm$9.24& \underline{92.31$\pm$2.07}      &  50.73$\pm$6.40             &  15.65$\pm$5.77    & 63.00$\pm$9.00& 92.95$\pm$2.95        \\
& S2L \citep{lee2020scribble2label}                     &  54.96$\pm$4.41            &  5.74$\pm$1.10  & 70.39$\pm$10.07&   89.94$\pm$3.77      &  54.66$\pm$2.44             &  10.56$\pm$4.63    & 71.51$\pm$13.18&  92.68$\pm$3.39       \\
& MLoss \citep{kim2019mumford}                          &  \underline{55.94$\pm$5.05}            & 5.78$\pm$1.14    & 69.05$\pm$10.91&   90.80$\pm$4.43     &  53.05$\pm$1.10             &  9.46$\pm$2.82   & 66.46$\pm$7.48& 93.15$\pm$2.48         \\
& EM \citep{grandvalet2004semi}                      &  47.89$\pm$6.25            &   7.06$\pm$1.95  & 52.86$\pm$10.15& 91.67$\pm$5.41       & 52.26$\pm$2.53              &  9.48$\pm$1.95    & \underline{75.75$\pm$7.50} & 90.63$\pm$3.48        \\
& Dense CRF \citep{tang2018regularized}               &  53.14$\pm$5.33            &   5.67$\pm$1.08   & 68.66$\pm$11.33& 89.33$\pm$5.09      &  52.15$\pm$1.17             &   12.91$\pm$5.86     & 63.37$\pm$10.22& 93.56$\pm$2.77      \\
& Gated CRF \citep{obukhov2019gated}                 &   40.80$\pm$3.37           & 9.78$\pm$1.18     & 69.68$\pm$6.57& 79.80$\pm$4.02      &  47.54$\pm$2.95             &  11.30$\pm$2.35    & 72.24$\pm$8.39&  89.31$\pm$3.62       \\
& AC \citep{chen2019learning}                      &   53.73$\pm$7.89           &  6.31$\pm$1.70    & 70.21$\pm$13.52& 88.57$\pm$8.60      &  52.54$\pm$2.03             &   10.91$\pm$4.41    & 70.71$\pm$12.23&  91.92$\pm$3.68      \\
& DBDM \citep{luo2022scribble}                    &  52.99$\pm$4.31            &   7.03$\pm$1.37   & 66.62$\pm$9.21&  90.03$\pm$3.19     &  53.29$\pm$2.39             &  11.42$\pm$2.31    & 59.26$\pm$5.12& 94.98$\pm$1.27        \\
& Tree Energy \citep{liang2022tree}             & 51.09$\pm$4.49             &  6.19$\pm$1.13  & 72.76$\pm$9.01 & 86.78$\pm$4.27        &  \underline{67.20$\pm$2.13}             &   \underline{5.69$\pm$1.80}     & 67.25$\pm$4.82& \underline{97.40$\pm$0.66}      \\
& Ours (coarse)                    &  \textbf{{77.99$\pm$1.97}}            & \textbf{{1.71$\pm$0.34}}    & \underline{73.74$\pm$4.55} &  \textbf{98.25$\pm$0.33}     & \textbf{{78.20$\pm$2.50}}              &  \textbf{{2.86$\pm$0.68}}     & \textbf{83.83$\pm$4.14} & \textbf{97.60$\pm$0.40}      \\
& FS\cellcolor{gray!30}                      &76.89$\pm$2.52\cellcolor{gray!30}              &2.05$\pm$0.49\cellcolor{gray!30}    & 78.51$\pm$6.06\cellcolor{gray!30} & 97.04$\pm$1.27\cellcolor{gray!30}       &66.00$\pm$8.11\cellcolor{gray!30}               &6.51$\pm$2.35\cellcolor{gray!30}    & 71.39$\pm$3.98\cellcolor{gray!30} &  96.16$\pm$2.61\cellcolor{gray!30}         \\ \specialrule{0.1em}{0pt}{0pt}
  \end{tabular}
  }
  \vspace{-0.2cm}

\end{table*}

\subsection{Comparison with SOTA}

Since noisy skeletons can be considered as sparse annotations with a certain degree of noise or simply noisy labels, we compare YoloCurvSeg with two categories of methods: (1) WSL methods and (2) noisy label learning (NLL) methods. The Dice similarity coefficient (DSC[\%]) and the average symmetric surface distance (ASSD[pixel]) are used as the evaluation metrics. Sensitivity (SE) and specificity (SP) are also employed for more comprehensive evaluations of the differences among various methods. All benchmarked WSL methods, NLL methods, and fully-supervised (FS) methods utilize the same segmentation network architecture as that adopted in $S_{coarse}$ and $S_{fine}$ (i.e., vanilla U-Net) for a fair comparison purpose. {Selecting the vanilla U-Net architecture is further motivated by its versatility and widespread application in medical image segmentation tasks \citep{ronneberger2015u,isensee2021nnu,antonelli2022medical}.}

\begin{figure*}[!h]

  \centering{\includegraphics[width=1\textwidth]{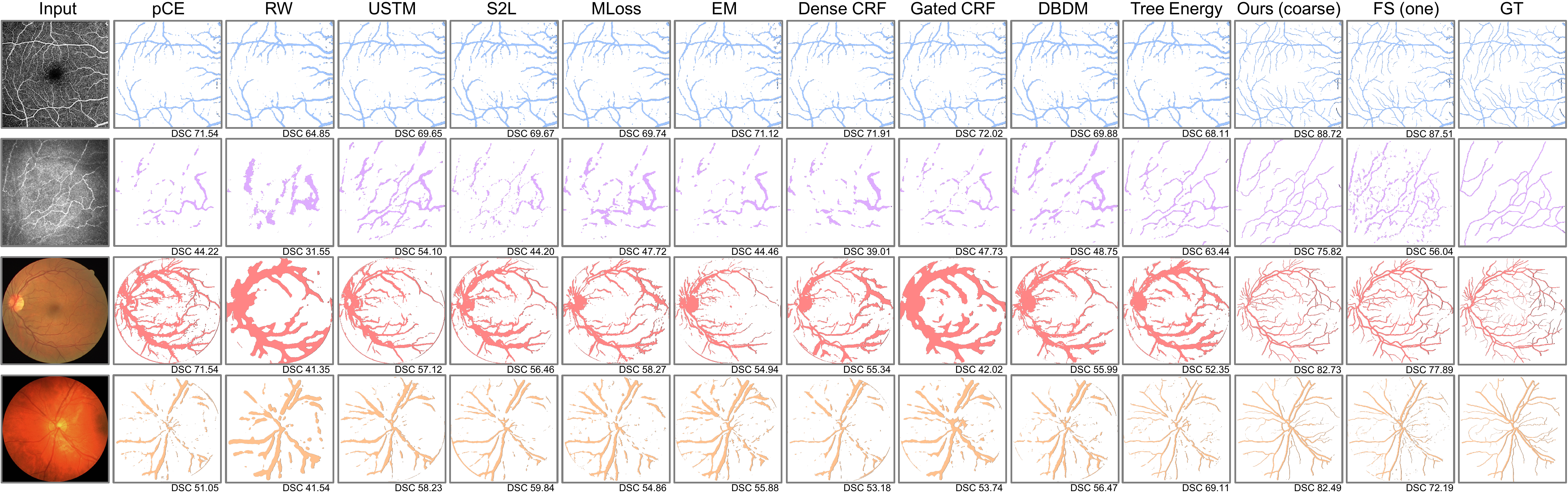}}
  \vspace{-0.5cm}
  {\caption{
      Qualitative visualization of representative results from our $S_{coarse}$ and other SOTA WSL methods under one-shot setting.
  }\label{fig:resultfig}}
\vspace{-0.3cm}
\end{figure*}

\begin{table*}[!t]
  \centering
  \setlength{\tabcolsep}{1.2mm}
  \footnotesize
  \renewcommand\theadgape{\Gape[1.8mm][0mm]}
  {\caption{Comparison with NLL methods on OCTA500 and DRIVE. \textbf{M} and \textbf{S} respectively indicate full mask and noisy skeleton. \textbf{FS} denotes fully-supervised learning. The best results are highlighted in bold, and the second-best results are underlined.}\label{tab:comparison3}}
  \renewcommand\arraystretch{1.15}
  \resizebox{\linewidth}{!}{
  \begin{tabular}{lcccccccccc}  
  \specialrule{0.1em}{0pt}{0pt}
  \multirow{2}{*}{Method} & \multirow{2}{*}{Label} & \multicolumn{4}{c}{OCTA500} & \multirow{2}{*}{Label} & \multicolumn{4}{c}{DRIVE} \\ \cline{3-6} \cline{8-11} 
                      &                        & DSC$\uparrow$          & ASSD$\downarrow$       & SE$\uparrow$    & SP$\uparrow$  &                        & DSC$\uparrow$         & ASSD$\downarrow$ & SE$\uparrow$    & SP$\uparrow$       \\
  \specialrule{0.1em}{0pt}{0pt}
GCE \citep{zhang2018generalized}                     & 200S                   &  80.51$\pm$1.80            &    0.89$\pm$0.20   & 85.27$\pm$3.46& 97.40$\pm$0.48      & 20S                    &  73.79$\pm$1.84           &    \underline{1.73$\pm$0.28}  &79.17$\pm$5.28 &95.92$\pm$0.77        \\
GCE \citep{zhang2018generalized}                      & 1M\&199S               &  80.74$\pm$1.78            &  0.86$\pm$0.18    & 82.00$\pm$3.95 & \underline{97.94$\pm$0.44}       & 1M\&19S                &   74.42$\pm$1.77          &   1.94$\pm$0.41      &76.06$\pm$5.46 &\underline{96.75$\pm$0.67}    \\
COT \citep{han2018co}                     & 200S                   &  76.58$\pm$3.66            &   1.31$\pm$0.37     & \textbf{88.83$\pm$2.99}  & 95.79$\pm$0.65   & 20S                    &   73.32$\pm$2.14          &    2.18$\pm$0.45   &74.52$\pm$5.36 &96.71$\pm$0.69     \\
COT \citep{han2018co}                     & 1M\&199S               &   77.52$\pm$3.72           &   1.28$\pm$.0.37    & 88.37$\pm$3.09 & 96.14$\pm$0.70      & 1M\&19S                &  73.82$\pm$2.24           &  2.22$\pm$0.45   &76.85$\pm$5.63 &96.40$\pm$0.95        \\
TriNet \citep{zhang2020robust}                  & 200S                   &   77.96$\pm$2.71           &   1.17$\pm$0.27   &88.18$\pm$3.37 &96.28$\pm$0.58        & 20S                    &    75.47$\pm$1.62         &    1.76$\pm$0.32    &\underline{82.37$\pm$4.99} &95.86$\pm$0.87     \\
TriNet \citep{zhang2020robust}                  & 1M\&199S               &  78.17$\pm$2.45            &   1.14$\pm$0.24     & \underline{88.52$\pm$3.31} & 96.28$\pm$0.57     & 1M\&19S                &  75.53$\pm$1.53           &    1.78$\pm$0.33   &81.41$\pm$5.13 &96.06$\pm$0.83      \\
CLSLS \citep{zhang2020characterizing}                  & 200S                   &  80.72$\pm$1.78            &  0.90$\pm$0.17   & 83.77$\pm$3.71 &  97.67$\pm$0.52       & 20S                    &   73.72$\pm$1.94          &    1.76$\pm$0.34    &78.34$\pm$5.71 &96.07$\pm$0.77     \\
CLSLS \citep{zhang2020characterizing}                  & 1M\&199S               &  \underline{80.77$\pm$1.78}            &   \underline{0.84$\pm$0.16}   & 82.87$\pm$3.78  &  97.81$\pm$0.49      & 1M\&19S                &   74.25$\pm$1.82          &     1.93$\pm$0.37   & 77.61$\pm$5.49& 96.39$\pm$0.73    \\
DAST \citep{yang2022learning}                    & 1M\&199S               &    66.20$\pm$4.74          &   2.70$\pm$0.73    & 70.32$\pm$4.76&95.94$\pm$0.95       & 1M\&19S                &  \underline{76.33$\pm$3.75}           &     1.97$\pm$0.49      &\textbf{85.17$\pm$5.25} &95.57$\pm$1.67 
\\
Ours (coarse)                    & 1S                     &   \textbf{{85.47$\pm$2.38}}           &  \textbf{{0.81$\pm$0.21}}      & 86.27$\pm$3.81 &  \textbf{98.47$\pm$0.42}    & 1S                     &   \textbf{{77.99$\pm$1.97}}          &    \textbf{{1.71$\pm$0.34}}     &73.74$\pm$4.55 & \textbf{98.25$\pm$0.33}   \\
FS\cellcolor{gray!30} & 1M\cellcolor{gray!30} & 84.60$\pm$3.33\cellcolor{gray!30}  & 0.98$\pm$0.35\cellcolor{gray!30} & 88.60$\pm$2.89\cellcolor{gray!30} & 97.93$\pm$0.81\cellcolor{gray!30} & 1M\cellcolor{gray!30} & 76.89$\pm$2.52\cellcolor{gray!30} & 2.05$\pm$0.49\cellcolor{gray!30} & 78.51$\pm$6.06\cellcolor{gray!30} & 97.04$\pm$1.27\cellcolor{gray!30}
\\
\specialrule{0.1em}{0pt}{0pt}
\end{tabular}
}

\end{table*}

\begin{table}[t]
  \centering
  \setlength{\tabcolsep}{0.7mm}
  \footnotesize
  \renewcommand\theadgape{\Gape[1.8mm][0mm]}
  {\caption{Comparison of YoloCurvSeg's coarse stage one-shot segmentation performance using background images respectively extracted by median filtering and the Inpainter in the image synthesization process. \textit{Median} represents median filtering.}  
  \label{tab:med}}
  \renewcommand\arraystretch{1.15}
  \tiny
  \resizebox{\linewidth}{!}{
  \begin{tabular}{cccc|cccc}
  \specialrule{0.1em}{0pt}{0pt}
  \multicolumn{4}{c|}{OCTA500}                                           & \multicolumn{4}{c}{DRIVE}                          \\
  Sample                   & Method    & DSC$\uparrow$            & ASSD$\downarrow$          & Sample                  & Method    & DSC$\uparrow$   & ASSD$\downarrow$ \\ \hline
  \multirow{2}{*}{NO. 3}  & Median    &  83.46          &    1.03        & \multirow{2}{*}{NO. 21} & Median    &  73.84  & 1.73   \\
                           & Inpainter & 84.94    & 0.87          &                         & Inpainter & 77.57  & 1.46 \\
                           \specialrule{0.05em}{0pt}{0pt}
  \multirow{2}{*}{NO. 12}  & Median    & 83.53          & 1.03          & \multirow{2}{*}{NO. 22} & Median    &  71.39  & 2.20   \\
                           & Inpainter & 85.47          & 0.81          &                         & Inpainter & 77.38 & 1.67 \\
                           \specialrule{0.05em}{0pt}{0pt}
  \multirow{2}{*}{NO. 42} & Median    & 82.97   & 1.11   & 
  \multirow{2}{*}{NO. 24} & Median    & 73.14   &  1.85  \\
                           & Inpainter & 85.16          & 0.87          &                         & Inpainter & 77.41 & 1.81\\
                           \specialrule{0.05em}{0pt}{0pt}
  \multirow{2}{*}{NO. 43}  & Median    & 83.08          & 1.11          & \multirow{2}{*}{NO. 25} & Median    & 73.24 & 2.01 \\
                           & Inpainter & 85.17    & 0.87          &                         & Inpainter & 77.99  & 1.71 \\
                           \specialrule{0.05em}{0pt}{0pt}
  \multirow{2}{*}{NO. 50}  & Median    & 82.93           &   1.11         & 
  \multirow{2}{*}{NO. 27} & Median    & 73.69 & 1.88 \\
                           & Inpainter & 85.08          & 0.87          &                         & Inpainter & 77.86 & 1.59 \\
                           \specialrule{0.05em}{0pt}{0pt}
  \multirow{2}{*}{NO. 102} & Median    & 82.86   & 1.11   & 
  \multirow{2}{*}{NO. 29} & Median     &  74.61  & 1.91  \\
                           & Inpainter & 84.97          & 0.86          &                         & Inpainter & 77.66 &  2.11 \\
                           \specialrule{0.05em}{0pt}{0pt}
  \multirow{2}{*}{NO. 149}  & Median    & 83.10          & 1.04          & \multirow{2}{*}{NO. 33} & Median    &  74.30  &  1.80  \\
                           & Inpainter & 85.27    & 0.84          &                         & Inpainter & 77.44  & 1.39 \\
                           \specialrule{0.05em}{0pt}{0pt}
  \multirow{2}{*}{NO. 175}  & Median    &    83.19         & 1.07            & \multirow{2}{*}{NO. 38} & Median    & 73.96 & 1.75 \\
                           & Inpainter & 85.17          & 0.85          &                         & Inpainter & 77.74 & 1.59 \\
                           \specialrule{0.05em}{0pt}{0pt}
  \multirow{2}{*}{NO. 184} & Median    &  82.82  &  1.09  & 
  \multirow{2}{*}{NO. 39} & Median    & 73.99   &   2.25 \\
                           & Inpainter & 85.17          & 0.84          &                         & Inpainter & 77.59 & 2.19\\
                           \specialrule{0.05em}{0pt}{0pt}
  \multirow{2}{*}{NO. 199}  & Median    &   82.58          &   1.13         & \multirow{2}{*}{NO. 40} & Median    &  71.31  &  2.53  \\
                           & Inpainter & 85.09    & 0.88          &                         & Inpainter & 77.53  & 2.13 \\
                           \specialrule{0.05em}{0pt}{0pt}
  \multirow{2}{*}{Overall}  & Median    &   83.05$\pm$0.29          &  1.08$\pm$0.04           & \multirow{2}{*}{Overall} & Median    & 73.35$\pm$1.14   & 1.99$\pm$0.26   \\
                           & Inpainter &   85.15$\pm$0.15          &   0.86$\pm$0.02          &         
                           & Inpainter &  77.61$\pm$0.20  &  1.77$\pm$0.29  \\
                           \specialrule{0.1em}{0pt}{0pt}
  \end{tabular}
  }
  \vspace{-0.2cm}
  \end{table}

\begin{figure*}[!h]
  \centering
  \hspace{-0.2cm}
  \includegraphics[width=0.42\linewidth]{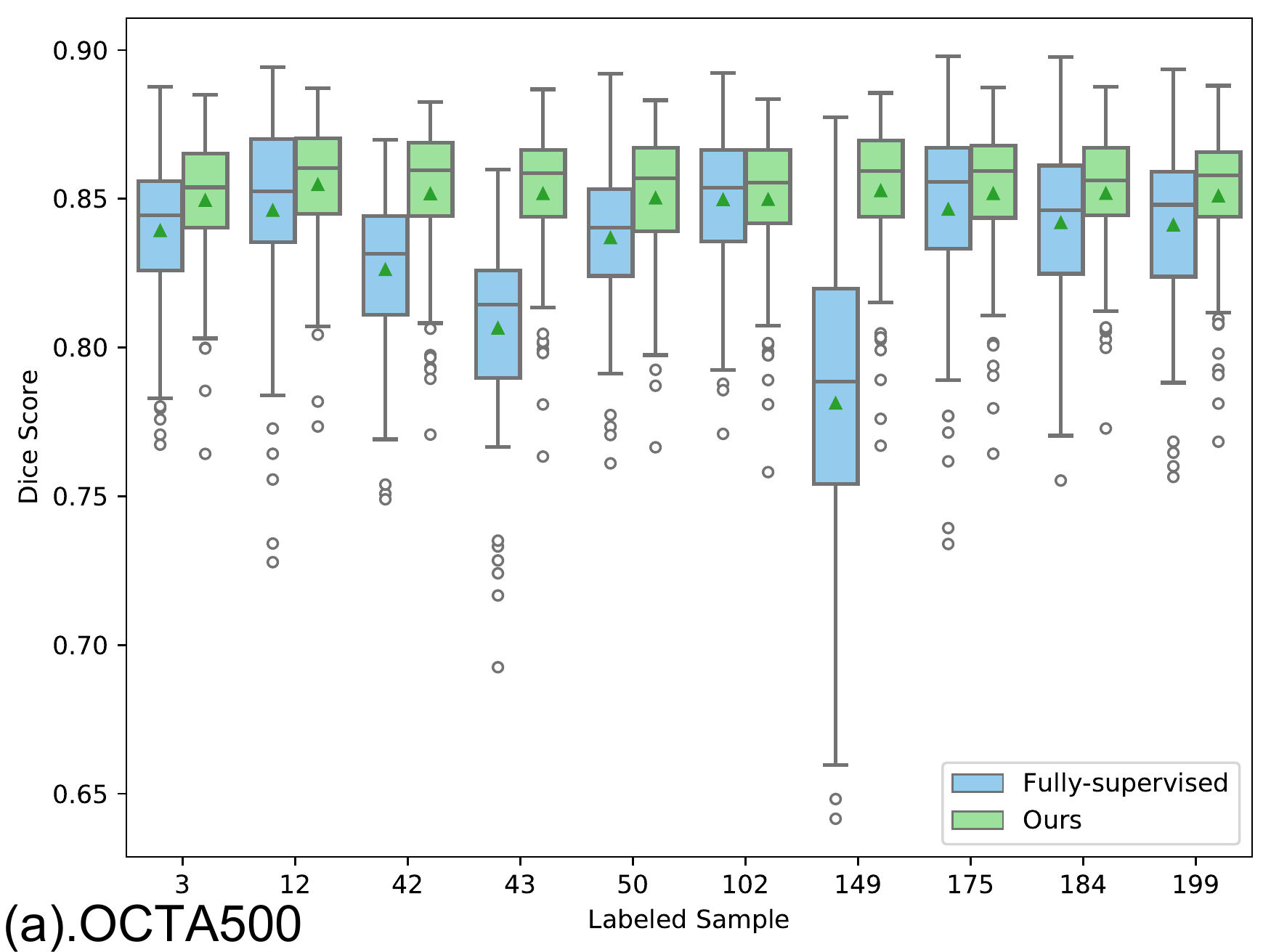}\hspace{+0.1cm}
  \includegraphics[width=0.42\linewidth]{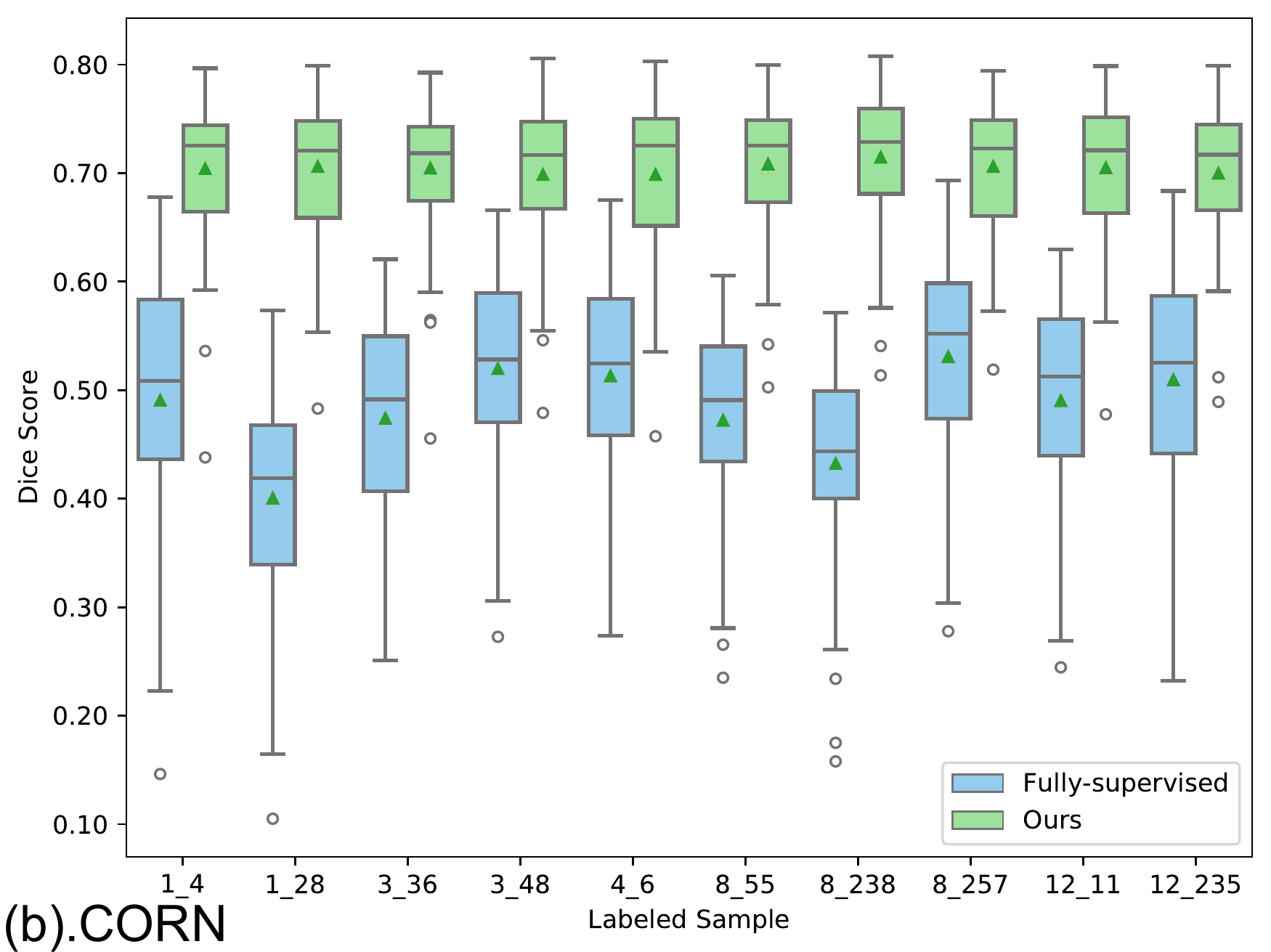}\\
  \vspace{+0.1cm}
  \includegraphics[width=0.42\linewidth]{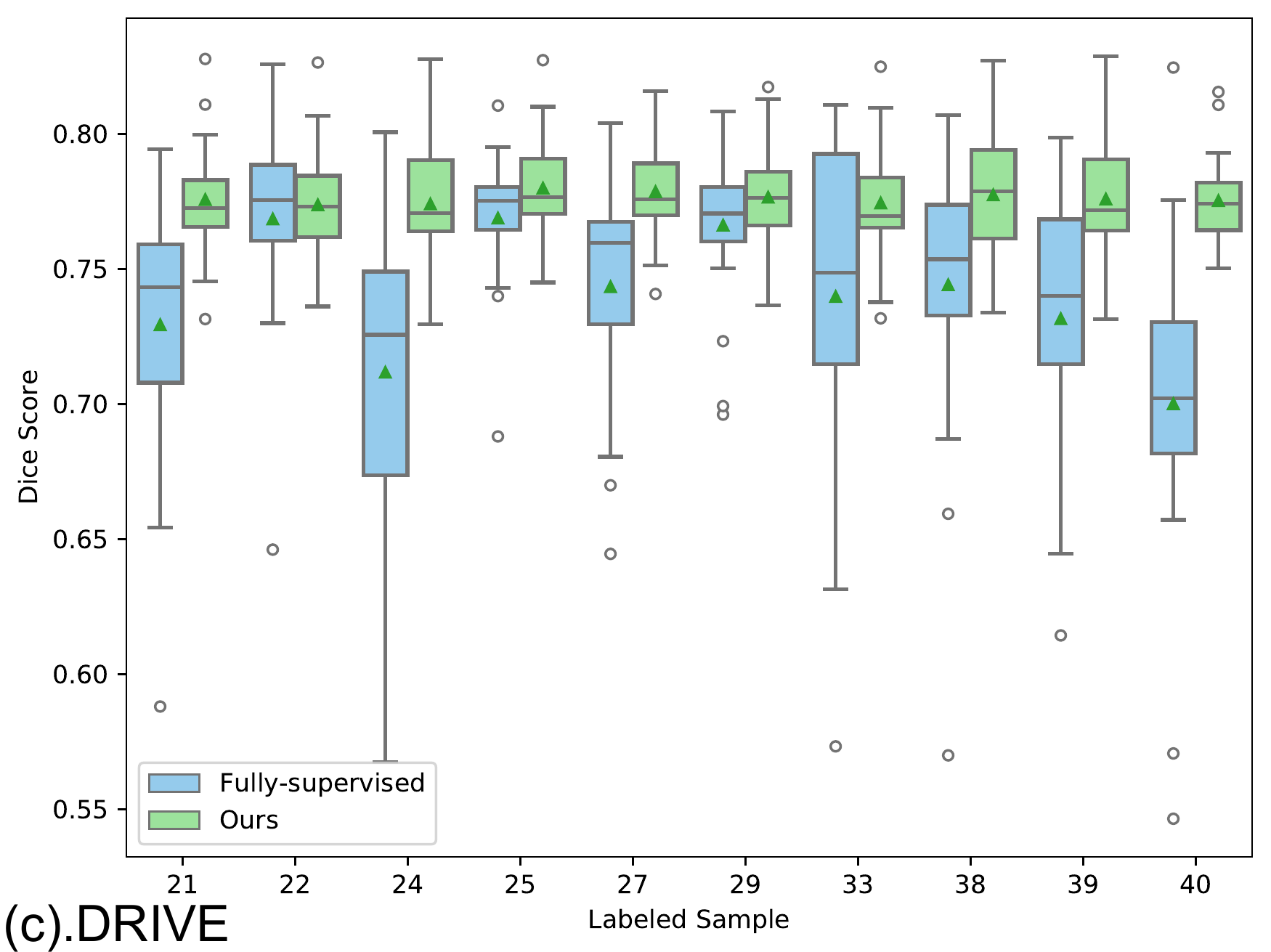}\hspace{+0.1cm}
  \includegraphics[width=0.42\linewidth]{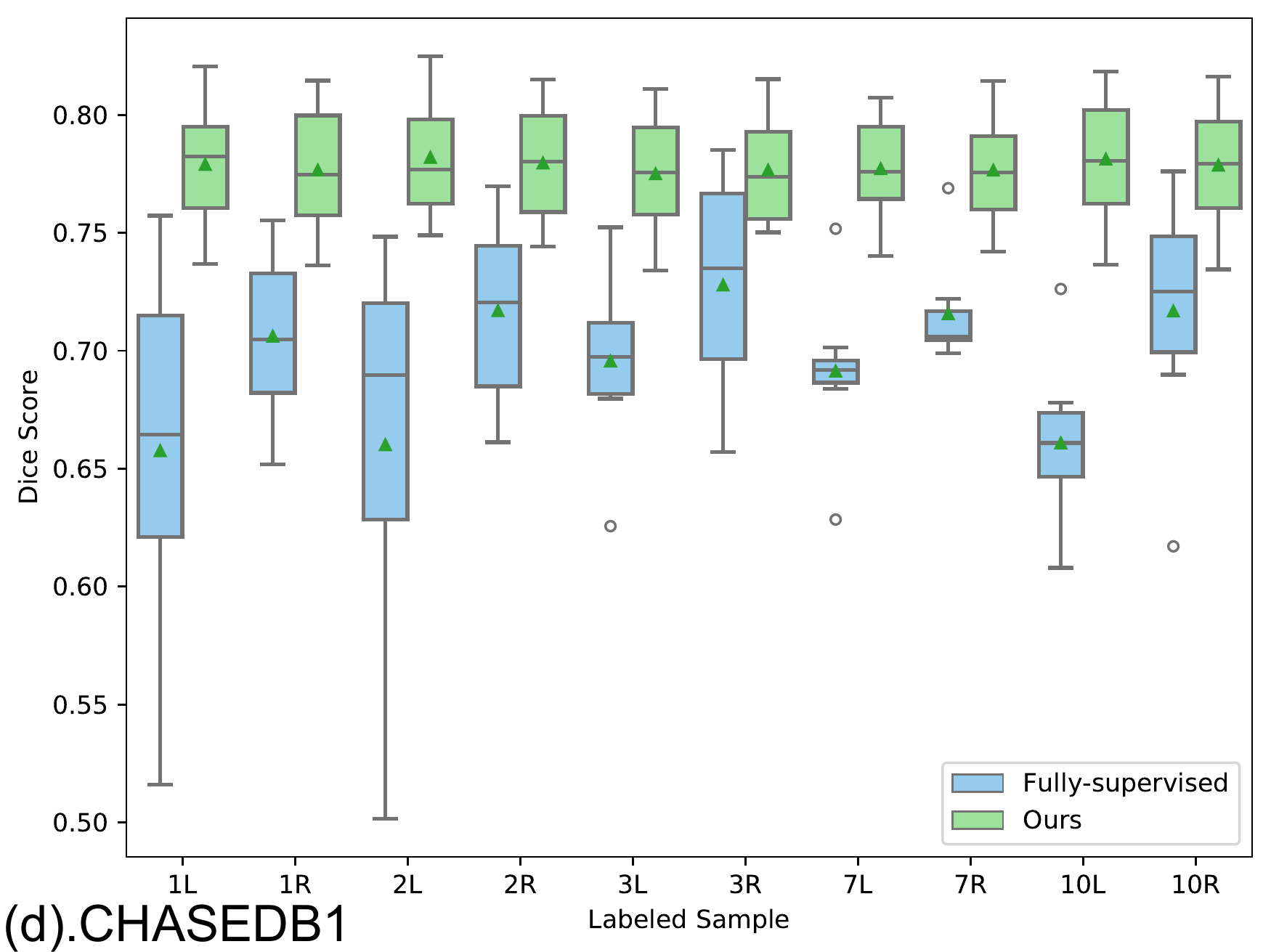}

  \vspace{-0.15cm}
\caption{Performance of YoloCurvSeg (coarse stage) given different labeled samples under one-shot setting.}
\label{fig:boxplot} 
\end{figure*}

\begin{figure*}[!h]
  \centering{\includegraphics[width=0.9\linewidth]{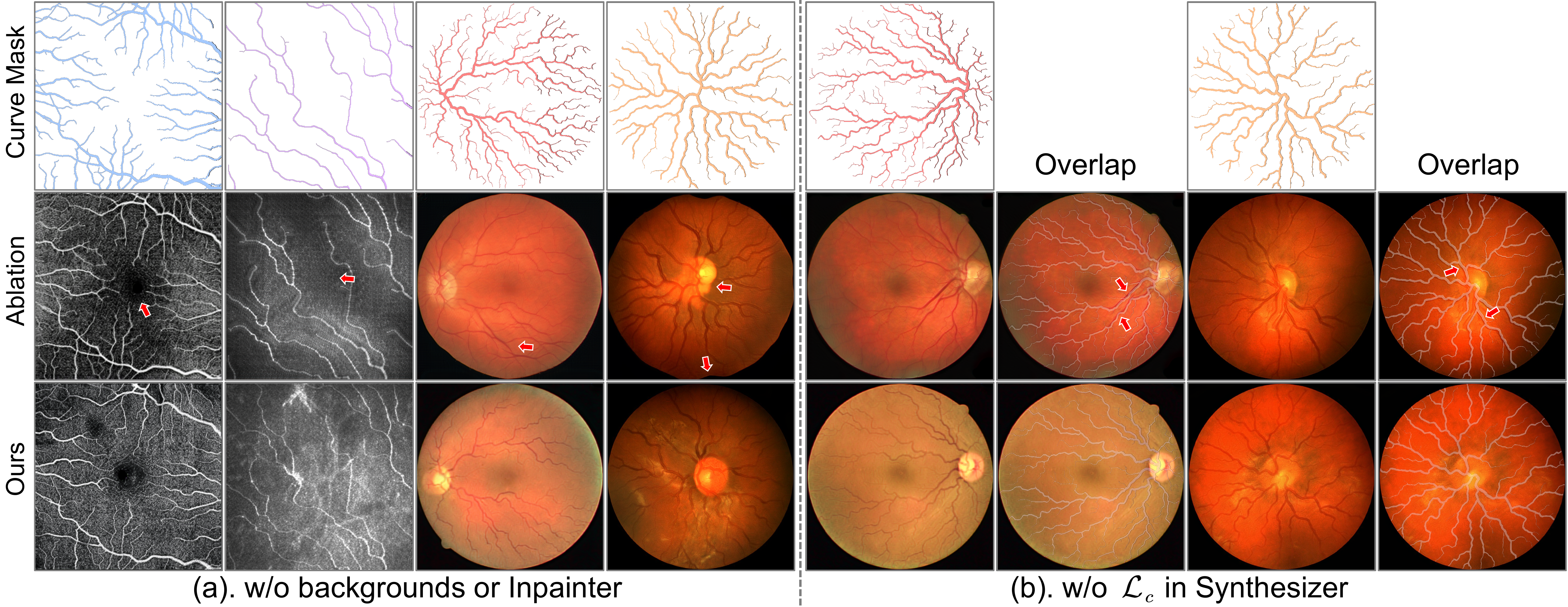}}
  \caption{
      Visualization of representative synthetic results from the ablation study. Arrows mark the unrealistic background regions, structures, or misalignments between the masks and the regions of interest.
  }
  \label{fig:ablation}
\end{figure*}

\subsubsection{Comparison with WSL methods} 

We compare YoloCurvSeg with 11 scribble-supervised segmentation methods employing the same skeleton set that we generate: pCE (partial cross-entropy loss, baseline); random walker pseudo labeling (RW); uncertainty aware self-ensembling and transformation-consistent model (USTM); Scribble2Label (S2L); Mumford–Shah Loss (MLoss); entropy minimization (EM); dense CRF loss; gated CRF loss; active contour loss (AC); dual-branch network with dynamically mixed pseudo label supervision (DBDM); and tree energy loss, the results of which are shown in Table \ref{tab:comparison1} and Table \ref{tab:comparison2}. For fair comparisons, YoloCurvSeg does not go through the fine stage and is denoted as \textit{Ours (coarse)} (i.e., the performance of $S_{coarse}$). 
{The upper part of each table indicates that all training data are sparsely labeled and all training set images are utilized to train the Segmenter in compared methods (for our method, they are used to train the Inpainter, Synthesizer and Segmenter). In the lower part, ``One'' indicates that only one sample is labeled and all other data are unlabeled and are \textbf{not utilized}. Please be aware that in the one-shot setting, the Inpainter and the Synthesizer are trained using only the single image of the sparsely annotated sample and this remains consistent throughout all experimental procedures outlined in the following sections. Additionally, all segmentation networks are randomly initialized without using any pre-trained parameters.}

As shown in Table \ref{tab:comparison1}, pCE achieves relatively low segmentation performance in most cases, as it only supervises sparsely annotated regions. RW is clearly not suitable for thin and elongated curvilinear structures, as its arbitrary expansion introduces a significant amount of noise in pseudo-labels, resulting in performance that is lower than the baseline. Most comparison methods attempt to generate and refine pseudo labels through introducing various forms of CRF loss (Dense CRF and Gated CRF), combining well-designed network architecture and consistency learning (e.g., USTM and DBDM), or employing more advanced forms of loss (e.g., MLoss, AC and Tree Energy). Although effective, these methods still show a significant performance gap compared to fully-supervised performance, even when all samples are annotated, let alone when only a single sample is annotated. Among all compared methods, Tree Energy achieves the second-best performance in most cases, but it still has a significant gap compared to YoloCurvSeg, and is highly affected by noise in sparse annotations. YoloCurvSeg achieves the best performance on all datasets under both settings, outperforming other WSL methods by \textbf{large margins}. Comparing "All" versus "One", apparently, YoloCurvSeg is not sensitive to the sample size of the labeled data and achieves \textbf{96.1\%}, \textbf{106.3\%}, \textbf{95.3\%} and \textbf{96.2\%} fully-supervised performance on the four datasets with only \textbf{0.14\%}, \textbf{0.03\%}, \textbf{1.40\%} and \textbf{0.65\%} labeled pixels in terms of DSC. It is worth noting that, even when only a single sample is annotated, YoloCurvSeg (bottom second last row of the table) still achieves superior performance compared to all comparison methods when all samples are annotated (top half of the table). Representative visualization results are shown in Fig. \ref{fig:resultfig}.

\subsubsection{Comparison with NLL methods}

In Table \ref{tab:comparison3}, we also compare YoloCurvSeg with several NLL methods, including generalized cross-entropy loss (GCE), co-teaching (COT), TriNet, confident learning with spatial label smoothing (CLSLS) and divergence-aware selective training (DAST) on OCTA500 and DRIVE. Most of these methods allow training under conditions of all noisy samples as well as mixed noisy (S in Table \ref{tab:comparison3}) and fully-supervised (M in Table \ref{tab:comparison3}) samples. It can be seen that with the inclusion of fully-supervised samples in training, these methods can generally achieve a certain degree of performance improvements. However, despite utilizing one full mask and multiple or even all skeleton samples, these methods are still inferior to YoloCurvSeg which employs only one skeleton sample.
We also find that all NLL methods perform worse than the fully-supervised model (FS in Table \ref{tab:comparison3}) trained solely with the same single fully labeled sample, illustrating that additional noisily labeled samples are not beneficial to model performance under such noisy conditions.

\begin{figure}[t]
  \centering
  \includegraphics[width=\linewidth]{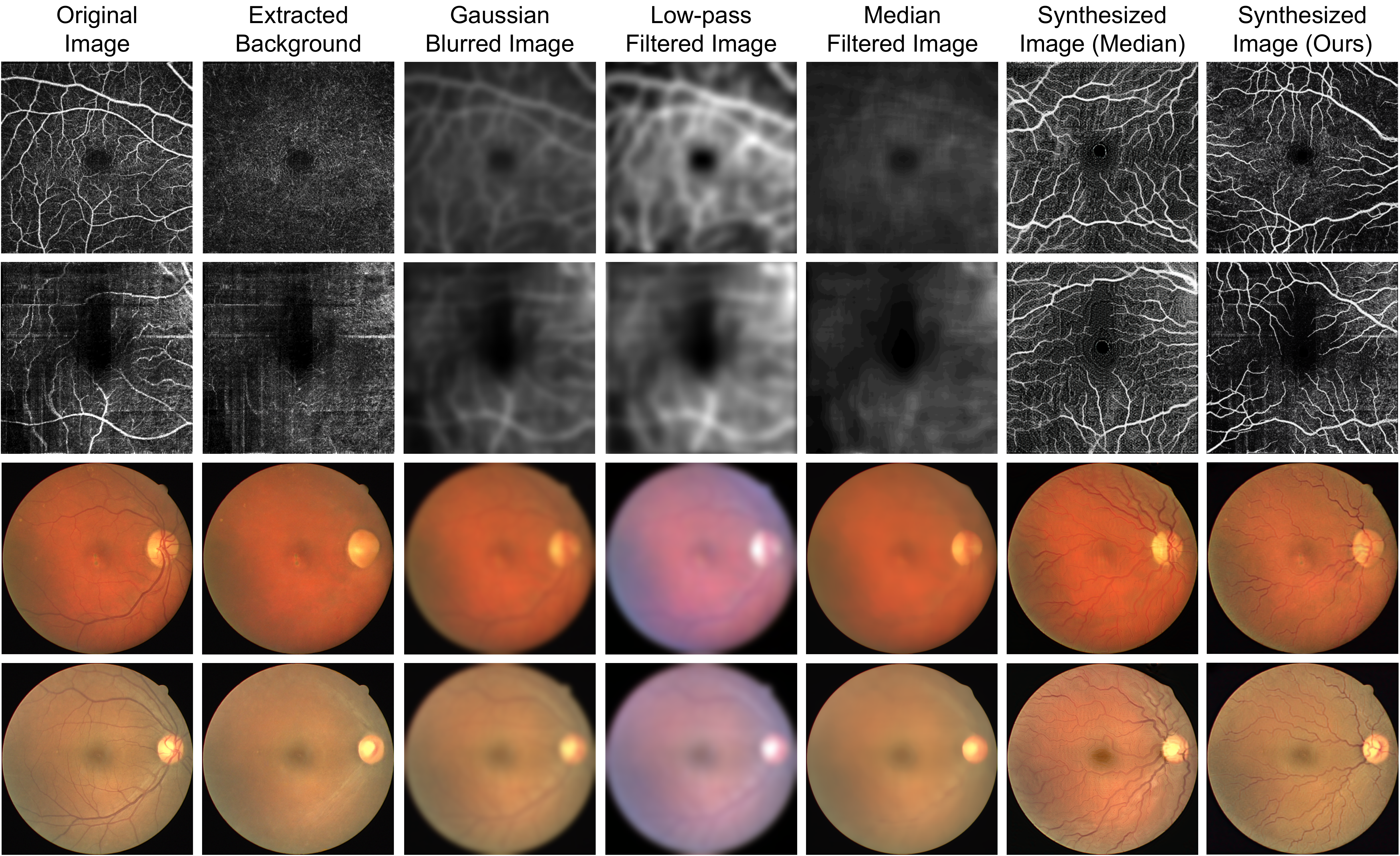}
  \vspace{-0.5cm}
{\caption{Examples of background images acquired through distinct unsupervised techniques, such as Gaussian blurring, low-pass filtering, and median filtering, along with instances of synthesized images using the median-filtered images and our Inpainter-extracted images as backgrounds. Zoom in for details.}\label{fig:ablation-med}}
\end{figure}


\subsection{Robustness Analysis and Ablation Study}

To verify the robustness of YoloCurvSeg for the selected one-shot sparsely labeled sample, we randomly select 10 samples from each dataset and compare the performance with the fully-supervised model trained on the same sample. As demonstrated in Fig. \ref{fig:boxplot}, YoloCurvSeg exceeds full supervision in almost all cases and delivers highly stable performance decoupled from image/annotation quality which nevertheless induces great fluctuations in the performance of the fully-supervised models. In addition to robustness, the predictions from YoloCurvSeg also have smaller variances. Both aspects indicate that YoloCurvSeg is sample-insensitive and can reduce the risk of selecting a wrong sample to label.

\begin{figure}[]
  \centering
  \includegraphics[width=\linewidth]{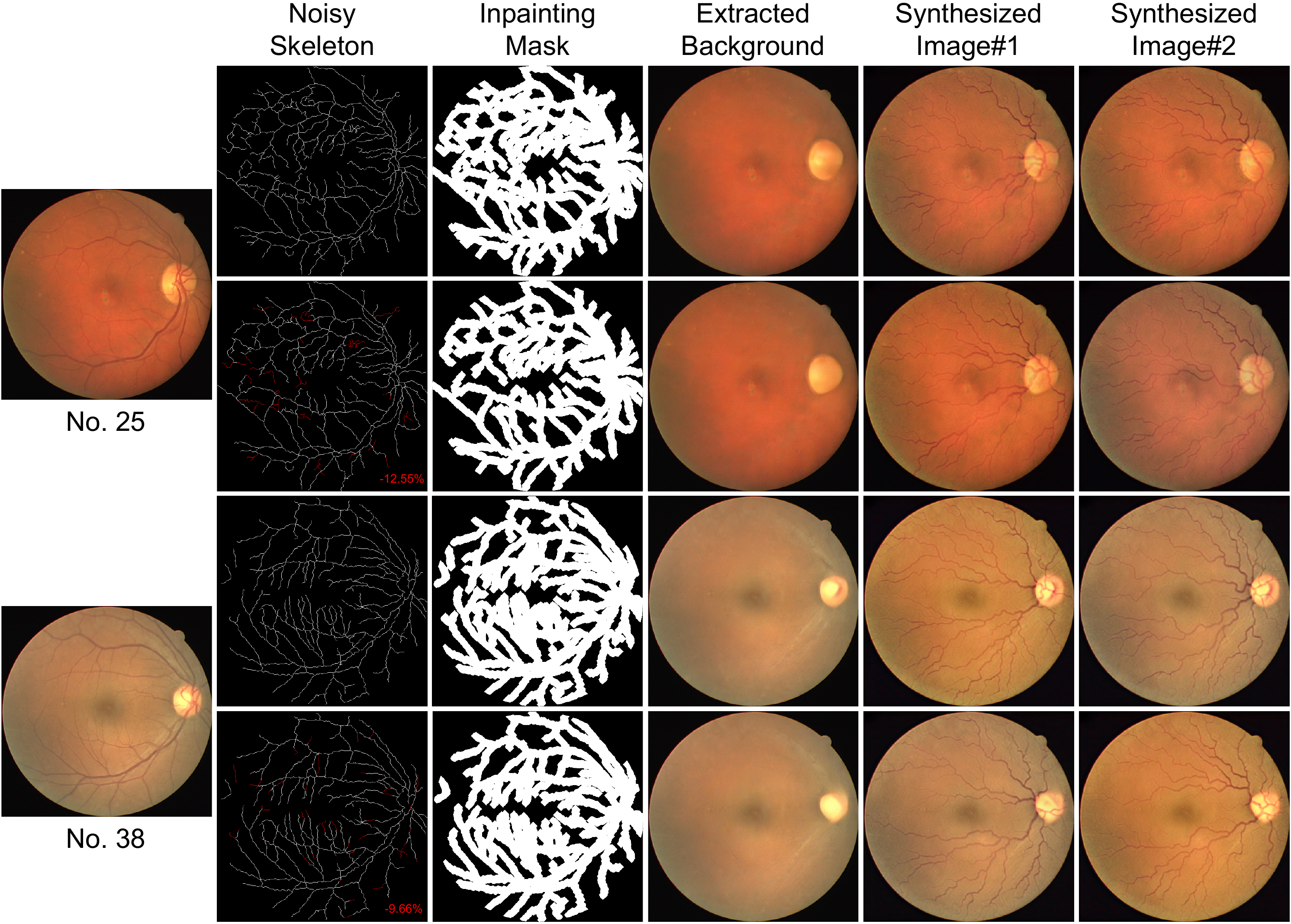}
  \vspace{-0.5cm}
{\caption{Visualization of the inpainting masks, extracted backgrounds, and synthesized images of two representative samples from the DRIVE dataset with varying levels of completeness in their noisy skeleton labels. In each noisy skeleton, the color red denotes the missing parts, and the red number indicates the percentage of omissions.}\label{fig:wipe}}
\vspace{-0.2cm}
\end{figure}

{ To investigate the impact of noisy skeleton's completeness on the segmentation performance, we conduct partial erasure analysis experiments on the skeletons. Due to the low contrast between small vessels and the background in fundus images, it is highly likely to have missing annotations on fundus images. Therefore, we select two samples from the DRIVE dataset and erase the noisy skeleton labels of some small vessels, as illustrated in Fig. \ref{fig:wipe}. Specifically, we respectively erase 12.55\% and 9.66\% of the annotated regions on samples No. 25 and No. 38. From the figure, we can clearly observe the erased areas on the noisy skeleton and the impact on the extracted background images and the synthesized images. The segmentation model's performance metrics (DSC, ASSD) on the two samples with complete noisy skeleton annotations are respective (77.99, 1.71) and (77.74, 1.59). After erasing some noisy skeletons and synthesizing the corresponding new training sets, the performance metrics of the segmentation model become (78.06, 1.83) and (78.11, 1.40), exhibiting only small fluctuations before and after erasure, thus demonstrating the proposed method's robustness. These slight performance fluctuations may be attributed to the significant dilation operation applied to the noisy skeleton, which may result in dilated masks covering the small vessels that may have been missed during annotation. Additionally, randomly generated foreground curves may also cover some small vessels, further reducing the labeling noise caused by missed annotations. On the other hand, as shown in Fig. \ref{fig:boxplot}, YoloCurvSeg demonstrates stable one-shot performance for all four datasets, despite the inevitable presence of varying degrees of label omissions in those samples. This is particularly evident in the CORN dataset (where there are already some missing annotations in the dataset itself), indicating that our approach is robust even in the presence of some degree of missing annotations.
}

As for the ablation study, we first remove the background bank extracted by the \textit{Inpainter} and perform direct curve-to-image translation. As shown in panel (a) of Fig. \ref{fig:ablation}, the synthetic images present unrealistic background texture due to the large gap between pre- and post-translation distributions (which is also shown in column B of Table \ref{tab:FID}). For high-resolution image datasets, the foregrounds of the synthetic images are distorted and fail to spatially align with the corresponding curve masks, which also occurs when we remove $\mathcal{L}_c$ of the \textit{Synthesizer} (and use CycleGAN \citep{zhu2017unpaired} for substitution), as shown in Fig. \ref{fig:ablation} (b). This indicates that the contrastive \textit{Synthesizer} (especially $\mathcal{L}_c$) is crucial for maintaining the corresponding local context at the same spatial location.

{ To more comprehensively demonstrate the significance of the Inpainter, we endeavor to generate background images through alternative unsupervised methods for subsequent image synthesis. We investigate Gaussian blurring, low-pass filtering, and median filtering. {Our objective is to selectively remove foreground vessels and simultaneously preserve the maximum amount of background details, by carefully adjusting the parameters. For Gaussian blurring, we utilize a kernel size of 9×9 and perform 25 iterations on samples from both OCTA500 and DRIVE datasets. For low-pass filtering, we generate two-dimensional Gaussian masks with a standard deviation of 0.05 based on the image dimensions. These masks are then applied to remove the low-frequency components. As for median filtering, we respectively apply kernel sizes of 29×29 and 31×31 to the OCTA500 and DRIVE datasets.} Visualization results indicate that median filtering is relatively more effective in eliminating the foreground, albeit at the cost of sacrificing details and introducing blurriness, as illustrated in Fig. \ref{fig:ablation-med}. As depicted in the last column of that figure, the image synthesis results with median-filtered backgrounds are superior to direct translation results from curves to real images. However, since the provided backgrounds are relatively blurry, there are still considerable amounts of artifacts. We then select {ten} samples from both OCTA500 and DRIVE for one-shot coarse stage performance comparisons, the results of which are presented in Table \ref{tab:med}. {Apparently, the overall performance of the ten randomly selected samples from the OCTA500 and DRIVE datasets indicates that utilizing the background extracted by the Inpainter induces overall performance gains of approximate 2\% and 4\% over the median-filtered background for one-shot (access to only one sample) segmentation}. Utilizing a noisy skeleton annotation to improve the segmentation performance is cost-effective, as evidenced by the measured annotation time for one such annotation described later.
}

\begin{figure}[htbp]
  \centering
  \includegraphics[width=\linewidth]{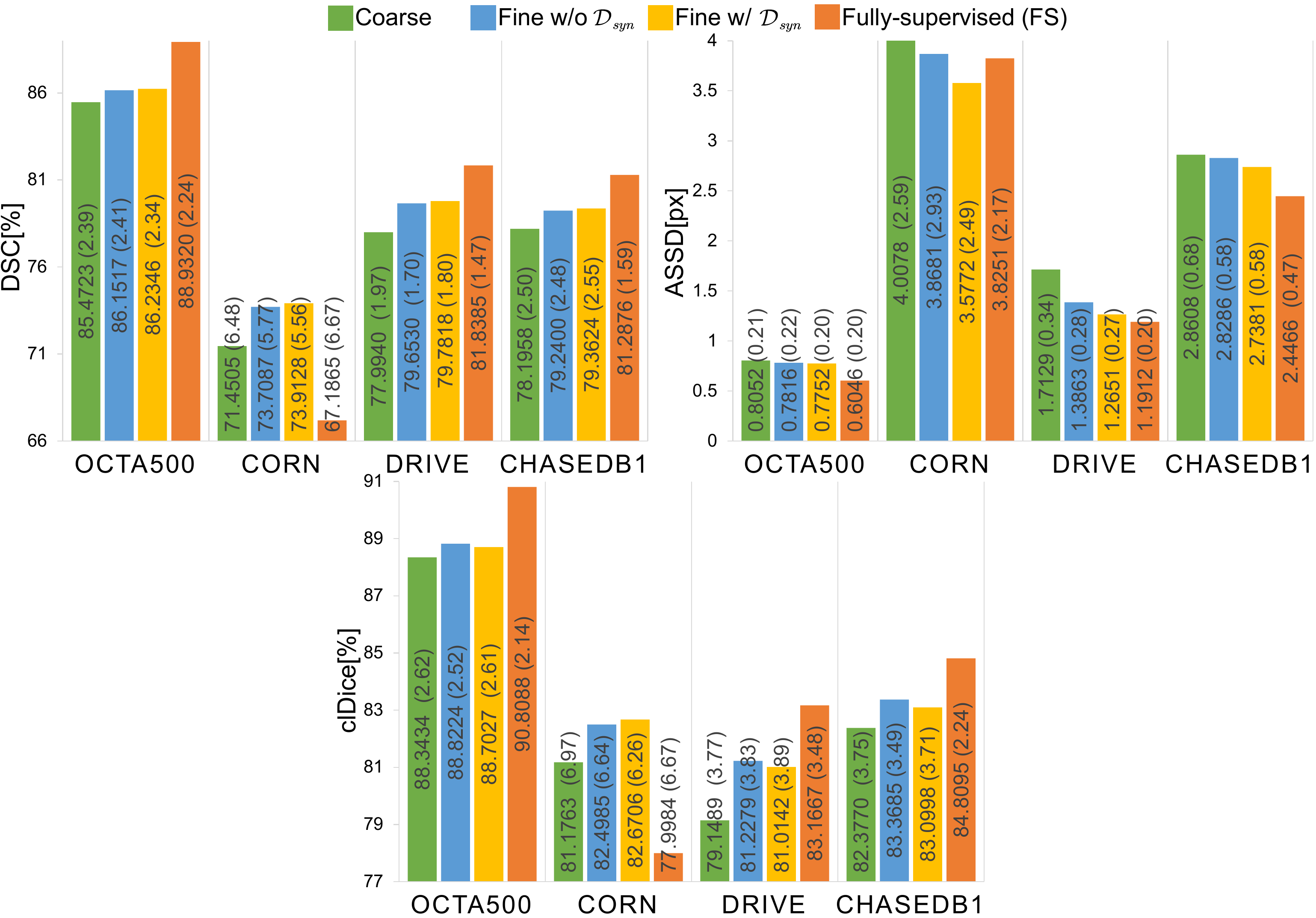}
  \vspace{-0.5cm}
{\caption{Performance of YoloCurvSeg under different training paradigms. \textbf{FS} denotes fully-supervised learning. Within the parentheses are the standard deviations.}\label{fig:barplot}}
\vspace{-0.3cm}
\end{figure}

It is worth pointing out that the results we have reported in the previous section represent the performance of $S_{coarse}$. We also explore the performance of $S_{fine}$ with and without utilizing $\mathcal{D}_{syn}$. 
{ Please noted ``without using $\mathcal{D}_{syn}$'' means that only $\mathcal{D}_{ori}$ (with predictions from $S_{coarse}$ as the pseudo-labels) is used for training $S_{fine}$.} { To comprehensively evaluate the topological connectivity and small vessel segmentation performance, the clDice \citep{shit2021cldice} metric is also computed.}
{ It can be observed from Fig. \ref{fig:barplot} that the DSC and ASSD metrics of $S_{fine}$ obtained with $D_{syn}$ in the training data are slightly better than those of $S_{fine}$ obtained without using $D_{syn}$, which may be attributed to the fact that the synthetic curves have high degrees of continuity and can reduce the model's outlier predictions. However, the clDice metric is slightly lower and this may be due to the fact that synthetic data inevitably exhibits a certain intensity gap compared to real data, especially in small vessel regions. Additionally,} 
we also conduct a performance comparison of the fully-supervised model with and without pretraining on $\mathcal{D}_{syn}$. Results show that the synthetic images from YoloCurvSeg also have great potential in serving as pretraining images; {pretraining on $\mathcal{D}_{syn}$,  followed by fine-tuning on fully supervised datasets, can further enhance the performance of the fully supervised model. Specifically, it increases the performance of the vanilla U-Net model in terms of DSC by 0.55, 0.63, 0.81, and decreases ASSD by 0.061, 0.041, 0.148, respectively on OCTA500, DRIVE and CHASEDB1}. {Ultimately, via further utilizing an additional unlabeled dataset $\mathcal{D}_{ori}$, YoloCurvSeg ($S_{fine}$) achieves \textbf{97.00\%}, \textbf{110.01\%}, \textbf{97.49\%} and \textbf{97.63\%} of the fully-supervised performance (with full masks of all available samples) with only one noisy skeleton annotation on the four datasets.}

\begin{figure*}[!h]
  \centering
  \includegraphics[width=0.49\linewidth]{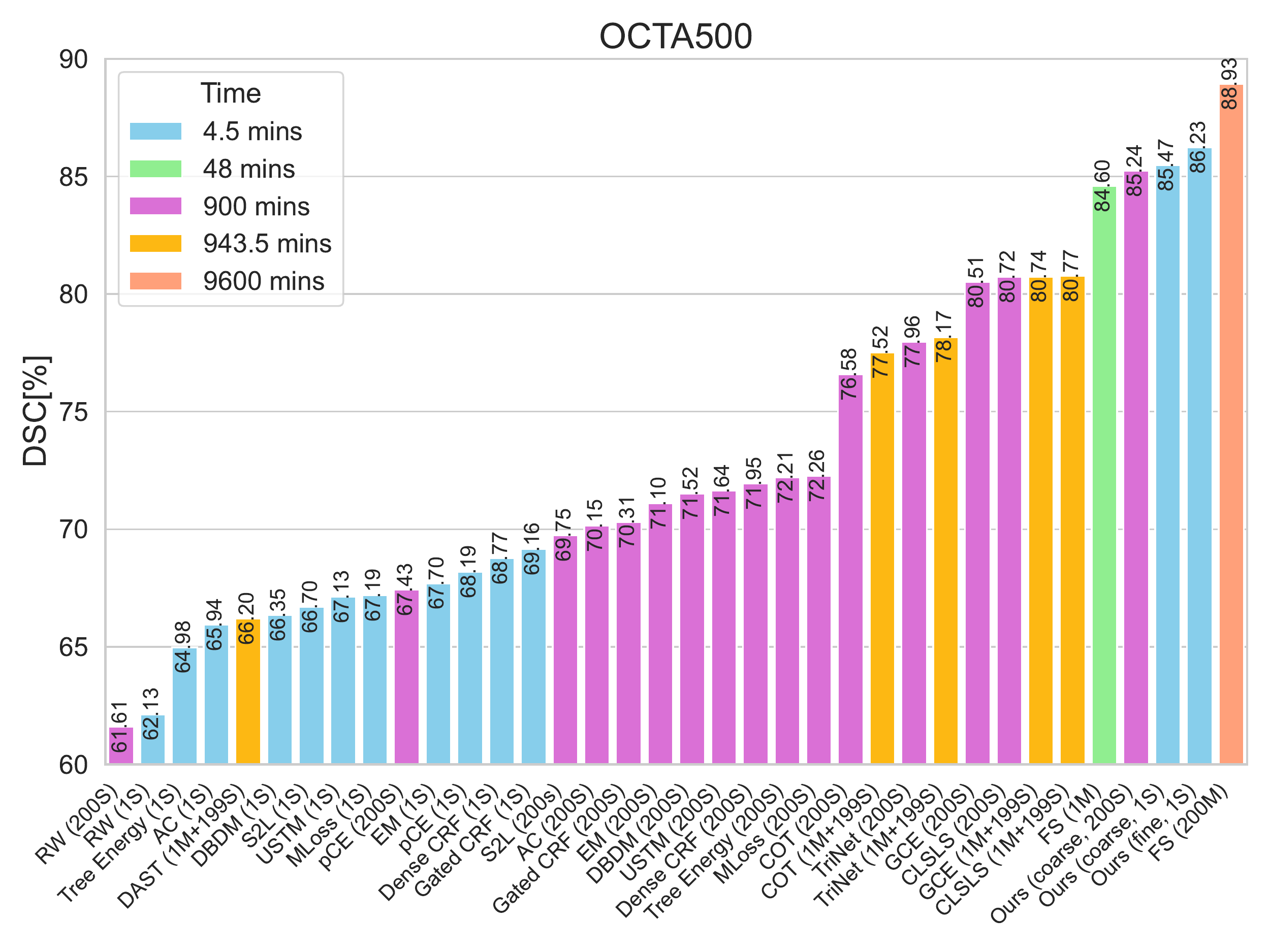}
  \includegraphics[width=0.49\linewidth]{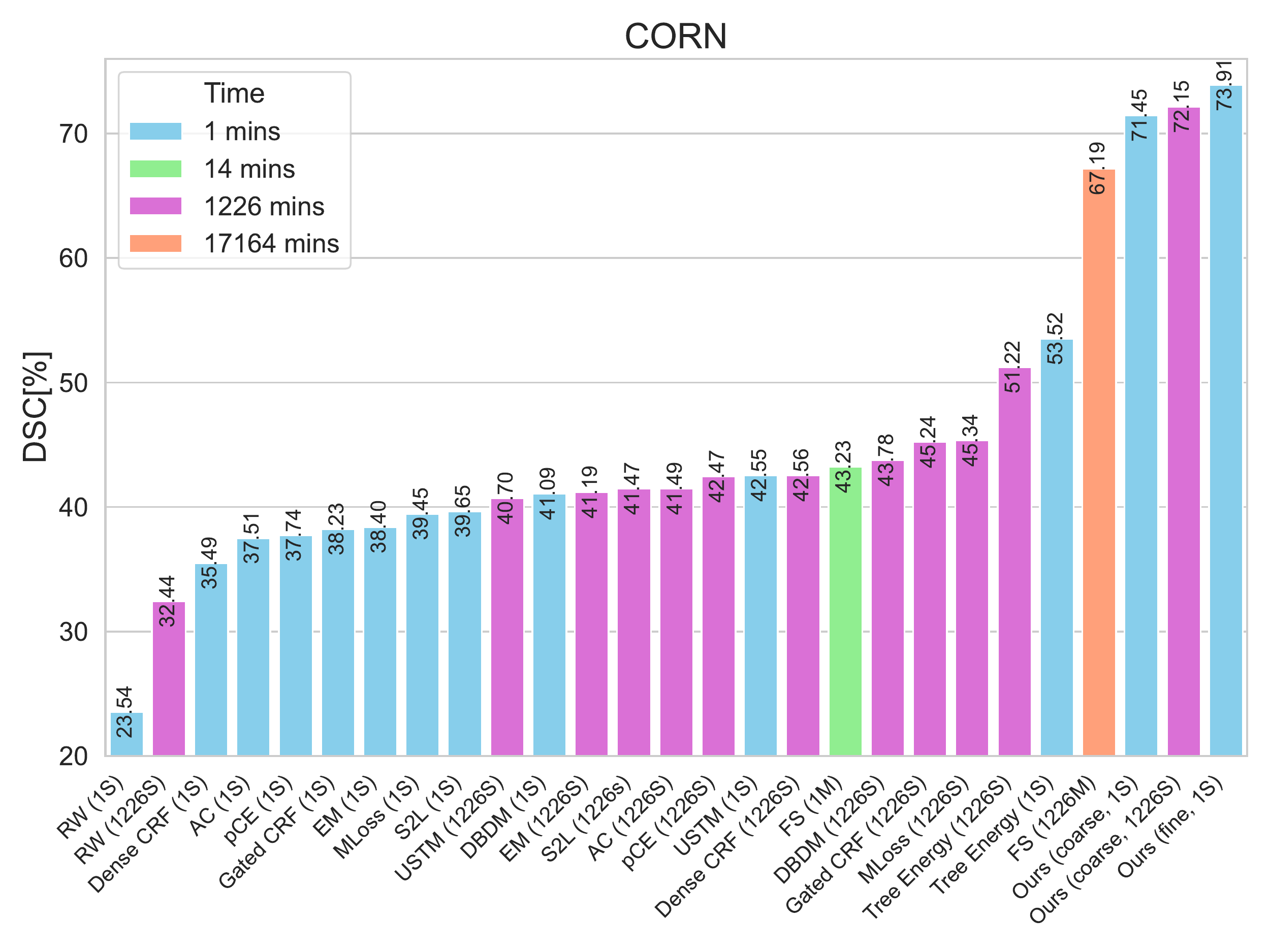}\\
  \includegraphics[width=0.49\linewidth]{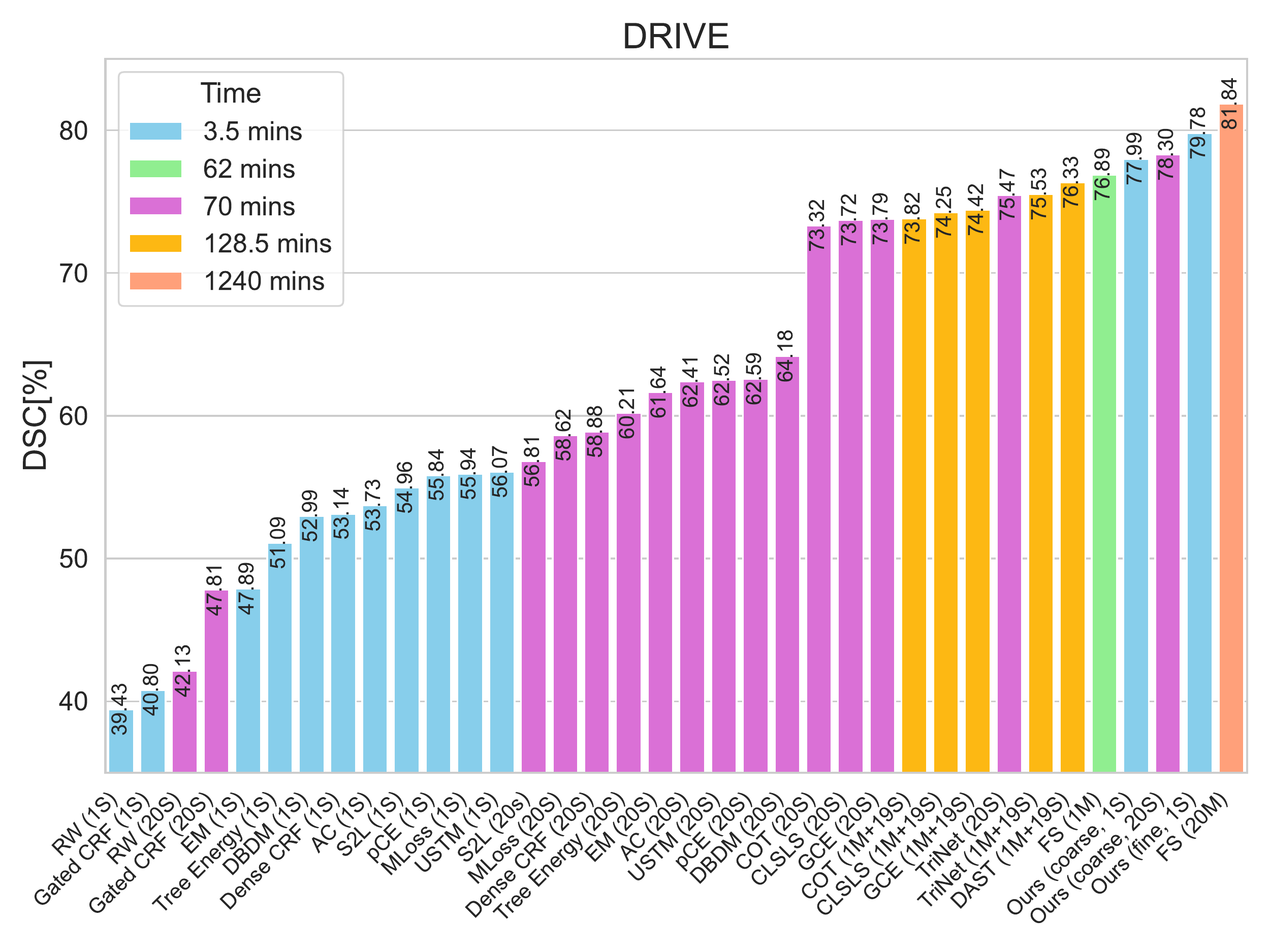}
  \includegraphics[width=0.49\linewidth]{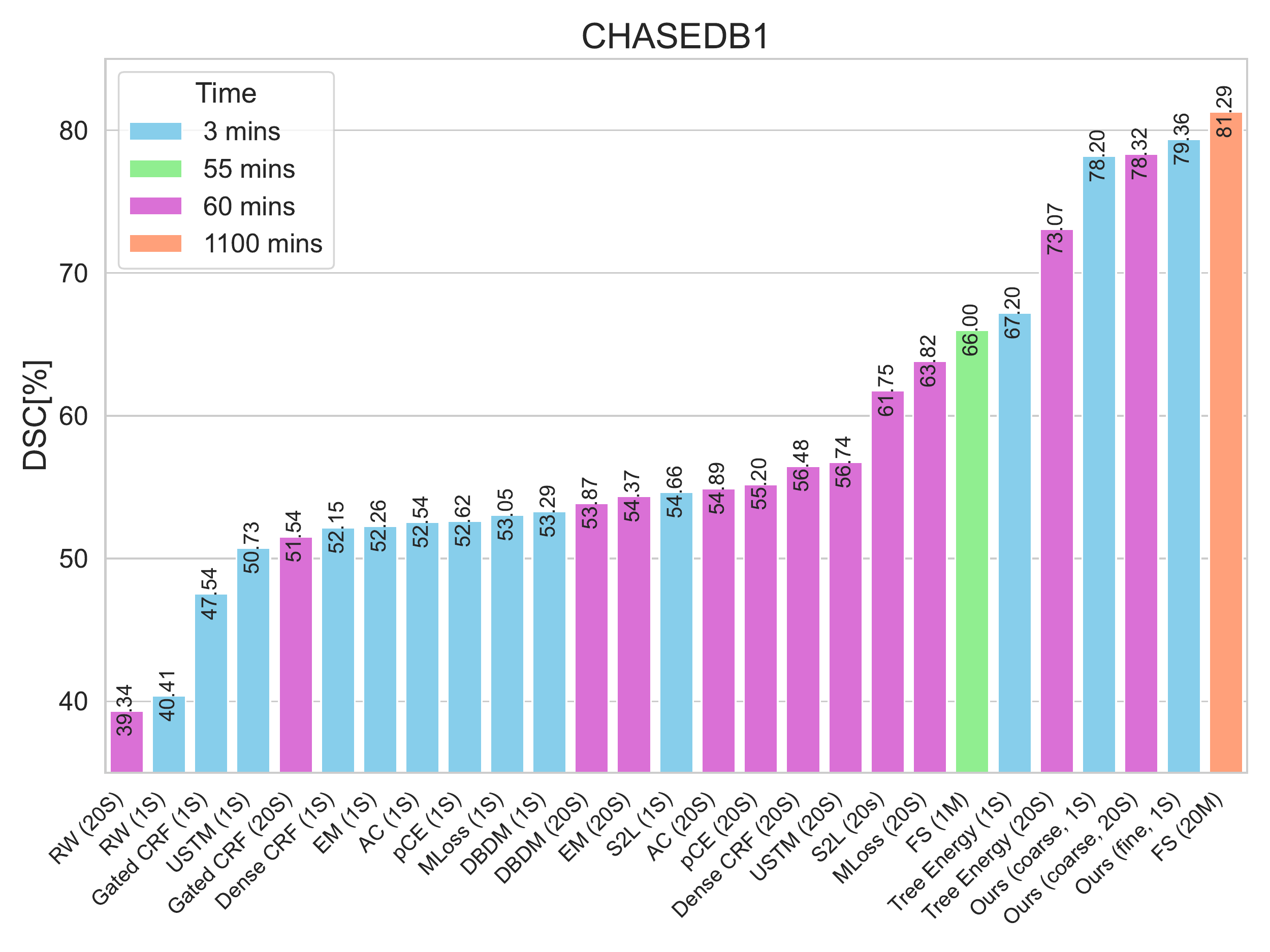}
{\caption{Segmentation accuracy (DSC) vs. annotation time for all benchmarked WSL and NLL methods, as well as the one- and all-shot fully-supervised (FS) settings. The number and type of labels used are indicated in the parentheses, with \textbf{M} and \textbf{S} respectively representing full mask and noisy skeleton.}\label{fig:time}}

\end{figure*}

\begin{figure*}[!h]
  \centering
  \includegraphics[width=0.95\linewidth]{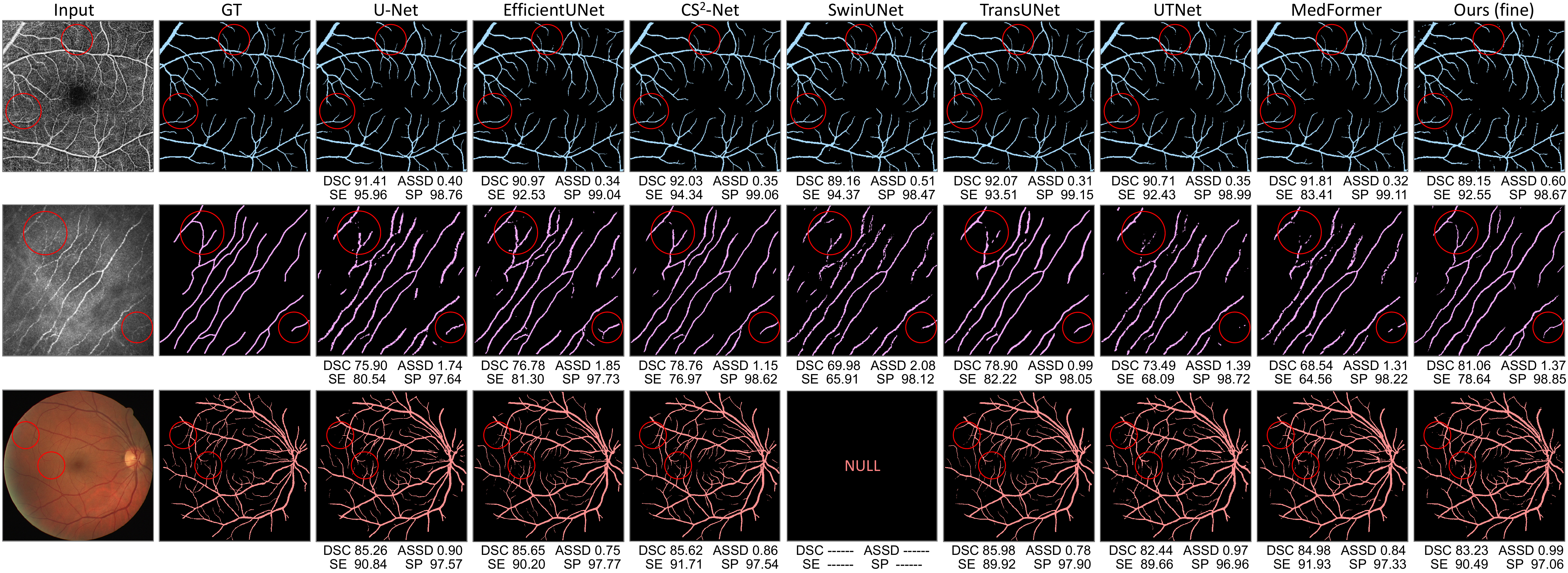}
{\caption{Qualitative visualization of representative results from SOTA FS (all samples utilized) methods and our $S_{fine}$ (with CS$^2$-Net architecture) under the one-shot setting. Red circles indicate the areas of interest worth noting. Zoom in for details.}\label{fig:sota}}
\end{figure*}

\begin{table*}[!t]
  \vspace{-0.2cm}
  \centering
  \setlength{\tabcolsep}{1.2mm}
  \footnotesize
  \renewcommand\theadgape{\Gape[1.8mm][0mm]}
  {\caption{Comparative performance of various segmentation networks under the fully-supervised setting of utilizing all samples, as evaluated on the OCTA500, CORN, and DRIVE datasets, alongside the fine stage's efficacy of our proposed method under the one-shot setting with different segmentation network architectures. * indicates the model is initialized with pre-trained weights on ImageNet. SwinUNet is not evaluated on DRIVE due to the lack of appropriate high-resolution pre-trained parameters. EfficientUNet employs the ImageNet pre-trained EfficientNet-B3 as its encoder.}\label{tab:sota}}
  \renewcommand\arraystretch{1.15}
  \resizebox{0.7\linewidth}{!}{
  \begin{tabular}{lcccccc}  
  \specialrule{0.1em}{0pt}{0pt}
  \multirow{2}{*}{Method} &  \multicolumn{2}{c}{OCTA500} &  \multicolumn{2}{c}{CORN} &  \multicolumn{2}{c}{DRIVE} \\ \cline{2-7}  
  &  DSC$\uparrow$ & ASSD$\downarrow$ &  DSC$\uparrow$ & ASSD$\downarrow$ &  DSC$\uparrow$ & ASSD$\downarrow$ \\ 
  \specialrule{0.1em}{0pt}{0pt} 
  U-Net (FS, all) \citep{ronneberger2015u}                 & 88.93$\pm$2.23                     & 0.60$\pm$0.20                    & \underline{67.19$\pm$6.67}          & 3.83$\pm$2.17                   & 81.84$\pm$1.47                    & 1.19$\pm$0.20                   \\
  EfficientUNet* (FS, all) \citep{tan2019efficientnet}        & 87.72$\pm$1.99                     & 0.63$\pm$0.17                    & 64.87$\pm$7.45                   & \textbf{2.85$\pm$1.74}          & 81.98$\pm$1.40                    & 1.20$\pm$0.27                   \\
  CS$^2$-Net (FS, all) \citep{mou2021cs2}              & \textbf{89.49$\pm$2.04}            & \textbf{0.52$\pm$0.16}           & 63.57$\pm$10.37                   & \underline{3.21$\pm$2.33}          & \textbf{82.38$\pm$1.38}           & \textbf{1.12$\pm$0.22}          \\
  SwinUNet* (FS, all) \citep{cao2023swin}             & 87.52$\pm$1.95                     & 0.60$\pm$0.15                    & 60.82$\pm$8.13                   & 3.82$\pm$2.40                   & --                       & --                     \\
  TransUNet* (FS, all) \citep{chen2021transunet}             & \underline{89.23$\pm$1.95}            & \underline{0.53$\pm$0.16}           & \textbf{68.55$\pm$5.90}          & 3.26$\pm$2.29                   & \underline{82.32$\pm$1.47}           & \underline{1.18$\pm$0.28}          \\
  UTNet (FS, all) \citep{gao2021utnet}                & 87.60$\pm$2.05                      & 0.60$\pm$0.19                      & 62.68$\pm$8.14                   & 4.21$\pm$2.52                   & 79.98$\pm$1.51                    & 1.37$\pm$0.37                   \\
  MedFormer (FS, all) \citep{gao2022data}             & 89.02$\pm$2.12                     & 0.54$\pm$0.17          & 62.17$\pm$6.95                  & 3.72$\pm$2.06                   & 81.63$\pm$1.63                    & 1.23$\pm$0.27                   \\ \hline
  Ours (Fine, one, UNet)      & 86.23$\pm$2.34                     & 0.78$\pm$0.20                    & 73.91$\pm$5.56                   & 3.58$\pm$2.49                   & 79.78$\pm$1.80                    & \textbf{1.27$\pm$0.27}                   \\
  Ours (Fine, one, TransUNet*) & \textbf{86.78$\pm$2.24}                     & \textbf{0.73$\pm$0.18}                    & \underline{74.22$\pm$5.57}                   & \underline{3.39$\pm$2.38}                   & \textbf{80.33$\pm$1.51}                    & 1.41$\pm$0.28                   \\
  Ours (Fine, one, CS$^2$-Net)     & \underline{86.70$\pm$2.27}                      & \underline{0.74$\pm$0.19}                    & \textbf{74.55$\pm$5.39}          & \textbf{3.06$\pm$1.75}          & \underline{80.20$\pm$1.55}            & \underline{1.30$\pm$0.23}    \\      
\specialrule{0.1em}{0pt}{0pt}
\end{tabular}
  }
\end{table*}

To better illustrate the time-saving benefits of our method in real clinical scenarios, {we randomly select samples from the four datasets (30 samples for OCTA500 and CORN, and 10 samples for DRIVE and CHASEDB1) and invite two ophthalmologists to annotate them in both noisy skeleton and full mask formats. We find that, annotating a noisy skeleton-style label for retinal vessels in a 6mm $\times$ 6mm OCTA image takes approximately 4.5 minutes, while annotating a full mask takes around 48 minutes due to the need for careful examination and modification of edges and details. Similarly, annotating the corneal nerve fibers in a CCM image, and the retinal vessels in DRIVE and CHASEDB1 style retinal fundus images, respectively takes about {1 minute, 3.5 minutes, 3 minutes} (noisy skeleton) and {14 minutes, 62 minutes, 55 minutes} (full mask) for a single sample.} We plot the segmentation performance and the annotation time consumption of all evaluated methods, including all WSL methods, NLL methods, and YoloCurvSeg, under both single-sample and all-sample conditions in Fig. \ref{fig:time}. Our method achieves the highest segmentation performance ($\geq$ 97\% of FS) with the lowest annotation time cost ($<$ 0.3\% of FS) across all four tasks.

\subsection{Comparison with SOTA FS methods and Discussion}

In previous experiments and analyses, we utilize the vanilla U-Net as the segmentation network architecture in each compared method, for a fair comparison purpose. In order to further demonstrate the practicality and scalability of YoloCurvSeg, we conduct both quantitative and qualitative comparisons and analyses with incorporations of more advanced segmentation networks and frameworks, especially those designed for curvilinear structure segmentation, and discuss some potential future directions for improvement. Concretely, we explore the fully-supervised performance of two more advanced CNN U-Net variants, namely EfficientUNet \citep{tan2019efficientnet} and CS$^2$-Net \citep{mou2021cs2}, as well as four Vision Transformer based segmentation networks, namely SwinUNet \citep{cao2023swin}, TransUNet \citep{chen2021transunet}, UTNet \citep{gao2021utnet} and MedFormer \citep{gao2022data}, on the OCTA500, CORN, and DRIVE datasets. { All networks are trained with the same hyperparameters, including initial learning rate, learning rate policy, optimizer, batch size, etc., following the specifications of our vanilla U-Net Segmenters, as described in previous sections, to ensure a fair comparison.} The quantitative results are listed in Table \ref{tab:sota}. It can be observed that plausibly due to maintaining a superior training paradigm in all comparisons, such as the optimizer and the learning rate schedule strategies, the vanilla U-Net does not exhibit significant gaps compared to the best methods for all three datasets. It even surpasses more advanced networks in some cases, which is consistent with findings reported in the nnU-Net paper \citep{isensee2021nnu}. Among all comparisons, CS$^2$-Net and TransUNet respectively achieve the best and the second-best overall performance. We thus explore replacing the two-stage Segmenters in YoloCurvSeg with those two better performing networks. The experimental results show that using more advanced architectures can further improve the segmentation performance. 

To elucidate the gap between YoloCurvSeg and SOTA FS methods, we conduct qualitative comparisons on some examples in Fig. \ref{fig:sota}. It can be observed that for structures with low image contrast or small peripheral vessels/nerves, such as the areas outlined by the red circles, our method shows certain degrees of gap in accuracy and structural coherence compared to the well-performing FS methods. This is mainly due to the morphological gap and intensity gap between synthetic images and real images. Potential solutions to address these issues include fine-tuning the hyperparameters of the first component of YoloCurvSeg to generate curves that better match real shapes. In addition, introducing new paradigms in image translation/synthesis, such as diffusion models \citep{ho2020denoising,cheng2023learning}, may further enhance the realism of the synthesized images. Improving the network structure by introducing various attention mechanisms, especially self-attention mechanisms, or defining objective functions that maintain topology \citep{cheng2021joint,shit2021cldice}, may further enhance the performance of our framework. The former has already been validated in the experiments in Table \ref{tab:sota}. Another future direction worth exploring is to employ noisy label learning methods \citep{zhang2018generalized,yang2022learning} to train the fine stage's Segmenter since the generated pseudo-labels are inevitably noisy.

\begin{figure}[t]
    \centering
    \includegraphics[width=\linewidth]{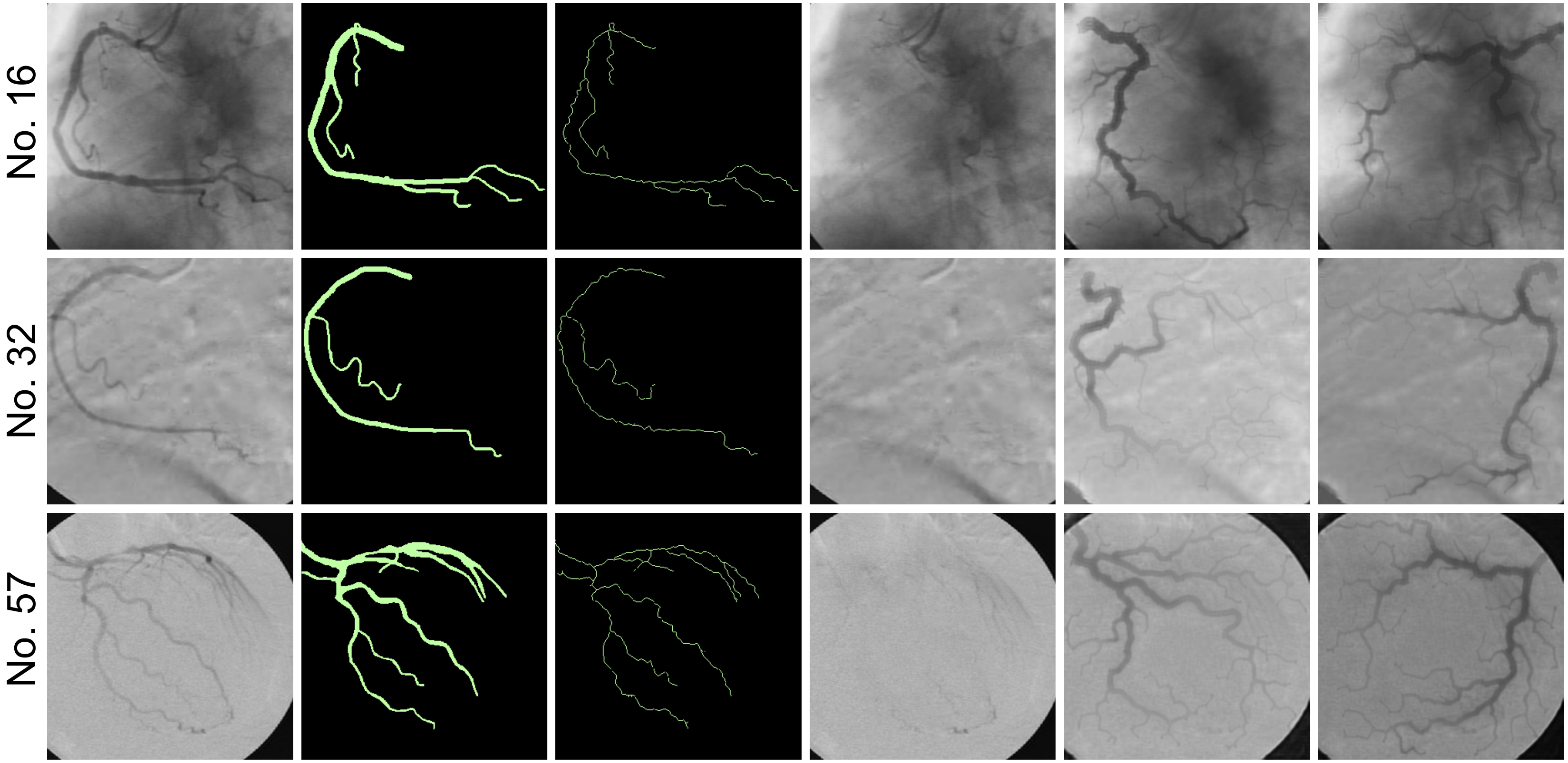}
    \vspace{-0.5cm}
  {\caption{Visualization of synthetic data from the DCA1 dataset. From left to right are examples of the original image, the full mask, the noisy skeleton, the extracted background and two synthesized images.}\label{fig:angio}}
  \vspace{-0.2cm}
  \end{figure}

Through extensive experiments, we have demonstrated that YoloCurvSeg can be applied to the two most common 2D curve structure segmentation tasks of three different modalities, the two structures being nerve fibers and retinal vessels, with good generalization. {To further demonstrate the scalability of the proposed pipeline, we conduct additional synthesis and segmentation validation analyses on an X-ray coronary angiography dataset, namely DCA1 \citep{cervantes2019automatic}. The first 100 samples from the dataset are used as the training and validation set, with the remaining 34 samples serving as the test set. All images are resized to $320\times320$. We arbitrarily select three samples from the DCA1 training set for fully supervised training and a comparison with YoloCurvSeg's one-shot performance. Concurrently, we also provide the performance of the fine segmenters, with the experimental results being detailed in Table \ref{tab:angio}. Compared to the fully-supervised setting using all samples and full masks, YoloCurvSeg still achieves exceptionally high one-shot performance, and this is accomplished without precisely tuning the Curve Generator of YoloCurvSeg. The three samples from the dataset and the corresponding synthesized images from YoloCurvSeg can be observed in Fig. \ref{fig:angio}.} Other similar and potentially transferable scenarios include cell membrane, crack, road (in aerial images) and leaf vein segmentation, etc.
With that being said, we acknowledge the challenge of applying and transferring our proposed YoloCurvSeg to specific curvilinear structure segmentation tasks in 3D scenarios such as brain vessel segmentation and cardiac vessel segmentation, as demonstrated in the works of Vessel-CAPTCHA \citep{dang2022vessel} and Examinee-Examiner Network \citep{qi2021examinee}. This is a major direction for future exploration. The current challenges mainly lie in the migration of YoloCurvSeg's second and third components, namely the Inpainter and the Multilayer Patch-wise Synthesizer, to 3D scenarios, which requires careful modification and design of the network's input format and size to balance their performance and the computational cost. Such explorations to some extent go beyond the scope of this work.

\begin{table}[t]
  \centering
  \setlength{\tabcolsep}{0.7mm}
  \footnotesize
  \renewcommand\theadgape{\Gape[1.8mm][0mm]}
  {\caption{YoloCurvSeg's performance on the DCA1 dataset. \textbf{FS} denotes fully-supervised learning.}  
  \label{tab:angio}}
  \renewcommand\arraystretch{1.15}
  \tiny
  \resizebox{0.65\linewidth}{!}{
  \begin{tabular}{cccc}
  \specialrule{0.1em}{0pt}{0pt}
  
   Method    & Label& DSC$\uparrow$            & ASSD$\downarrow$           \\ \hline
    Ours (Coarse, No. 16) & 1S & 69.81$\pm$5.30 & 6.06$\pm$3.00 \\
    FS (one, No. 16) & 1M & 59.02$\pm$9.94 & 8.07$\pm$4.68 \\ \hline
    Ours (Coarse, No. 32) & 1S & 69.96$\pm$5.06 & 5.71$\pm$2.80 \\
    FS (one, No. 32) & 1M & 48.92$\pm$14.05 & 15.18$\pm$9.45 \\ \hline
    Ours (Coarse, No. 57) & 1S & 70.68$\pm$5.83 & 5.71$\pm$3.46 \\
    FS (one, No. 57) & 1M & 26.05$\pm$17.62 & 20.59$\pm$9.15 \\ \hline
    Ours (Fine, No. 16) & 1S & 73.66$\pm$4.54 & 4.51$\pm$2.34 \\
    Ours (Fine, No. 32) & 1S & 73.80$\pm$4.11 & 4.31$\pm$2.26 \\
    Ours (Fine, No. 57) & 1S & 75.23$\pm$4.24 & 3.82$\pm$1.97 \\
    FS\cellcolor{gray!30} & 100M\cellcolor{gray!30} & 79.16$\pm$3.35\cellcolor{gray!30}  & 3.32$\pm$2.39\cellcolor{gray!30} \\
    \specialrule{0.1em}{0pt}{0pt}
  \end{tabular}
  }
  \end{table}

{ One concern regarding the YoloCurvSeg pipeline is whether the performance advantage arises from the overall framework's parameters or whether it introduces unfairness in comparison to other methods. In Table \ref{tab:para}, we provide the total parameters of the training and testing models between the YoloCurvSeg pipeline and comparative methods. We acknowledge that there may exist a certain degree of unfairness in our comparison. However, it is important to emphasize that this parameter comparison is to some extent, merely illustrative, as certain methods, despite not augmenting the model's trainable parameters, necessitate additional non-training parameters and data processing, such as tree filters and minimum spanning tree calculations in Tree Energy Loss, and energy field computations in Gated CRF, etc. Additionally, the first three components of YoloCurvSeg primarily contribute to the image synthesis process and are somewhat decoupled from the segmentation task. They can be considered as preprocessing steps that do not directly participate in the training of the segmenter. This distinguishes YoloCurvSeg from other methods that may incorporate additional trainable or non-trainable parameters in the segmentation process. We ensure fairness in the segmentation model and its training paradigm as much as possible. Furthermore, during testing or prediction in YoloCurvSeg, only the segmenter's involvement is required, which aligns with the majority of all compared methods.
}

  \begin{table}[t]
    \centering
    \setlength{\tabcolsep}{0.7mm}
    \footnotesize
    \renewcommand\theadgape{\Gape[1.8mm][0mm]}
    {\caption{Comparison of the total parameters in the training and testing phases between the YoloCurvSeg pipeline and comparative methods. Inp: Inpainter; Syn: Synthesizer; Seg: Segmenter.}  
    \label{tab:para}}
    \renewcommand\arraystretch{1.15}
    \tiny
    \resizebox{\linewidth}{!}{
    \begin{tabular}{ccc}
    \specialrule{0.1em}{0pt}{0pt}
    
     Method    &  Training Params (M)            & Testing Params (M)           \\ \hline
     pCE, GCE, RW & 1.81& 1.81\\
     S2L, MLoss, EM &1.81 &1.81 \\
     Dense CRF, Gated CRF &1.81 &1.81 \\
     AC, Tree Energy &1.81 &1.81 \\
     DBDM, DAST &2.45 & 2.45\\
     COT &1.81$\times$2=3.62 &1.81$\times$2=3.62 \\
     TriNet &1.81$\times$3=5.43 &1.81$\times$3=5.43 \\
     USTM, CLSLS &1.81$\times$2=3.62 & 1.81 \\
     YoloCurvSeg & 19.07(Inp)+11.38(Syn)+1.81(Seg)=32.26 & 1.81\\
      \specialrule{0.1em}{0pt}{0pt}
    \end{tabular}
    }
    \end{table}

\section{Conclusion}
This paper presents a novel sparsely annotated segmentation framework for curvilinear structures, named YoloCurvSeg. YoloCurvSeg is an image synthesis based pipeline comprising of a Curve Generator, an Inpainter, a Synthesizer and a two-stage Segmenter. Extensive experiments are conducted on four publicly accessible datasets, with superiority of our proposed framework being successfully established. Potential future directions are transferring YoloCurvSeg to 3D scenarios and exploring a better pipeline to further reduce the domain gap between synthetic images and real images.

\section*{Declaration of competing interest}

The authors declare that they have no known competing financial interests or personal relationships that could have appeared to influence the work reported in this paper.

\section*{Acknowledgments}
This study was supported by the Shenzhen Basic Research Program JCYJ20190809120205578; the National Natural Science Foundation of China 62071210; the Shenzhen Science and Technology Program RCYX20210609103056042; the Shenzhen Basic Research Program JCYJ20200925153847004; the Shenzhen Science and Technology Innovation Committee Program KCXFZ2020122117340001. 

\section*{Data availability}

The original datasets used in this paper are publicly available. The source code, generated skeleton annotations and synthetic datasets will be made available at \href{https://github.com/llmir/YoloCurvSeg}{https://github.com/llmir/YoloCurvSeg} upon acceptance.






\bibliographystyle{model2-names.bst}\biboptions{authoryear}
\balance
\bibliography{refs}



\end{document}